%% file: main.tex
\documentclass[11pt]{article}
\usepackage[top=1 in, bottom=1 in, left=0.85 in, right=0.85 in]{geometry}
\input{math_commands}

\usepackage{float}
\usepackage{times}
\usepackage{mathtools}
\usepackage[round]{natbib}
\usepackage{makecell}
\usepackage{amsmath, amsthm, amssymb,graphicx,url}
\usepackage{thmtools,thm-restate,wrapfig,enumitem,mathabx}
\usepackage{booktabs}
\usepackage{color}
\usepackage{colortbl}
\usepackage{multirow}

\usepackage[utf8]{inputenc} 
\usepackage[T1]{fontenc}    
\usepackage{hyperref}       
\usepackage{url}            
\usepackage{booktabs}       
\usepackage{amsfonts}       
\usepackage{nicefrac}       
\usepackage{microtype}      
\usepackage{xcolor}         

\usepackage{caption}
\usepackage{subcaption}

\usepackage{algorithm}
\usepackage[noend]{algorithmic}
\usepackage{titletoc}
\usepackage{bbm}
\usepackage{diagbox}   
\usepackage{tikz}
\usetikzlibrary{calc}
\usepackage{etoc}

\newtheorem{assumption}{Assumption}

\newtheorem{remark}{Remark}
\newtheorem{lemma}{Lemma}

\newcommand{\prob}{\mathbb{P}}
\newcommand{\bp}{\mathrm{bp}}
\newcommand{\mix}{\mathrm{mix}}

\title{Zeroth-Order Policy Optimization for Scheduling in Queueing Networks}
\author{
    Deyi Wang \\
	University of Michigan, Ann Arbor\\
	\texttt{deyiwang@umich.edu} \\
	\and
	Qining Zhang \\
	University of Michigan, Ann Arbor\\
	\texttt{qiningz@umich.edu} \\
    \and
	Lei Ying \\
	University of Michigan, Ann Arbor\\
	\texttt{leiying@umich.edu} \\
}
\date{}
\begin{document}

\maketitle

\begin{abstract}
    Reinforcement learning in queueing network scheduling presents significant challenges because of the countably infinite state space and the high stochasticity in transitions. In this paper, we propose zeroth-order policy optimization algorithms that are easy to implement and rely solely on estimates of the average cost, a single scalar quantity, without requiring the estimation of value or advantage functions.
    Our analysis focuses on back-pressure guarded policies and establishes uniform smoothness of the average cost objective with respect to the policy parameters.
    Building on this property, we establish finite-time convergence guarantees to an $\epsilon$-stationary policy for the proposed algorithms.
    Numerical experiments on multiple queueing networks, including systems beyond classical exponential service distribution, demonstrate competitive performance, outperforming gradient-based RL methods like PPO and traditional queueing baselines.
\end{abstract}
\etocdepthtag.toc{main}

\input{sections/introduction}

\input{sections/problem_formulation}

\input{sections/ZOalg}

\input{sections/main_results}

\input{sections/experiments}

\bibliographystyle{unsrtnat}
\bibliography{refs,inlab-refs}

\appendix
\etocdepthtag.toc{appendix}
\input{sections/appendix}

\end{document}

%% file: math_commands.tex
\usepackage{amsmath,amsfonts,bm}

\def\eqref#1{equation~\ref{#1}}

\def\1{\bm{1}}

\def\rr{{\textnormal{r}}}

\def\rx{{\textnormal{x}}}
\def\ry{{\textnormal{y}}}
\def\rz{{\textnormal{z}}}

\def\rvu{{\mathbf{i}}}

\def\rvu{{\mathbf{u}}}

\def\ervu{{\textnormal{u}}}

\def\vzero{{\bm{0}}}
\def\vone{{\bm{1}}}
\def\vmu{{\bm{\mu}}}
\def\vtheta{{\bm{\theta}}}
\def\vlambda{{\bm{\lambda}}}

\def\vnu{{\bm{\nu}}}

\def\va{{\bm{a}}}

\def\vc{{\bm{c}}}

\def\ve{{\bm{e}}}

\def\vp{{\bm{p}}}
\def\vq{{\bm{q}}}

\def\vs{{\bm{s}}}

\def\vu{{\bm{u}}}
\def\vv{{\bm{v}}}
\def\vw{{\bm{w}}}
\def\vx{{\bm{x}}}
\def\vy{{\bm{y}}}

\def\mI{{\bm{I}}}

\def\mP{{\bm{P}}}

\def\mR{{\bm{R}}}

\def\mX{{\bm{X}}}

\DeclareMathAlphabet{\mathsfit}{\encodingdefault}{\sfdefault}{m}{sl}
\SetMathAlphabet{\mathsfit}{bold}{\encodingdefault}{\sfdefault}{bx}{n}

\def\gA{{\mathcal{A}}}

\def\gC{{\mathcal{C}}}
\def\gD{{\mathcal{D}}}

\def\gN{{\mathcal{N}}}

\def\gQ{{\mathcal{Q}}}

\def\gU{{\mathcal{U}}}

\def\gX{{\mathcal{X}}}

\def\sP{{\mathbb{P}}}

\def\sR{{\mathbb{R}}}

\def\sZ{{\mathbb{Z}}}

\def\emR{{R}}

\newcommand{\E}{\mathbb{E}}

\DeclareMathOperator*{\argmax}{arg\,max}

%% file: sections/introduction.tex
\section{Introduction}

Queueing network scheduling problem can be formulated as an average-cost Markov decision process, making reinforcement learning (RL) a promising approach to policy optimization for queueing systems (see, e.g., \cite{Dai_2022,LiuXieMod22,WeiLiuWan_23,CheDonNam24,murthy2024}). However, infinite buffers lead to countably infinite state spaces, and long-run-average performance depends on policy-dependent stationary distributions that are often difficult to characterize. These features make both policy evaluation and optimization challenging.

Policy-gradient methods have shown promising results. In particular, \cite{Dai_2022} developed PPO with variance reduction and establish local policy-improvement bounds, without establishing the convergence to a \(\epsilon\)-stationary policy. Local improvement alone does not guarantee sufficient progress toward stationarity. \cite{murthy2024} derived optimality-gap bounds for {\em tabular} Natural Policy Gradient (NPG) updates with state-dependent step sizes, but the guarantees only hold for tabular NPG. Both approaches require estimating value or advantage functions over the infinite state spaces. 

This paper develops two zeroth-order policy optimization algorithms that require only the average-cost (a scalar value), without estimating the value or advantage functions over the infinite state space. Our analysis considers back-pressure guarded policies, which use learned scheduling when the total queue length is below a threshold and back-pressure scheduling otherwise to guarantee stability when traffic load is strictly within the stability region. For this class of policies, we establish {\em uniform smoothness} of the average-cost objective with respect to the policy parameters, under standard regularity assumptions on finite-dimensional policy parameterizations. We further prove the finite-time convergence to a \(\epsilon\)-stationary policy for the proposed algorithms under exact average-cost evaluation. Experiments across different queueing networks demonstrate competitive performance, with improvements over PPO and priority-based benchmarks on more complex networks.

%% file: sections/problem_formulation.tex
In the rest of the paper, for any positive integer $N$, we use $[N]$ to denote the set $\{1,2,\ldots,N\}$. 
We use boldface $\vx$ to denote vectors, and boldface capital $\mX$ to denote matrices. 
The $i$-th entry of vector $\vx$ is denoted using $x_i$, and the element on the $i$-th row and $j$-th column of matrix $\mX$ is denoted using $X_{ij}$. 
Sets are denoted with calligraphic $\gX$. For scalar $x$, we use $[x]^+$ to denote $\max\{x,0\}$.
\section{Problem Formulation} \label{section: problem formualtion}

This section introduces the multi-class queueing network model, the average-cost Markov Decision Process (MDP) for the queueing network, and the class of stable policies considered in the paper.
\subsection{Multi-Class Queueing Networks}

\begin{figure}[h]
    \centering
    \includegraphics[width=0.5\linewidth]{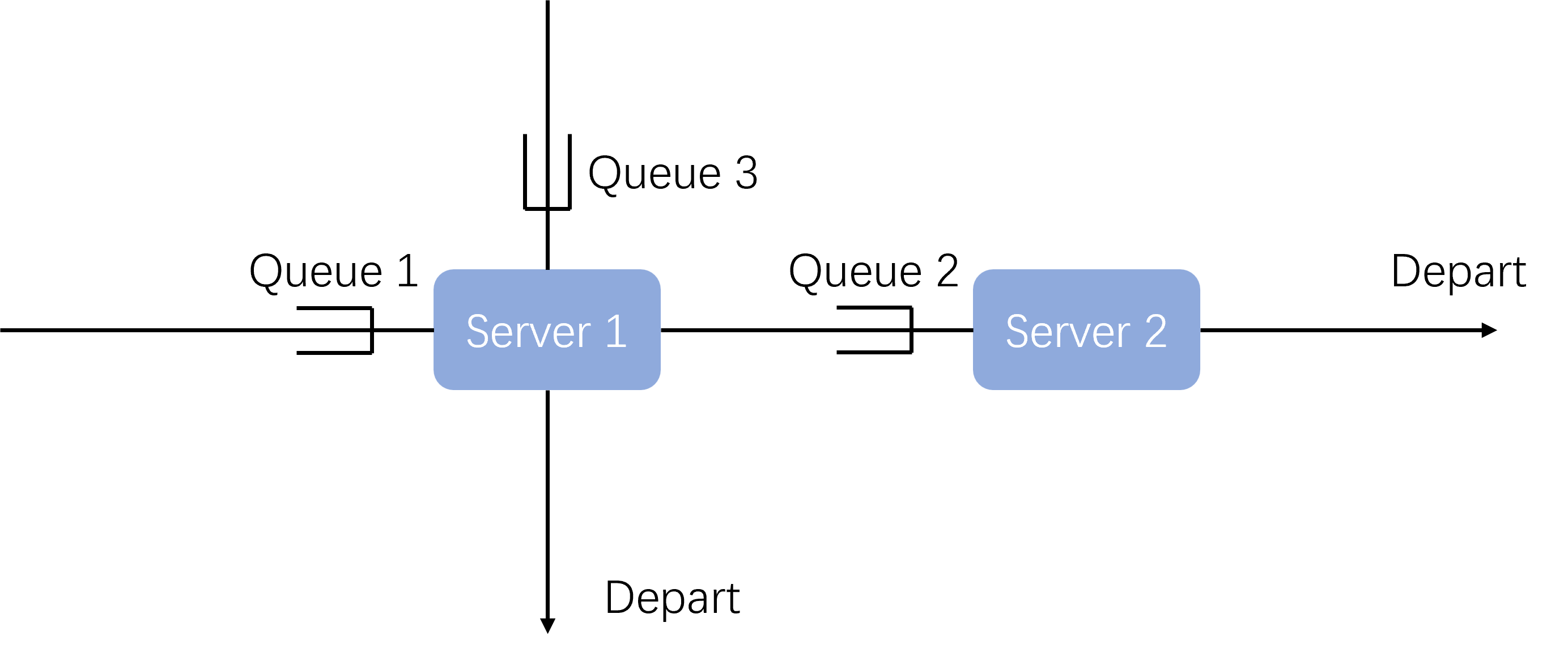}
    \caption{Illustration of a criss-cross queueing network.}
    \label{fig:queue_illustration}
\end{figure}

We consider a discrete-time open multi-class queueing network, with time indexed by slots $t\in\{1,2,\ldots\}$. The network consists of $N$ queues with infinite-buffers and $M$ servers, where $M\leq N$. The queues are indexed by $[N]$ and the servers by $[M]$. For each server $m\in[M]$, let $\gN_m$ denote the set of queues that can be served by server $m$. Each queue is associated with exactly one server; hence, the collection $\{\gN_m\}_{m=1}^M$ forms a partition of $[N]$. A server may serve multiple queues, with the scheduling decision determining which of its associated queues to serve at the beginning of each time slot.
Figure~\ref{fig:queue_illustration} shows a illustration for the simple criss-cross queueing network.
In this network, there are $N=3$ queues and $M=2$ servers, and $\gN_1 = \{1,3\}$, $\gN_2 = \{2\}$.

\noindent\textbf{Arrival and Service processes:}
Let $q_i(t)$ denote the number of jobs in queue $i$ at the beginning of time slot $t$, and let
$
\vq(t)=[q_1(t),\ldots,q_N(t)]^\top
$
denote the queue-length vector. For each queue $i\in[N]$, let $a_i(t)\in\{0,1\}$ denote the number of exogenous arrivals to queue $i$ at the beginning of time slot $t$. For each $i$, the exogenous arrival process $\{a_i(t)\}_{t\geq 1}$ is i.i.d. Bernoulli with mean $\lambda_i\in(0,1)$, i.e.,
$
a_i(t)\sim\operatorname{Ber}(\lambda_i).
$
Let $\va(t) = [a_1(t), \cdots, a_N(t)]^\top$. Let $u_i(t)\in\{0,1\}$ denote the scheduling decision for queue $i$ in slot $t$. If $u_i(t)=1$ and queue $i$ is nonempty, the head-of-line job in queue $i$ will be served by the associated server during time slot $t$.  Let $\vu(t) = [u_1(t),\cdots, u_N(t)]^\top$. Notice that a server can only serve at most one job at a single time slot. 
So the set of feasible scheduling actions can be defined as:
\begin{align}\label{eq:feasible-actions}
    \gU = \left\{\vu \in \{0,1\}^N: \sum_{i \in \gN_m} u_i \leq 1,~\forall m \in [M] \right\}
\end{align}
 Let $s_i(t)\in\{0,1\}$ be the potential service-completion process independent of the scheduling decision, i.e., whether a job in queue $i$ completes its service during slot $t$ if being served. We assume a geometric and preemptive service model where $s_i(t)$ is an i.i.d. Bernoulli process with rate $\mu_i\in(0,1)$, i.e., $s_i(t) \sim\operatorname{Ber}(\mu_i)$. Let $\vs(t) = [s_1(t),\cdots, s_N(t)]^\top$. Both the arrival process $\va(t)$ and the potential service-completion process $\vs(t)$ are independent of each other and across queues. For simplicity of notation, let $\tilde{s}_i(t)\in\{0,1\}$ be the actual service-completion indicator of whether a job at queue $i$ will finish service, which requires queue $i$ is scheduled and nonempty and the head-of-line job completes its service, so $ \tilde{s}_i(t)= \mathbbm{1}\{q_i(t)>0\}u_i(t)s_i(t)$. Similarly, let $\tilde{\vs}(t) = [\tilde{s}_1(t), \cdots, \tilde{s}_N(t)]^\top$. We assume the exogenous arrival rate $\vlambda=[\lambda_1, \cdots, \lambda_N]^\top$ is unknown, but the service rate $\vmu=[\mu_1, \cdots, \mu_N]^\top$ is known.

\noindent\textbf{Job Routing:} After a job completes service in queue $i$, it either leaves the system or is routed to a unique downstream queue for further service following a deterministic route. Let $\mR\in\{0,1\}^{N\times N}$ be the routing matrix where $\emR_{ij} = 1$ means that a job in queue $i$ is routed to queue $j$ after completing its service. We assume exogenous arrivals and jobs routed from upstream queues during slot $t$ are added to the queue lengths at the beginning of slot $t+1$ so that the queue evolution can be written as
\begin{align*}
    q_i(t+1) =& q_i(t) + a_i(t) - \tilde{s}_i(t) + \sum_{j=1}^N R_{ji} \tilde{s}_j(t)\\
    =& q_i(t) + a_i(t) - \mathbbm{1}\{q_i(t)>0\}u_i(t)s_i(t) + \sum_{j=1}^N R_{ji} \mathbbm{1}\{q_j(t)>0\}u_j(t)s_j(t)
\end{align*}
or in the vector form
\begin{align}\label{eq:queue-recursion}
    \vq(t+1)
    =
    \vq(t)+\va(t)+(\mR^\top-\mI)\tilde{\vs}(t).
\end{align}
Notice that $\sum_{j=1}^N \emR_{ij}\in \{0,1\}$ and it is $0$ when a job in queue $i$ leaves the system after service and $1$ if the job is routed to a unique downstream queue. We assume the network is open: starting from any queue, a job will eventually leave the system after finitely many routing transitions. 

\subsection{Average-Cost Markov Decision Process (MDP) Formulation}
This paper considers a queueing scheduling problem formulated as an average-cost Markov decision process, denoted as a tuple $(\gQ, \gU, P, c)$, where $\gQ = \sZ_{\geq 0}^N$ is the state space representing all possible queue lengths of the $N$ queues, $\gU$ is the action space representing the set of feasible scheduling actions in~\eqref{eq:feasible-actions}, 
and $P$ is the transition kernel induced by the exogenous arrivals, service completions, and deterministic job routing following the queueing recursion in~\eqref{eq:queue-recursion}, i.e.,
\begin{align*}
    P(\vq' | \vq, \vu) =& \prob\left( \vq(t+1) = \vq'|\vq(t) = \vq, \vu(t) = \vu \right)\\
    =& \prob\left( \vq+\va(t)+(\mR^\top-\mI)\tilde{\vs}(t) = \vq'|\vq(t) = \vq, \vu(t) = \vu \right).
\end{align*}

At each time $t$, the holding cost is defined as the weighted total queue length, i.e., $c(\vq(t)) = \sum_{i=1}^N c_i q_i(t)$ for some $\vc = [c_1, c_2, \dots, c_N]^\top \in (0,1]^N$.
For general positive holding cost weights $\vc \notin (0,1]^N$, we can normalize the costs by dividing each weight by $c_{\max}=\max_i c_i,$ which does not change the optimal solution to the problem. 
A stationary policy $\pi$ specifies, for each queue-length vector
$\vq\in\mathbb{Z}_+^N$, a probability distribution over the feasible scheduling actions $\gU$. In particular, $\vu(t) \sim \pi(\cdot\mid\vq(t))$. For a policy $\pi$, let the average cost be
\begin{align*}
    J(\pi) = \limsup_{T\to \infty}\frac{1}{T}\E_{\pi}\left[\sum_{t=1}^T c(\vq(t)) \right],
\end{align*}
where the expectation is taken with respect to the randomness in the
arrival processes, service processes, and the scheduling policy. Let $\Pi$ denote the set of all such admissible stationary scheduling policies. 

\begin{remark} \label{remark: MDPformulation}
    Although we focus on discrete-time queueing networks with Bernoulli arrivals and geometric service times, the MDP formulation above is not restricted to this setting.
    A continuous-time queueing network with Poisson arrivals and exponential service times can also be represented as a discrete-time MDP through uniformization. See, e.g.,~\cite{Dai_2022}.
    The resulting MDP differs from the one considered above primarily in its transition kernel $P$, which is determined by the uniformized transition probabilities. Our analysis can be easily generalized to bounded arrival distributions beyond Bernoulli such that the probabilities of no arrival or exactly one job arrival are both strictly positive. 
\end{remark}

\subsection{Stability Region}
The average-cost problem is only meaningful when there exists some policy to stabilize the queues, i.e., the traffic load is within the stability region because otherwise the average cost is $\infty$. Next, we  introduce the stability region. Given the exogenous arrival rate vector $\vlambda,$ define $\tilde{\lambda}_i$ 
such that \[
    \tilde{\vlambda} = \vlambda + \mR^\top \tilde{\vlambda},
\]
which is the effective arrival rate vector to the queues, including both exogenous arrivals and jobs routed from upstream queues.
Since the network is open, according to~\citep{Bramson2008}, $\mI - \mR^\top$ is invertible and $(\mI - \mR^\top)^{-1} = \sum_{n = 0}^\infty (\mR^\top)^n.$ Hence $\tilde{\vlambda} = (\mI - \mR^\top)^{-1} \vlambda$ is well-defined. 
For each server $m\in [M]$, we can define the load as $\varrho_m \coloneqq \sum_{i\in \gN_m} \tilde{\lambda}_i / \mu_i$.
The {\bf stability region} of the queueing network  is 
\begin{align*}
    \gC = \left\{ \vlambda \ge \vzero: \varrho_m \le 1, \forall m \in [M] \right\}
\end{align*} because no scheduling algorithm can stabilize the system if $\vlambda \not\in\gC$ \cite{SriYin_14,DaiHar20}. 

\begin{assumption}\label{assumption:stability-condition}
    We assume the traffic load is strictly within the stability region, i.e., $\varrho_m<1$, for all $m\in [M]$.
\end{assumption}

Under Assumption~\ref{assumption:stability-condition}, the system is strictly underloaded and $\varrho_m = \sum_{i\in \gN_m}{\tilde\lambda_i}/{\mu_i} < 1,$ where $\tilde{\vlambda} = (\mI - \mR^\top)^{-1}\vlambda $.
Now given $\delta>0,$ let $\tilde{\vlambda}(\delta) = (\mI - \mR^\top)^{-1}(\vlambda + \delta \vone)$.
We define the traffic slackness constant $\delta_c>0$ to be
\begin{align}\label{eq:traffic-slackness}
    \delta_c \coloneqq \max \left\{ {\delta>0}: \sum_{i\in \gN_m} \frac{\tilde{\lambda_i}(\delta)}{\mu_i} \leq 1 ,\forall m \in [M] \right\}.
\end{align}

\subsection{Back-Pressure Guarded Policy} \label{section: policies}
\textbf{Back-pressure action:} we first introduce the celebrated back-pressure policy \cite{TasEph_92}, which is throughput optimal. For a queue-length vector $\vq\in \gQ$, the back-pressure policy selects, at each server, the non-empty queue with the largest nonnegative service-rate-weighted job pressure to downstream queues. Specifically, let the back-pressure weight of queue $i$ be
\begin{align*}
    w_i(\vq) = \mu_i\left[ q_i - \sum_{j=1}^N R_{ij}q_j \right]\mathbbm{1}\{q_i>0\}, \quad i\in[N].
\end{align*}
Here, $\sum_{j=1}^N \emR_{ij}q_j$ is the queue length of the unique downstream queue of queue $i$, and is zero if a job completing service at queue $i$ leaves the system. The back-pressure policy selects the action $\vu_{\mathrm{bp}}$ that maximizes the total back-pressure weight, i.e., 
\begin{align*}
    \vu_{\bp}(\vq) \in \argmax_{\vu \in \gU} \sum_{i=1}^N w_i(\vq) u_i,
\end{align*}
where a fixed tie-breaking rule is used whenever the maximizer is not
unique. Let $\pi_{\bp}$ be the classic back-pressure policy which selects $\vu_{\bp}(\vq)$ for all $\vq \in \gQ$. The back-pressure policy is a stationary policy and it has been proved in \cite{TasEph_92} that it is throughput optimal and stabilizes all queues when the load within the stability region.  

\noindent\textbf{Back-pressure guarded policies:} This paper considers a class of stationary policies that use a learnable scheduling policy $\pi_\vtheta$ when the total queue length is below a prescribed threshold $H$ and switch to back-pressure when the total queue length is above the threshold. For the learnable policies, we consider ``weakly'' work-conserving policies such that at least one of the servers will serve a non-empty queue when at least one of the queues is nonempty. Note that any work-conserving policy is also weakly work-conserving. 
For any queue vector $\vq$ such that $\|\vq\|_1>0$, define the set of weakly work-conserving actions as:
\begin{align}\label{eq:work-conserving-actions}
    \gU_{\mathrm{wwc}}(\vq) = \left\{\vu \in \{0,1\}^N: \sum_{i \in \gN_m} u_i \le 1\ \forall m, ~\exists i' \in [N]~s.t.~u_{i'}  \mathbbm 1 \left\{ q_{i'} > 0 \right\}=1 \right\}.
\end{align}
Let the set of stationary weakly work-conserving policies be $\Pi_{\rm WWC}\subset \Pi$. Let $\pi_\vtheta\in \Pi_{\rm WWC}$ denote a weakly work-conserving policy parameterized by $\vtheta\in \sR^d$, where $d$ is the number of parameters.
For a threshold $H<\infty$ and a stationary policy $\pi_\vtheta$, we further define the corresponding back-pressure guarded policy $\pi_{H,\vtheta}$ by
\begin{align*}
    \pi_{H,\vtheta}(\vu\mid\vq)
    \coloneqq
    \begin{cases}
        \pi_{\vtheta}(\vu\mid\vq),
        & \|\vq\|_1\leq H,\\
        \mathbbm{1}\{\vu=\vu_{\bp}(\vq)\},
        & \|\vq\|_1>H.
    \end{cases}
\end{align*}
In other words, the policy follows policy $\pi_\vtheta$
when the queues are within the bounded set $\{\vq\in\gQ:\|\vq\|_1\leq H\}$, and follows the back-pressure policy outside. Let
$\Pi_H=\{\pi_{H,\vtheta}: \vtheta \in \sR^d\}\subseteq \Pi_{\rm WWC}$ denote the resulting class of back-pressure guarded policies.
Proposition~\ref{prop:well_defined_J} shows the properties of any policy in $\Pi_H$.
We include the proof of Proposition~\ref{prop:well_defined_J} in Appendix~\ref{section: prop:well_defined_J: proof}.

\begin{restatable}{proposition}{propwelldefinedJ} \label{prop:well_defined_J}
   Under Assumption~\ref{assumption:stability-condition}, the Markov chain under any back-pressure-guarded policy in $\pi_{H,\vtheta}\in\Pi_{H}$ is irreducible, aperiodic, and positively recurrent. Furthermore, $J(\pi_{H,\vtheta})$ is finite.
\end{restatable}

\subsection{Value Functions and Q-Functions}
Based on Proposition \ref{prop:well_defined_J}, the Markov chain under any policy $\pi \in \Pi_H$ has a unique stationary distribution $\nu^\pi(\vq)$ and $J(\pi)$ is finite. 
For any policy $\pi \in \Pi_H$, define the (relative) value function $V^\pi: \gQ \mapsto \sR$ by for any $\vq \in \gQ$,
\begin{align*}
    V^\pi(\vq)=\sum_{t=0}^\infty \left(\E_\pi[c(\vq(t))|\vq(0)=\vq]-\E_\pi[c(\vq(t))|\vq(0)=0]\right).
\end{align*}
Proposition~\ref{prop: well_defined_value} shows that the (relative) value function $V^\pi$ is well defined and satisifies the Poisson equation.
We include the proof of Proposition~\ref{prop: well_defined_value} in Appendix~\ref{section: prop:well_defined_value:proof}.
\begin{restatable}{proposition}{propwelldefinedvalue}
    For any policy $\pi \in \Pi_H$, the (relative) value function $V^\pi$ is well defined.
    Specially, for any $\vq \in \gQ$,
    \[
    \left| V^\pi(\vq) \right| \le 2 C_\mix (W(\vq)+W(\vzero))\left( \frac{\|\vq\|_1}{1-\rho_\mix} + \frac{N \rho_\mix}{(1-\rho_\mix)^2} \right),
    \]
    where the function $W: \gQ \mapsto [1,\infty)$ is defined in Appendix~\ref{section: geometric mixing} in detail.    Moreover, the (relative) value function satisfies the Poisson equation, i.e., for $\vq \in \gQ$,
    \begin{align*}
        V^\pi(\vq) = c(\vq) - J(\pi) + \sum_{\vq'\in\gQ} P^\pi(\vq'|\vq) V^\pi(\vq').
    \end{align*}
        \label{prop: well_defined_value}
\end{restatable}
Given the (relative) value function $V^\pi,$ we then define the (relative) $Q$ function $Q^\pi: \gQ\times\gU\mapsto\sR$ and the advantage function $A^\pi:\gQ\times \gU \mapsto \sR$ to be
\begin{align*}
    Q^\pi(\vq,\vu) =&c(\vq) - J(\pi) + \sum_{\vq'\in\gQ} P(\vq'|\vq,\vu) V^\pi(\vq')\\
A^\pi(\vq, \vu) = & Q^\pi(\vq, \vu) - V^\pi(\vq).
\end{align*}
 The objective of this paper is to find the policy that minmize the average cost $J(\pi)$ within the policy class. 
We define 
\begin{align*}
    J^*=\inf_{\pi\in\Pi_H} J(\pi) = \inf_{\pi\in\Pi_H}\sum_{\vq} \nu^{\pi}(\vq) c(\vq).
\end{align*}

%% file: sections/ZOalg.tex
\section{Zeroth-Order Policy Optimization for Queueing Network Controls} \label{section: algorithm}

We propose the zeroth-order policy optimization method for the average-cost problem, which is based on a three-step approach: perturbation, evaluation and update. 
In particular,  we first perturb the current policy to another close policy, and then evaluate the performance of the perturbed policy.
If the perturbed policy performs better than the current policy, then we update the current policy towards the perturbed policy; otherwise, we retain the current policy and try another different perturbation at the next policy iteration. The algorithm is presented in Algorithm~\ref{alg:zeroth-order}.

\begin{algorithm}[h]
\caption{Zeroth-order policy optimization method for queueing network controls}
\label{alg:zeroth-order}
\begin{algorithmic}[1]
    \REQUIRE initialize the actor-network parameter $\vtheta_0$, learning rate $\alpha$, perturbation distance $\beta=\alpha$, and threshold $H$; 
    \FOR{iteration $k=0,1,\cdots K-1$}
        \STATE sample a random vector $\vv_k$ such that each coordinate $v_{k,i}$ for $i = 1, 2, \dots, d$ is sampled from $\{-1,+1\}$ equally likely and independently;
        \STATE obtain a perturbed policy $\pi_{\vtheta_k'}$ where $\vtheta_k' = \vtheta_k + \beta \vv_k$;
        \STATE evaluate the average cost of the original back-pressure guarded policy $J(\pi_{H, \vtheta_k})$;
        \STATE evaluate the average cosst of the the perturbed back-pressure guarded policy $J(\pi_{H, \vtheta_k'})$;
        \STATE update the actor network
        $$\vtheta_{k+1} = \vtheta_k + \alpha \mathbbm{1}\left\{ J(\pi_{H,\vtheta_k'}) < J(\pi_{H,\vtheta_k})\right\}\vv_k.$$
    \ENDFOR
\end{algorithmic}
\end{algorithm}

We next make a few remarks: 
\begin{enumerate}
    \item \textbf{Perturbation:} in each training iteration, we use Rademacher distribution for perturbation, which can be replaced with other distirbutions such as Guassian perturbation. 
    \item \textbf{Evaluation:} Algorithm~\ref{alg:zeroth-order} assumes perfect policy evaluation. Our algorithm only requires perfect evaluation of the average cost $J(\pi)$ and does not require the evaluation of the value function or the Q-function like in policy-gradient-based algorithms \citep{Dai_2022,murthy2024}. In other words, the algorithm only needs to evaluate a single scalar value for each policy instead of a function over its state and action spaces.  
    
    \item \textbf{Update:} Unlike regular zeroth-order methods which update the current policy to the opposite direction when the perturbed policy performs worse, Algorithm~\ref{alg:zeroth-order} only updates when the perturbed policy performs better, which guarantees monotonic performance improvement. 
\end{enumerate}

Inspired by the federated learning zeroth-order policy optimization method in~\cite{deyi2026_fl}, we also propose a partitioned learning variation of Algorithm~\ref{alg:zeroth-order}.
We partition the parameter set into $D$ subsets.
In each training iteration, we generate $D$ perturbed policies, each of which perturbs one of the subsets and leaves other parameters unchanged.
Then the algorithm evaluates each perturbed policy, and compare it with the current policy.
If the perturbed policy performs better, the algorithm updates the parameters of the corresponding subset; and otherwise, the corresponding parameters remain unchanged.

For the parameter indices set $[d]$, we use $\gD_1, \gD_2, \dots, \gD_D \subseteq [d]$ to denote a partition of $[d]$, i.e.,  $\bigcup_{j=1}^D \gD_j=[d]$ and  $\gD_j \cap \gD_{j'} = \emptyset$ for any $j\neq j'$. 
For each $\gD_j$, define a $d$-dimension vector $\ve_{\gD_j}$ such that $e_{\gD_j,i} = 1$ if $i \in \gD_j$; otherwise, $e_{\gD_j,i} = 0$.
We also define a vector operator $\circ$ that takes entry-wise product such that $\vv\circ\vu$ is a $d$-dimension vector with $(v\circ u)_i = v_iu_i$.
The algorithm is presented in Algorithm~\ref{alg:zeroth-order-partition}.
Instead of one potential update direction, Algorithm~\ref{alg:zeroth-order-partition} considers $2^D$ possible update directions.

\begin{algorithm}[h]
\caption{Partitioned zeroth-order policy optimization method for queueing network controls}
\label{alg:zeroth-order-partition}
\begin{algorithmic}[1]
    \REQUIRE initialize the actor-network parameter $\vtheta_0$, learning rate $\alpha$, perturbation distance $\beta$, threshold $H$, and a partition of the parameter indices $\gD_1, \gD_2, \dots, \gD_D \subseteq [d]$ for a given $D$;
    \FOR{iteration $k=0,1,\cdots K-1$}
        \STATE sample a random vector $\vv_k$ such that each coordinate $v_{k,i}$ for $i = 1, 2, \dots, d$ is sampled from $\{-1,+1\}$ equally likely and independently;
        \STATE obtain $D$ perturbed policies $\pi_{\vtheta_{k,j}'}$ for $j=1,2,\dots,D$, where $\vtheta_{k,j}' = \vtheta_k + \beta \vv_k \circ \ve_{\gD_j}$;
        \STATE evaluate the average cost of original back-pressure guarded policy $J(\pi_{H, \vtheta_k})$;
        \STATE evaluate the average costs of all the perturbed back-pressure guarded policies,i.e., $J(\pi_{H, \vtheta_{k,1}'}), J(\pi_{H, \vtheta_{k,2}'}), \dots, J(\pi_{H, \vtheta_{k,D}'})$;
        \STATE update the actor network
        $$\vtheta_{k+1} = \vtheta_k + \alpha \sum_{j=1}^D \mathbbm{1}\left\{ J(\pi_{H,\vtheta_{k,j}'}) < J(\pi_{H,\vtheta_k})\right\}\vv_k \circ \ve_{\gD_j}.$$
    \ENDFOR
\end{algorithmic}
\end{algorithm}

%% file: sections/main_results.tex
\section{Main Results and Proof Roadmaps} \label{section: main results}

In this section, we study the convergence of Algorithm~\ref{alg:zeroth-order} and Algorithm~\ref{alg:zeroth-order-partition}.
We first state the assumptions needed for theoretical analysis, and then provide the lemmas and theorems that characterize the convergence of Algorithm~\ref{alg:zeroth-order} and Algorithm~\ref{alg:zeroth-order-partition}.

\subsection{Main Results}

Besides the stability assumption (Assumption~\ref{assumption:stability-condition}),
we assume all policies $\pi_\vtheta$ are uniformly Lipschitz continuous and smooth with respect to the parameter $\vtheta$ for the states such that the total queue length $\| \vq \|_1 \le H.$
The assumption is formally stated as follows:
\begin{assumption}\label{assumption:smooth-policy}
    There exist constants $L_0, L_1 < \infty$ such that
    \begin{align*}
    &\sup_{\vq: \| \vq \|_1 \le H} \sum_{\vu \in \gU} \left| \pi_{\vtheta'} (\vu | \vq) - \pi_\vtheta (\vu | \vq) \right|
    \le
    L_0 \| \vtheta' - \vtheta \|_2,
    \qquad
    \forall \vtheta, \vtheta',\\
    &\sup_{\vq: \| \vq \|_1 \le H} \sum_{\vu \in \gU} \left| \pi_{\vtheta'} (\vu | \vq) - \pi_\vtheta (\vu | \vq) - \nabla_\vtheta \pi_\vtheta (\vu | \vq)^\top (\vtheta' - \vtheta) \right|
    \le
    \frac{L_1}{2} \| \vtheta' - \vtheta \|_2^2,
    \qquad
    \forall \vtheta, \vtheta'.
    \end{align*}
\end{assumption}
Assumption~\ref{assumption:smooth-policy} holds, for example, for the tabular softmax policy parameterization, as a direct corollary using Lemma 24 in~\cite{papini2022}.
Based on the assumption above, we will first show that the average cost $J(\pi_{H, \vtheta})$ is smooth with respect to the parameter $\vtheta$. 

\begin{restatable}{lemma}{smoothnesslemma} \label{lemma: smoothness}
    Under Assumptions~\ref{assumption:stability-condition} and \ref{assumption:smooth-policy} and with any policy in $\Pi_H,$
    the average cost $J(\pi_{H,\vtheta})$ is $L_J$-smooth with respect to $\vtheta$, i.e.,
    \[
    \left| J(\pi_{H,\vtheta'}) - J(\pi_{H,\vtheta}) - \nabla_\vtheta J(\pi_{H,\vtheta})^\top (\vtheta' - \vtheta) \right|
    \le
    \frac{L_J}{2} \| \vtheta' - \vtheta \|_2^2,
    \qquad
    \forall \vtheta, \vtheta'.
    \]
    for some $L_J$ which only depends on $\delta_c$ (the traffic slackness constant)
    , $H$ (the threshold for activating the back-pressure), $N$ (the number of queues), and $L_0$ and $L_1$ (the constants in Assumption \ref{assumption:smooth-policy}).
\end{restatable}

The proof of Lemma~\ref{lemma: smoothness} is in Appendix~\ref{section: lemma: smoothness: proof}.
The smoothness of $J(\pi_{H, \vtheta})$ ensures that 
when the policy perturbation $\Delta\vtheta=\vtheta' - \vtheta$ is small, the first-order term $\nabla_\vtheta J(\pi_{H,\vtheta})^\top \Delta\vtheta$ dominates the performance difference over higher-order terms. 
Since $\nabla_\vtheta J(\pi_{H,\vtheta})^\top \Delta\vtheta$ and $\nabla_\vtheta J(\pi_{H,\vtheta})^\top (-\Delta\vtheta)$ have opposite signs, half of the perturbation directions that are not orthogonal to the gradient make the first-order term negative. Therefore, 
there is a positive probability that a randomly perturbed policy improves the performance, as stated in the following theorem. 

\begin{restatable}{theorem}{periterationthm} \label{thm:positive_prob_per_iteration}
    At the $k$-th iteration of Algorithm~\ref{alg:zeroth-order}, if the gradient norm satisfies $\|\nabla_\vtheta J(\pi_{H, \vtheta_k})\|_2 \geq 2 \beta L_J d$ where $\beta$ is the perturbation distance, the perturbed policy improves by at least $\beta^2L_Jd$ with probability at least $1/32,$ i.e., 
    \begin{align*}
        \prob \left( J\left(\pi_{H,\vtheta_k'}\right) \leq J\left(\pi_{H,\vtheta_k}\right) - \beta^2 L_J d \right) \geq \frac{1}{32}.
    \end{align*}
\end{restatable}
Based on the theorem above,  the learned policy under Algorithm~\ref{alg:zeroth-order} does not become worse and improves with a constant probability when the gradient norm is  lower bounded. The next theorem establishes the convergence of the learned policy to an $\epsilon$-stationary point. 

\begin{restatable}{theorem}{mainthm} \label{thm: main_thm}
    Consider Algorithm~\ref{alg:zeroth-order} with the perturbation distance and learning rate being $\alpha = \beta = \frac{\varepsilon}{2 L_J d}$.  Given $\omega \in (0,1)$, if we pick a learned policy from $\{\vtheta_0, \vtheta_1, \cdots, \vtheta_{K-1}\}$ after $K$ policy iterations, uniformly  at random, where \begin{align*}
        K \geq  \frac{128 L_J d}{\omega \varepsilon^2} [J(\pi_{H,\vtheta_0}) - J^*],
    \end{align*} 
    then with probability at least $1-\omega$, the selected policy is an $\epsilon$-stationary point, i.e., the selected policy $\pi_{H, \vtheta_\rr}$ satisfies $\left\| \nabla_\vtheta J(\pi_{H,\vtheta_\rr}) \right\|_2 \leq \varepsilon$. Moreover, the last learned policy, parameterized by $\vtheta_K,$ satisfies $J(\pi_{H,\vtheta_K}) \leq J(\pi_{H,\vtheta_\rr})$ almost surely.
\end{restatable}

The proofs of Theorem~\ref{thm:positive_prob_per_iteration} and Theorem~\ref{thm: main_thm} are included in Appendix~\ref{section: thm 1 proof} and~\ref{section: thm 2 proof}.

For Algorithm~\ref{alg:zeroth-order-partition},following a similar argument as the proof of Theorem~\ref{thm:positive_prob_per_iteration},  we can derive a partition-wise performance improvement guarantee, i.e., at each iteration $k$, for each partition $j = 1, 2, \dots, D$, $\sP\left( J(\pi_{H,\vtheta_{k,j}'}) < J(\pi_{H,\vtheta_k}) \right)$ is lower bounded.
However, the partition-wise performance improvement does not guarantee an overall performance improvement.
Thus, we follow an approrach similar to the proof of Theorem 1 in~\cite{deyi2026_fl} to derive the convergence result of Algorithm~\ref{alg:zeroth-order-partition}.
\begin{restatable}{theorem}{partitionthm} \label{thm: partition}
    Consider Algorithm~\ref{alg:zeroth-order-partition} with the  perturbation distance and learning rate being $\alpha = \beta = \sqrt{\frac{J(\pi_{H,\vtheta_0}) - J^*}{L_J d K}}$.     
  If we pick a policy from $\{\vtheta_0, \vtheta_1, \cdots, \vtheta_{K-1}\}$ uniformly at random, then the selected policy $\pi_{H, \vtheta_\rr}$ satisfies
    \[
    \E\left[ \| \nabla_\vtheta J(\pi_{H,\vtheta_\rr}) \|_2 \right] \le 4 \sqrt{3} \sqrt{\frac{\left( J(\pi_{H,\vtheta_0}) - J^* \right) L_J d}{K}}.
    \]
\end{restatable}

The proof of Theorem~\ref{thm: partition} is included in Appendix~\ref{section: thm 3 proof}.

\subsection{Proof Roadmaps}

In this section, we show the proof roadmaps of Lemma~\ref{lemma: smoothness}, Theorem~\ref{thm:positive_prob_per_iteration}, Theorem~\ref{thm: main_thm} and Theorem~\ref{thm: partition}.

\subsubsection{Proof Roadmap of Lemma~\ref{lemma: smoothness}}
Considering two policies $\pi = \pi_{H, \vtheta}$ and $\pi' = \pi_{H, \vtheta'}$, we first use the performance difference lemma (Lemma A.9,~\cite{murthy2024}) and policy gradient theorem (Theorem 1,~\cite{Sutton1999}) to obtain
\begin{align}
    &J(\pi') - J(\pi) - \nabla_\vtheta J(\pi)^\top (\vtheta' - \vtheta) \nonumber \\
    =& \underbrace{ \sum_{\vq: \|\vq\|_1 \le H} \nu^{\pi} (\vq) \sum_{\vu \in \gU} (\pi' (\vu | \vq) - \pi (\vu | \vq) - \nabla_\vtheta \pi(\vu | \vq)^\top (\vtheta' - \vtheta) ) A^\pi (\vq, \vu) }_{\mathsf{I_1}} \nonumber \\
    &+ \underbrace{ \sum_{\vq: \|\vq\|_1 \le H} \left( \nu^{\pi'} (\vq) - \nu^{\pi} (\vq) \right) \sum_{\vu \in \gU} (\pi' (\vu | \vq) - \pi (\vu | \vq) ) A^\pi (\vq, \vu) }_{\mathsf{I_2}}.
    \label{eq: mainbody_smoothness_decom}
\end{align}
For term $\mathsf{I_1}$, we will show
\begin{align} \label{eq: mainbody_I_1}
    \left| \mathsf{I_1} \right| \le \max_{\substack{\vq: \|\vq\|_1 \le H \\ \vu \in \gU}} \left| A^\pi (\vq, \vu) \right| \frac{L_1}{2} \|\vtheta' - \vtheta\|_2^2,
\end{align} 
based on the smoothness of the policy class assumed in Assumption~\ref{assumption:smooth-policy}.
For term $\mathsf{I_2}$, we can derive:
\begin{align} 
    |\mathsf{I_2}| \le \max_{\substack{\vq: \|\vq\|_1 \le H \\ \vu \in \gU}} \left| A^\pi (\vq, \vu) \right| 
    L_0 \|\vtheta' - \vtheta\|_2 
    \|\vnu^{\pi'} - \vnu^\pi\|_1 ,
    \label{eq: mainbody_I_2_pre}
\end{align}
based on the Lipschitz continuity of the policy class assumed in Assumption~\ref{assumption:smooth-policy}.
We will further bound the advantage function and the stationary distribution difference such that
\[
    \max_{\substack{\vq: \|\vq\|_1 \le H \\ \vu \in \gU}} \left| A^\pi (\vq, \vu) \right|
    \quad \text{and} \quad
    \|\vnu^{\pi'} - \vnu^\pi\|_1.
\]
To establish these bounds, we first show uniform geometric mixing of the Markov chains under any policy in the policy class $\Pi_H$ (Lemma~\ref{lemma: Uniform geometric mixing}). 
Formally speaking, there exist a Lyapunov function $W:\gQ\mapsto[1,\infty)$, constants $C_\mix$ and $\rho_\mix \in (0,1)$ such that for all policies $\pi \in \Pi_H$,
\begin{align} \label{eq: geomix_main}
\left\| \left( \mP^\pi \right)^t (\vq, \cdot) - \vnu^\pi \right\|_{\rm TV} \le C_\mix W(\vq) \rho_\mix^t,
\qquad
\forall \vq \in \gQ, t\ge0,
\end{align}
where $\|\cdot\|_{\rm TV}$ is the total variance distance, and $\left( \mP^\pi \right)^t (\vq, \cdot)$ represents the $t$-step transition distribution starting from the state $\vq$ under policy $\pi$. This will be established by considering an exponential Lyapunov function $W(\vq) = \exp(\kappa \|\vq\|_2)$ with a policy-independent constant $\kappa >0$ and  by showing that the Lyapunov function satisfies the geometric drift condition, i.e., for all policy $\pi \in \Pi_H$, there exist constants $\gamma \in (0,1)$, $B, H_{\delta,N} < \infty$ such that
\[
    \E_\pi\left[
        \left. W(\vq(t+1))
        \right| \vq(t)=\vq
    \right]
    \le
    \gamma W(\vq) + B \mathbbm{1} \left\{ \| \vq \|_1 \le H_{\delta, N}\right\},
    \qquad
    \forall \vq\in\gQ.
\]
The details can be found in Lemma~\ref{geometrical_drift}.

With the uniform geometric mixing property, Lemma~\ref{lemma: advantage_bound} characterizes the policy-independent bound of $\max_{\substack{\vq: \|\vq\|_1 \le H \\ \vu \in \gU}} \left| A^\pi (\vq, \vu) \right|$ and Lemma~\ref{lemma: stationary distribution difference} characterizes the bound of $\|\vnu^{\pi'} - \vnu^\pi\|_1$.
Formally, we have for any policies $\pi$ and $\pi' \in \Pi_H$,
\[
\max_{\substack{\vq: \|\vq\|_1 \le H \\ \vu \in \gU}} \left| A^\pi (\vq, \vu) \right| 
\le 
2 C_{\mix} (H+N) \exp(\kappa (H+N)) \frac{1+\rho_\mix}{\rho_\mix (1 - \rho_\mix)^2},
\]
and
\[
\| \vnu^{\pi'} - \vnu^\pi \|_1 \le \frac{2 C_{\mix} \exp(\kappa (H+N))}{1 - \rho_{\mix}}  L_0 \| \vtheta'-\vtheta \|_2.
\]
Substituting these two bounds into~\eqref{eq: mainbody_I_1} and ~\eqref{eq: mainbody_I_2_pre}, and then into~\eqref{eq: mainbody_smoothness_decom} gives that
\begin{align} \label{eq: mainbody_smoothness_final}
    \left| J(\pi') - J(\pi) - \nabla_\vtheta J(\pi)^\top (\vtheta' - \vtheta) \right| &\le |\mathsf I_1| + |\mathsf I_2| \nonumber \\
    &\le \max_{\substack{\vq: \|\vq\|_1 \le H \\ \vu \in \gU}} \left| A^\pi (\vq, \vu) \right| \left( \frac{L_1}{2} + L_0^2 \frac{2 C_{\mix} \exp(\kappa (H+N))}{1 - \rho_{\mix}} \right) \| \vtheta'-\vtheta \|_2^2 \nonumber \\
    &= \frac{L_J}{2} \|\vtheta' - \vtheta\|_2^2,
\end{align}
where
\[
    L_J = 4 C_\mix (H+N) \exp(\kappa(H+N)) \frac{1+\rho_\mix}{\rho_\mix (1-\rho_\mix)^2} 
    \left( \frac{L_1}{2} + \frac{2 C_\mix \exp(\kappa(H+N))}{1-\rho_\mix} L_0^2 \right),
\] which completes the proof.

\subsubsection{Proof Roadmap of Theorem~\ref{thm:positive_prob_per_iteration}}

At iteration $k,$ let $\vtheta_k' = \vtheta_k + \beta \vv_k$. Since $J(\pi_{H,\vtheta})$ is $L_J$-smooth according to Lemma~\ref{lemma: smoothness}, we have,
\begin{align}
   \beta \nabla_\vtheta J(\pi_{H,\vtheta_k})^\top \vv_k - \frac{L_J}{2} \beta^2 \|\vv_k\|_2^2\le J(\pi_{H,\vtheta_k'}) - J(\pi_{H,\vtheta_k})  \le \beta \nabla_\vtheta J(\pi_{H,\vtheta_k})^\top \vv_k +\frac{L_J}{2} \beta^2 \|\vv_k\|_2^2.
\end{align}
Since the perturbation of each coordinate is sampled according to a Rademacher distribution,  $\|\vv_k\|_2^2 = \sum_{i = 1}^d v_{k,i}^2 = d$ and a sufficient condition that the policy performance improves by at least $\beta^2 L_J d$ 
is 
\begin{align} 
    \nabla_\vtheta J(\pi_{H,\vtheta_k})^\top \vv_k < -\frac{L_J \beta}{2} \|\vv_k\|_2^2 - \beta L_J d = -\frac{L_J \beta}{2} d - \beta L_J d = -\frac{3 \beta L_J d}{2}.
\end{align}
 Fix $\vtheta_k$,  notice that the distribution of $\nabla_\vtheta J(\pi_{H,\vtheta_k})^\top \vv_k$ is symmetric around $0$, so
\begin{align}     
\sP \left( \nabla_\vtheta J(\pi_{H,\vtheta_k})^\top \vv_k < -\frac{3 \beta L_J d}{2} \right) = 
    \frac{1}{2} \sP \left( \left( \nabla_\vtheta J(\pi_{H,\vtheta_k})^\top \vv_k \right)^2 > \frac{9 \beta^2 L_J^2 d^2}{4} \right).
\end{align}

To lower bound $\sP\left( \left(\nabla_\vtheta J(\pi_{H,\vtheta_k})^\top \vv_k\right)^2 >  \frac{9 \beta^2 L_J^2 d^2}{4} \right)$, we apply Paley–Zygmund inequality (Lemma 19,~\cite{Paley_Zygmund_1932}):
given $\rz \ge 0$ be a random variable with $\E[\rz^2] < \infty$, and let $\iota \in [0,1]$, then
\[
\sP(\rz > \iota\E[\rz]) \ge (1 - \iota)^2 \frac{\E[\rz]^2}{\E[\rz^2]}.
\]
Considering $\left(\nabla_\vtheta J(\pi_{H,\vtheta_k})^\top \vv_k\right)^2$ as the random variable $\rz$ in Paley–Zygmund inequality, we will show that
\begin{align*}
    \E \left[ \left( \nabla_\vtheta J(\pi_{H,\vtheta_k})^\top \vv_k \right)^2 \right] &= \| \nabla_\vtheta J(\pi_{H,\vtheta_k}) \|_2^2 \ge 4 \beta^2 L_J^2 d^2 \nonumber \\
    \E \left[ \left( \nabla_\vtheta J(\pi_{H,\vtheta_k})^\top \vv_k \right)^4 \right] &\leq 3 \| \nabla_\vtheta J(\pi_{H,\vtheta_k}) \|_2^4.
\end{align*}
Substituting these into Paley–Zygmund inequality and applying $\iota = 9/16$ gives the probability lower bound $1/32$.

\subsection{Proof Roadmap of Theorem~\ref{thm: main_thm}}

At iteration $k$, when $\left\|\nabla_\vtheta J(\pi_{H,\vtheta_k})\right\|_2 \geq 2\beta L_Jd$, Theorem~\ref{thm:positive_prob_per_iteration} shows that, with probability at least $p=1/32$, the perturbed policy reduces the average cost by at least $\beta^2L_Jd$.
Since Algorithm~\ref{alg:zeroth-order} only accepts a perturbed policy when it has a smaller average cost, we obtain
\[
    \E\left[
        J(\pi_{H,\vtheta_{k+1}})
        -
        J(\pi_{H,\vtheta_k})
    \right]
    \leq
    -p\beta^2L_Jd\,
    \prob\left(
        \left\|\nabla_\vtheta J(\pi_{H,\vtheta_k})\right\|_2
        \geq 2\beta L_Jd
    \right).
\]
Taking a telescoping sum over $k=0,1,\ldots,K-1$, and using $J(\pi_{H,\vtheta_K})\geq J^*$, we obtain
\[
    \frac{1}{K}
    \sum_{k=0}^{K-1}
    \prob\left(
        \left\|\nabla_\vtheta J(\pi_{H,\vtheta_k})\right\|_2
        \geq 2\beta L_Jd
    \right)
    \leq
    \frac{
        J(\pi_{H,\vtheta_0})-J^*
    }{
        p\beta^2L_JdK
    }.
\]
Since index $r$ is chosen from
$\{0, \cdots, K-1\}$ uniformly at random, each learned policy is selected with probably $1/K$, which implies that 
\[
    \prob\left(
        \left\|\nabla_\vtheta J(\pi_{H,\vtheta_\rr})\right\|_2
        \geq 2\beta L_Jd
    \right) = \frac{1}{K}
    \sum_{k=0}^{K-1}
    \prob\left(
        \left\|\nabla_\vtheta J(\pi_{H,\vtheta_\rr})\right\|_2
        \geq 2\beta L_Jd | \vtheta_\rr=\vtheta_k\right)
    \leq
    \frac{
        J(\pi_{H,\vtheta_0})-J^*
    }{
        p\beta^2L_JdK
    }.
\]

Finally, choosing $\alpha=\beta=\varepsilon/2L_Jd$ and substituting them and $p=1/32$ into the probability bound yield that
\[
    K
    \geq
    \frac{128 L_Jd}{\omega\varepsilon^2}
    \left[
        J(\pi_{H,\vtheta_0})-J^*
    \right]
\]
is sufficient to guarantee
\[
    \prob\left(
        \left\|\nabla_\vtheta J(\pi_{H,\vtheta_\rr})\right\|_2
        \leq\varepsilon
    \right)
    \geq 1-\omega.
\]
Finally, since Algorithm~\ref{alg:zeroth-order} only accepts policy updates that decrease the average cost, $J(\pi_{H,\vtheta_k})$ is non-increasing along every sample path and
\[
    J(\pi_{H,\vtheta_K})
    \leq
    J(\pi_{H,\vtheta_\rr})
\]
almost surely.

\subsection{Proof Roadmap of Theorem~\ref{thm: partition}}

Given the smoothness shown in Lemma~\ref{lemma: smoothness} and linearity of the first-order term, the per step performance difference $J(\pi_{H,\vtheta_{k+1}}) - J(\pi_{H,\vtheta_k})$ can be bounded as follows:
\begin{align}
    J(\pi_{H,\vtheta_{k+1}}) - J(\pi_{H,\vtheta_k})
    \le \alpha \sum_{j=1}^D \left\langle \nabla_\vtheta J(\pi_{H,\vtheta_k}), \mathbbm{1}\left\{ J(\pi_{H,\vtheta_{k,j}'}) < J(\pi_{H,\vtheta_k})\right\} \vv_k \circ \ve_{\gD_j} \right\rangle + \frac{\alpha^2 L_J d}{2}.
    \label{eq: proof_roadmap_perstep_partition}
\end{align}
For partially perturbed policy, 
we can further upper bound the expectation of the first-order term in~\eqref{eq: proof_roadmap_perstep_partition} by following the proof of Lemma 2 in~\cite{deyi2026_fl} and obtain
\begin{align*}
    \E \left[\left. \left\langle \nabla_\vtheta J(\pi_{H,\vtheta_k}), \mathbbm{1}\left\{ J(\pi_{H,\vtheta_{k,j}'}) < J(\pi_{H,\vtheta_k})\right\} \vv_k \circ \ve_{\gD_j} \right\rangle \right| \vtheta_k \right] \le -\frac{\| \nabla_\vtheta J(\pi_{H,\vtheta_k}) \circ \ve_{\gD_j} \|_2}{2 \sqrt{3}} + \frac{\beta L_J |\gD_j|}{2}.
\end{align*}
With this bound, applying the same telescoping sum as in the proof of Theorem~\ref{thm: main_thm} gives that
\begin{align*}
    \frac{\sum_{k=0}^{K-1} \E\left[ \sum_{j=1}^D \| \nabla_\vtheta J(\pi_{H,\vtheta_k}) \circ \ve_{\gD_j} \|_2 \right]}{K} \le \frac{2 \sqrt{3} \left( J(\pi_{H,\vtheta_0}) - J^* \right)}{\alpha K} + 2 \sqrt{3} \alpha L_J d.
\end{align*}
Since $\sum_{j=1}^D \| \vv \circ \ve_{\gD_j} \|_2 \ge \|\vv\|_2$ (Proposition 2 in~\cite{deyi2026_fl}), we can obtain 
\begin{align*}
    \E\left[\| \nabla_\vtheta J(\pi_{H,\vtheta_\rr}) \|_2\right] \le \frac{2 \sqrt{3} \left( J(\pi_{H,\vtheta_0}) - J^* \right)}{\alpha K} + 2 \sqrt{3} \alpha L_J d.
\end{align*}
The proof completes by choosing a proper learning rate $\alpha$.

%% file: sections/experiments.tex
\section{Experiments} \label{section: experiments}

This section presents our experimental results of Algorithm~\ref{alg:zeroth-order} and Algorithm~\ref{alg:zeroth-order-partition} in different queueing networks, and comparisons with other benchmark policies.

\subsection{Queueing Networks and Simulation platform}
Our experiments include three types of queueing networks (criss-cross network, N-model network, and nine-layer Exp Reentrant network) that can be formulated as Markov Decision Processes with state space being all possible queue lengths, and one queueing network (five-layer HyperExp Reentrant network) that \textit{cannot} be formulated as Markov Decision Processes based on queue lengths alone. 
We simulated these queueing networks in QGym~\citep{QGym}. 

\begin{figure}[htbp]
    \centering
    \begin{subfigure}[b]{0.38\textwidth}
        \centering
        \includegraphics[width=\linewidth]{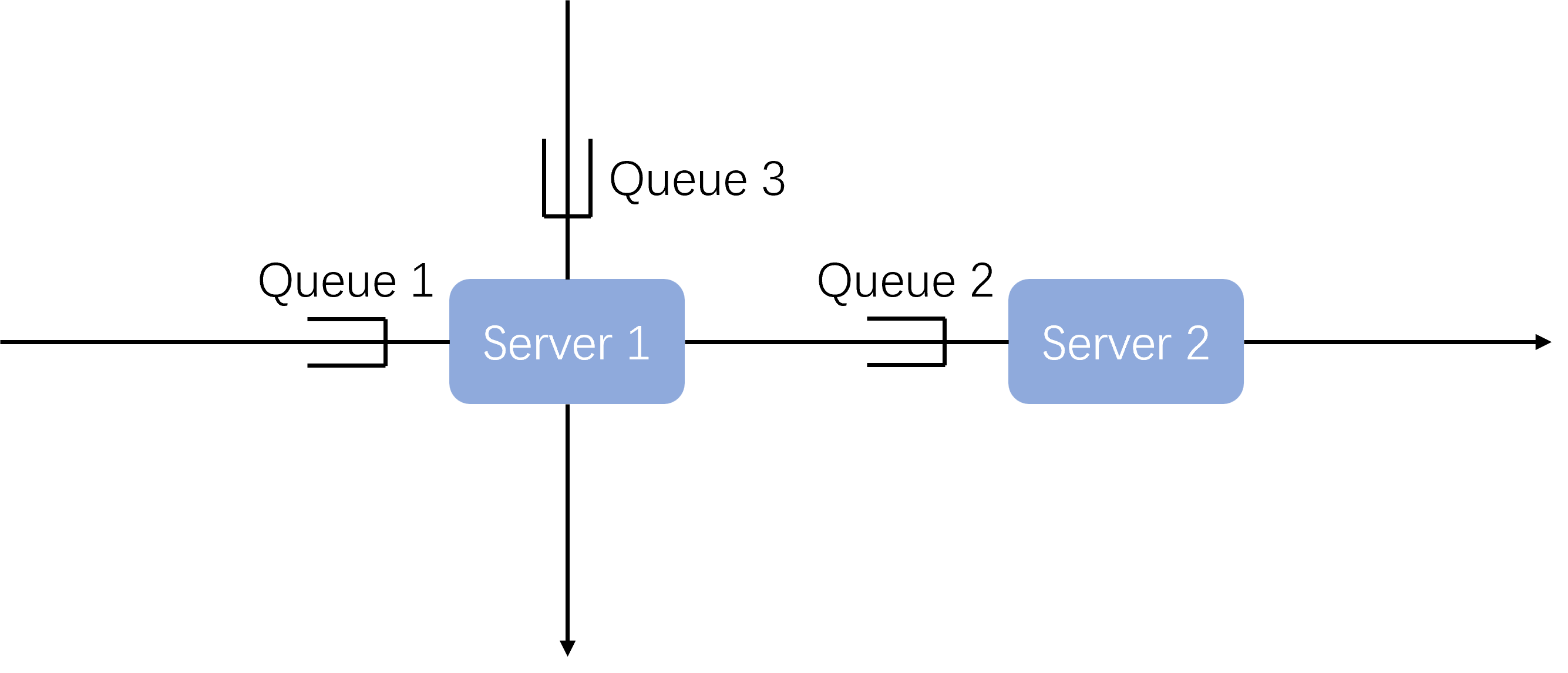}
        \caption{Criss-cross network}
        \label{fig:criss-cross-experiment}
    \end{subfigure}
    \hfill
    \begin{subfigure}[b]{0.6\textwidth}
        \centering
        \includegraphics[width=\linewidth]{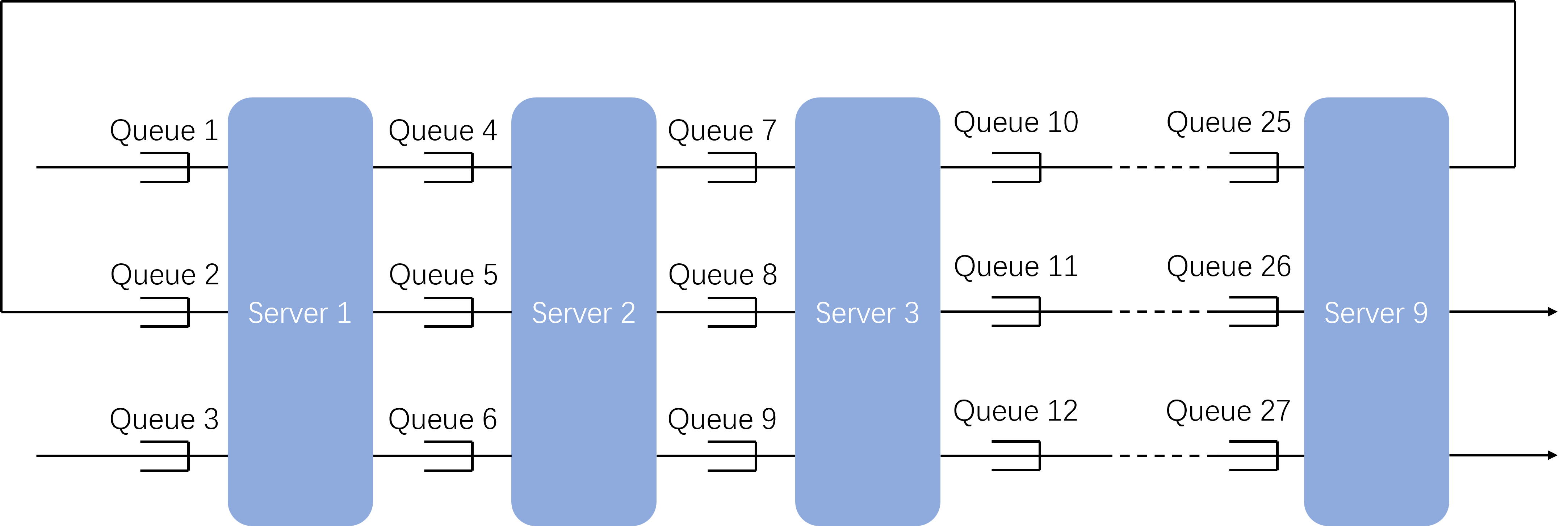}
        \caption{Nine-layer Reentrant network}
        \label{fig:9_reentrant_experiment}
    \end{subfigure}
    \label{fig:main-cluster}    
    \begin{subfigure}[b]{0.38\textwidth}
        \centering
        \includegraphics[width=\linewidth]{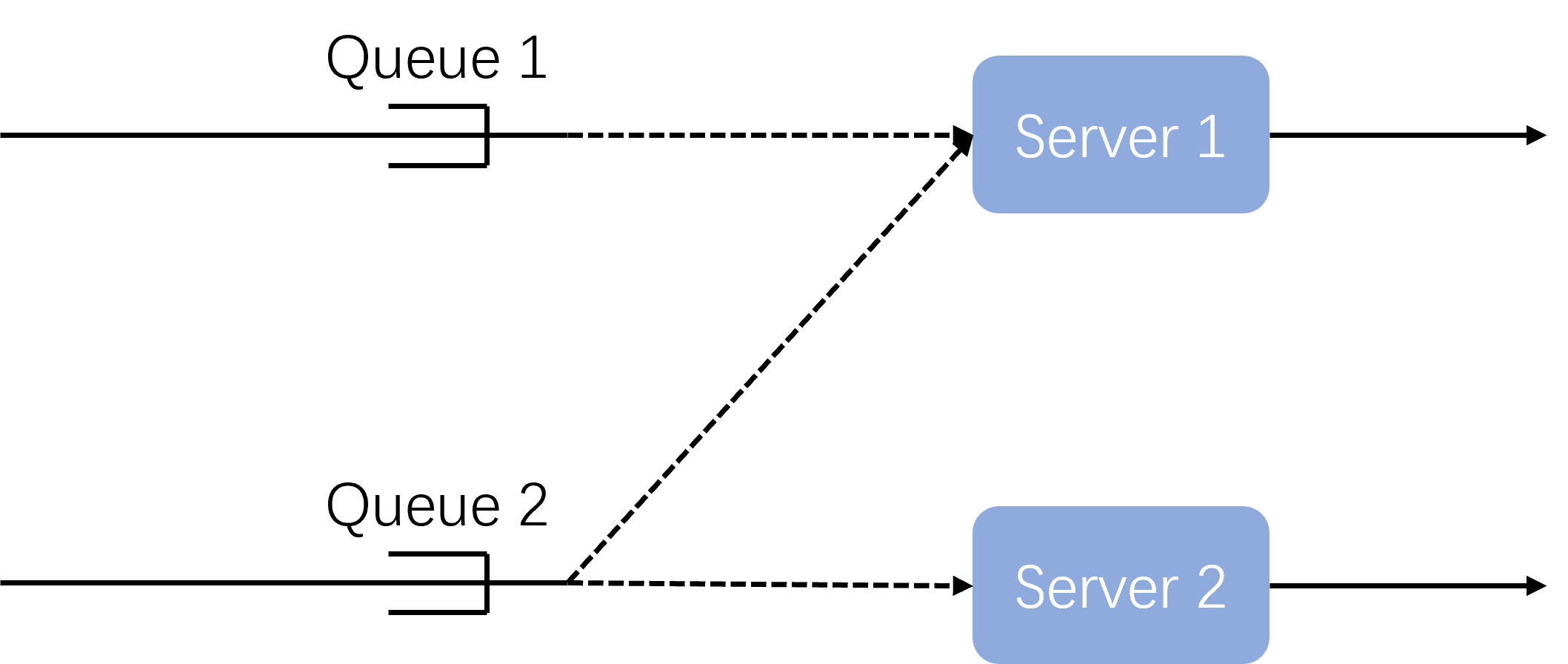}
        \caption{N-model network}
        \label{fig:n-model-experiment}
    \end{subfigure}
    \begin{subfigure}[b]{0.6\textwidth}
        \centering
        \includegraphics[width=\linewidth]{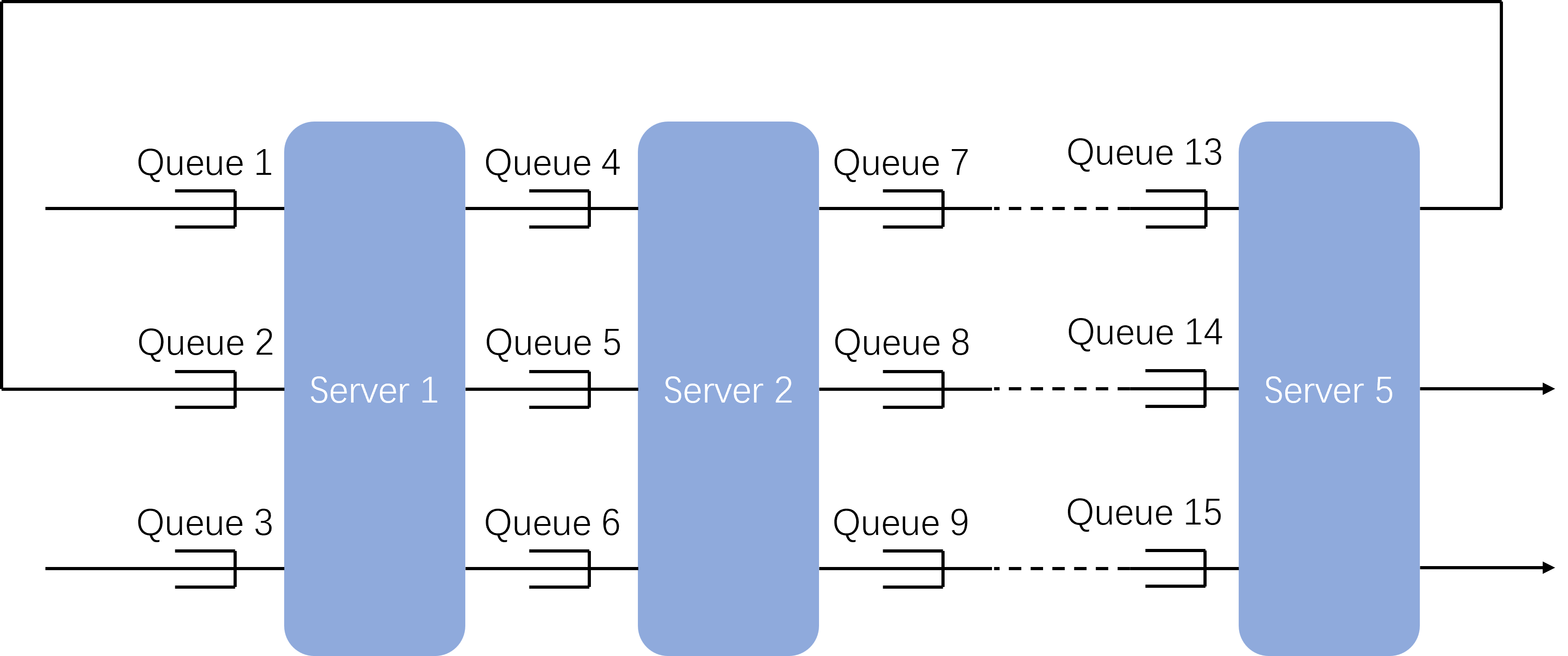}
        \caption{Five-layer Reentrant Network}
        \label{fig:5_reentrant_experiment}
    \end{subfigure}
    \caption{Topologies of queueing networks.}
\end{figure}

\textbf{Criss-cross network (Figure~\ref{fig:criss-cross-experiment}):}
There are $N = 3$ queues and $M = 2$ servers in the network.
The exogenous arrival rate is $\vlambda = [0.9, 0, 0.9]^\top$,
and the service rate is $\vmu = [2, 1, 2]^\top$.
The holding cost $c(\vq) = \sum_{i=1}^3 q_i$.
The service of any server is preemptive, i.e., the server can pause its service to one queue and switch to serve another queue from one time slot to another.
The load of each server in the system is $\varrho=0.9$.

\textbf{N-model network (Figure~\ref{fig:n-model-experiment}):}
There are $N=2$ queues and $M=2$ servers in the network.
The N-model network does not satisfy the assumption  we made in Section~\ref{section: problem formualtion} that each queue can only be served by a unique server but can be still formulated an MDP problem. 
In the N-model network, $\gN_1 = \{1,2\}$ and $\gN_2 = \{2\}$.
Both queues have exogenous Poisson arrivals with rates $\vlambda = [0.38, 1.235]^\top$. The service times follow exponential distributions defined below:
Let $\mu_{i,m}$ denote the service rate of a job in Queue $i$ being served by Server $m.$
We set $\mu_{1,1} = 1$, $\mu_{2,1} = 0.5$ and $\mu_{2,2} = 1$.
If both Server $1$ and $2$ are scheduled to serve Queue $2$, Server 2 gets the job if there is only one job in Queue $2.$
The holding cost is $c(\vq) = q_1 + 3q_2$.

\textbf{Nine-layer Exp reentrant network (Figure~\ref{fig:9_reentrant_experiment}):}
There are $N = 27$ queues and $M = 9$ servers in the network.
Queue 1 and Queue 3 have exogenous Poisson arrivals with arrival rate $\lambda_1 = \lambda_3 = 9/140$.
Servers with odd indices have the same service rate, and servers with even indices also have the same service rate. The service times follows an exponential distributions.
For $m = 1,3,5,7,9$, we have $\mu_{3m-2} = 1/8$, $\mu_{3m-1} = 1/2$, and $\mu_{3m} = 1/4$;
for $m = 2,4,6,8$, we have $\mu_{3m-2} = 1/6$, $\mu_{3m-1} = 1/7$, and $\mu_{3m} = 1$.
The holding cost $c(\vq) = \sum_{i=1}^{27} q_i$. The load of each server in the system is $\varrho=0.9$.

\textbf{Five-layer HyperExp reentrant network (Figure~\ref{fig:5_reentrant_experiment}):}
There are $N = 15$ queues and $M = 5$ servers in the network.
Queue 1 and Queue 3 have exogenous Poisson arrivals with arrival rate $\lambda_1 = \lambda_3 = 9/140$.
Service times follow hyper-exponential distributions, defined by the following desnity function $$f_\rx(x) = \sum_{i=1}^n p_i f_{\ry_i}(x),$$ where each $\ry_i$ is an exponentially distributed random variable, and $\{p_i\}_{i=1}^n$ a discrete probability distribution.
Specifically, in this queueing network, given a service rate $\mu$, the service time of each job is independently sampled from a hyper-exponential distribution with $n=2$, $p_1 = p_2 = 1/2$, $\ry_1 \sim \text{Exp} (\mu/1.8)$, and $\ry_2 \sim \text{Exp} (\mu/0.2)$.
This ensures the mean service time is still $1/\mu$.
Servers with odd indices have the same service rate, and servers with even indices also have the same service rate.
For $m = 1,3,5$, we have $\mu_{3m-2} = 1/8$, $\mu_{3m-1} = 1/2$, and $\mu_{3m} = 1/4$;
for $m = 2,4$, we have $\mu_{3m-2} = 1/6$, $\mu_{3m-1} = 1/7$, and $\mu_{3m} = 1$.
The holding cost $c(\vq) = \sum_{i=1}^{15} q_i$. The load of each server in the system is $\varrho=0.9$.

\subsection{Baselines}

We compare our algorithm to two traditional queueing policies: $c\mu$ policy and MaxWeight policy, and also PPO tailored for the queueing application~\citep{QGym}.

\textbf{$\boldsymbol{c\mu}$ policy.}
Given a queue state $\vq$, for each queue $i \in [N]$, let $w_{c\mu,i} = c_i \mu_i$ denote the weight, and $c\mu$ policy chooses the scheduling action $\vu_{c\mu}$ that maximizes $\sum_{i=1}^N w_{c\mu,i}u_i$.

\textbf{MaxWeight policy.}
Given queue state $\vq$, for each queue $i \in [N]$, define $w_{\text{MW},i} = c_i \mu_i q_i$ to be the weight of the queue, then the MaxWeight policy chooses the scheduling action $\vu_{\text{MW}}$ that maximizes $\sum_{i=1}^N w_{\text{MW},i}u_i$.

Note that for the N-model network, both servers can serve Queue 2. Thus, feasible scheduling actions are defined different from in Section~\ref{section: problem formualtion}, and c$\mu$ policy and MaxWeight policy will also be defined differently accordingly.
For the N-model network, we define the set of feasible scheduling actions as
$$\gU_\text{N-model} = \left\{ \vu \in \{ 0,1 \}^{2\times 2}: u_{1,2}=0,\ u_{1,1}+u_{2,1}\le1,\ u_{2,2}\le1 \right\},$$
where for $i=1,2$, $m=1,2$, $u_{i,m} \in \{0,1\}$ represents if server $m$ serves queue $i$.
We then define the scheduling actions under the c$\mu$ policy and MaxWeight policy as:
\begin{align*}
    \vu_{c\mu} &= \argmax_{\vu \in \gU_\text{N-model}} \sum_{i=1}^2 \sum_{m=1}^2 c_i \mu_{i,m} u_{i,m}, \\
    \vu_{\rm MW} &= \argmax_{\vu \in \gU_\text{N-model}} \sum_{i=1}^2 \sum_{m=1}^2 c_i q_i \mu_{i,m} u_{i,m}.
\end{align*}

We further remark that though the back-pressure policy is throughput optimal, $c\mu$ policy and MaxWeight policy outperform backpressure for the criss-cross network and the two reentrant networks as shown in \cite{QGym}, and for the N-model network, since there is no routing, the back-pressure policy is identical to the MaxWeight policy.
Therefore, we did not include back-pressure in this paper as a baseline. 

\textbf{PPO.}
PPO algorithm for the queueing system also aims to minimize the clipped objective:
\[
\min_{\vtheta}\ L(\vtheta) = \hat\E_t \left[ \max\left\{ \frac{\pi_{\vtheta} (\vu_t | \vq_t)}{\pi_{\vtheta_{old}} (\vu_t | \vq_t)} \hat A (\vq_t, \vu_t), \operatorname{clip}\left( \frac{\pi_{\vtheta} (\vu_t | \vq_t)}{\pi_{\vtheta_{old}} (\vu_t | \vq_t)}, 1-\epsilon, 1+\epsilon \right) \hat A (\vq_t, \vu_t) \right\} \right],
\]
where $\hat A$ is the estimated advantage function.
\cite{QGym} provided an open-source implementation of PPO algorithms over various queueing networks.

\subsection{Policy Parameterization} \label{section: exp-policy-parameterization}

For the PPO Algorithm and our proposed zeroth-order algorithms,
we use the default work-conserving policy (never assign a server to an empty queue) parameterization instead of the weakly work-conserving policies introduced in Section~\ref{section: policies}. For the queueing networks considered in this section, the optimal policy class includes a work-conserving policy. 
After pretraining, we observed the pre-trained policies are stable, so we did not set the back-pressure threshold $H$, which can be regarded as choosing a very large threshold.  
The neural network for policy parameterization contains one input layer, three hidden layers, and one output layer.
Each hidden layer equips tanh activation function.
For a queueing system with $N$ queues and $M$ servers, the size of each layer was chosen as follows: 
\[
N \xrightarrow{\rm tanh} 10N \xrightarrow{\rm tanh} \left\lfloor10\sqrt{NM}\right\rceil \xrightarrow{\rm tanh} 10M \rightarrow MN.
\]
The output is $\vp \in \sR^{M\times N}$.
Then for each server $m$, when $\sum_{i' \in \gN_m} q_{i'} > 0$, it will serve queue $i$ (if it belongs to server $m$) with probability
\[
\frac{ \exp(p_{m,i}) \mathbbm 1 \{q_i>0\} }{ \sum_{i' \in \gN_m} \exp(p_{m,i'}) \mathbbm 1 \{q_{i'}>0\} }.
\]
Note that for the PPO Algorithm, a value-function network used to construct advantage estimates is also needed with the same structure but different output layer.

\subsection{Training Process and Policy Evaluation}

\textbf{Pretrain.}
Prior to PPO training and zeroth-order training, the policy network is pretrained using behavioral cloning.
Specifically, we uniformly sample $10^6$ queue states from $\{0,1,\dots,100\}^N \in \gQ$, and for each sample $\vq$, we generate the corresponding weight $\vw_{\rm P}(\vq) \in \sR^{M\times N}$ such that
$$[w_{\rm P}(\vq)]_{m,i} = \frac{c_i \mu_{i} q_i}{\sum_{i' \in \gN_m} c_{i'} \mu_{i'} q_{i'}},$$
and $[w_{\rm P}(\vq)]_{m,i} = 0$ for $i,m$ such that $i \notin \gN_m$.
Note that when for some server $m$, $\sum_{i' \in \gN_m} q_{i'} = 0$, a uniform distribution over queues $i' \in \gN_m$ will be applied, i.e., for $i' \in \gN_m$, $[w_{\rm P}(\vq)]_{m,i'} = 1/|\gN_m|$.
For the N-model network, for $i,m$ such that $i \in \gN_m$, the target weight is
$$[w_P(\vq)]_{m,i} = \frac{c_i \mu_{i,m} q_i}{\sum_{i' \in \gN_m} c_{i'} \mu_{i',m} q_i}, $$
and for other cases, the target weight is defined the same as before.
Then we minimize the MSE between the softmax of the neural network output logits and $\vw_{\rm P}(\vq)$.
We use Adam to pretrain the policy network. The batch size is $100$, the training epoch is $1$, and the learning rate is $3\times10^{-4}$.

\textbf{Algorithm~\ref{alg:zeroth-order} training.}
The total number of policy iterations during training is $K=100$.
Let $\theta_{0,\max}$ denote the maximum absolute value among all parameters of the initial policy network (after pretraining).
We apply time-varying perturbation distance and learning rate.
Let $\{\beta_k\}_{k=0}^{99}$ denote the perturbation scheduler, and $\{\alpha_k\}_{k=0}^{99}$ denote the learning rate scheduler.
The initial perturbation distance is $\beta_0 = 0.5 \theta_{0,\max}$, and it decreases logarithmically to $\beta_{99} = 0.1 \theta_{0,\max}$ along training.
At the $k$-th iteration, the learning rate $\alpha_k = 0.2 \beta_k$.
Note that in experiments, the learning rate is chosen to be smaller than the the perturbation distance to account for the imperfect policy evaluation from finite samples.
In each training iteration for both the original and the perturbed policy, we use the mean of the time-average cost among $10$ trajectories to estimate the average cost $J(\pi)$, and update the policy accordingly. The trajectory length varies based on the queueing network environment and is shown in Table~\ref{tab:fair_table}.

\textbf{Algorithm~\ref{alg:zeroth-order-partition} training.}
The total number of policy iterations during training is $K=100$.
The parameter space is partitioned into $D=8$ subsets, and each subset includes the weights and biases of one neural network layer.
For $j = 1,2,\dots,D$, let $\theta_{0,\max,j}$ denote the maximum absolute value among all parameters in the $j$-th partition of the initial policy network (after pretraining).
For each partition, we apply the same perturbation distance and learning rate scheduler as in Algorithm~\ref{alg:zeroth-order} using $\theta_{0,\max,j}$. To estimate the average cost $J(\pi)$ for the original policy and all $8$ partition-perturbed policies, we use the mean of time-averaged cost from $10$ trajectories, where the lengths for different environments are summsraized in Table~\ref{tab:fair_table}. The lengths are chosen so that the total numbers of simulated rollout steps approximately equal across the two algorithms.

\textbf{PPO training.}
The total number of training iterations is $K=100$.
We inherit all training settings from~\cite{QGym}, except that in each training iteration, we collect $50$ trajectories. The length of the trajectory depends on the queueing network and is shown in Table~\ref{tab:fair_table}.

\begin{table}[h]
\centering
\small
\begin{tabular}{cccc}
\toprule
Queueing Networks & Algorithm~\ref{alg:zeroth-order} & Algorithm~\ref{alg:zeroth-order-partition} & PPO  \\
\midrule
Criss-cross 
& $10000$
& $2200$
& $4000$ \\

N-model
& $4500$ 
& $1000$ 
& $1800$ \\

9-layer Exp Reentrant
& $135000$
& $30000$ 
& $54000$ \\

5-layer HyperExp Reentrant
& $121500$
& $27000$ 
& $50000$ \\

\bottomrule
\end{tabular}
\caption{The length (number of steps) of each collected trajectory in RL training. The trajectory lengths are chosen to make the sample complexities almost the same across algorithms.}
\label{tab:fair_table}
\end{table}

\textbf{Policy evaluation.}
For each policy, we collect $400$ long trajectories, obtain their time-averaged holding costs, and use the mean to evaluate the average holding cost and report the $95\%$ confidence interval.
The evaluation trajectory length of each queueing network is summarized in Table~\ref{tab: eval_traj}.
\begin{table}[h]
\begin{tabular}{cccccc}
\toprule
Queueing Networks & Criss-cross & N-model & 9-layer Exp Reentrant & 5-layer HyperExp Reentrant  \\
\midrule
Trajectory length 
& $10000$
& $10000$
& $300000$
& $150000$ \\
\bottomrule
\end{tabular}
\caption{The length (number of steps) of each evaluation trajectory.}
\label{tab: eval_traj}
\end{table}

\subsection{Results}
\begin{figure}[h]
    \centering
    \begin{subfigure}[b]{0.49\textwidth}
        \centering
        \includegraphics[width=\textwidth]{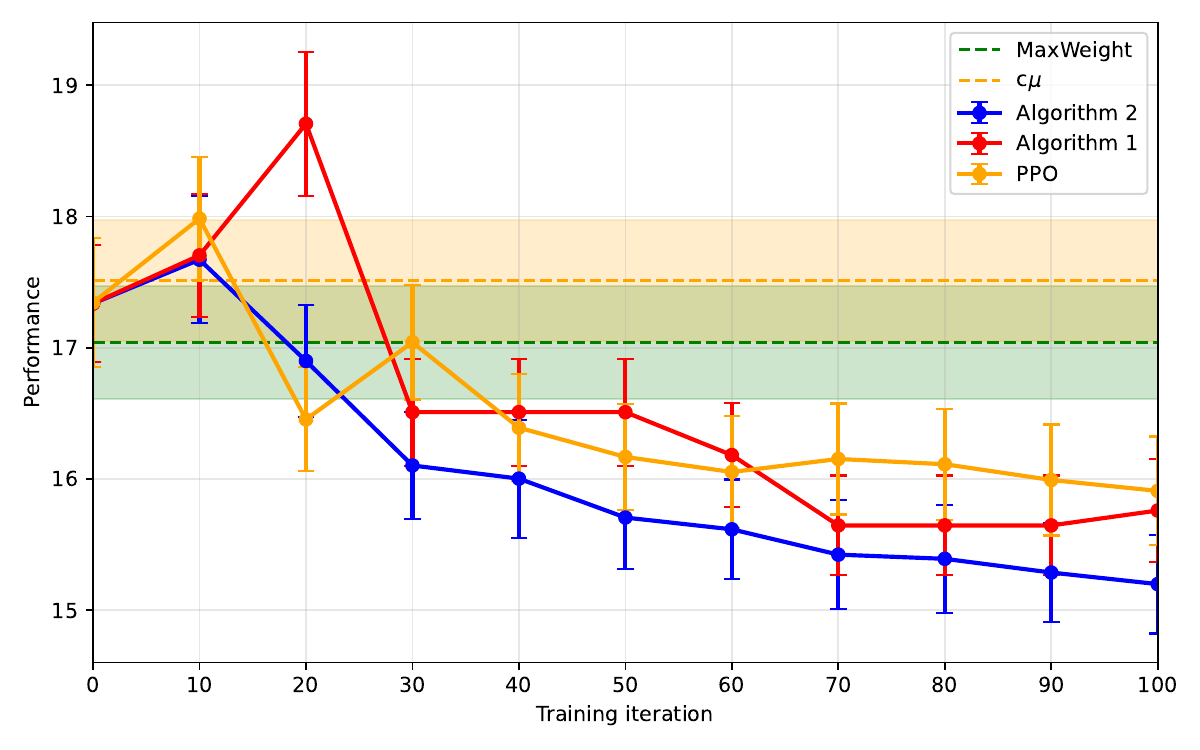}
        \caption{Criss-cross}
    \end{subfigure}
    \begin{subfigure}[b]{0.49\textwidth}
        \centering
        \includegraphics[width=\textwidth]{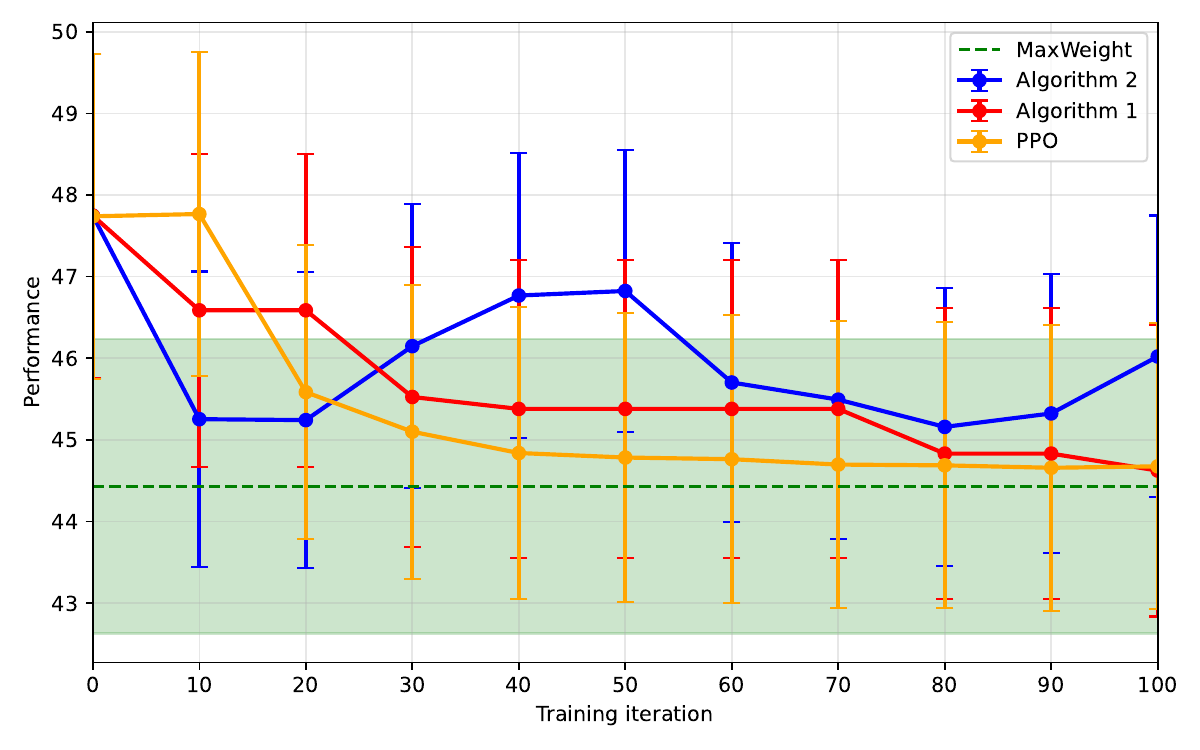}
        \caption{N-model}
    \end{subfigure}
    \begin{subfigure}[b]{0.49\textwidth}
        \centering
        \includegraphics[width=\textwidth]{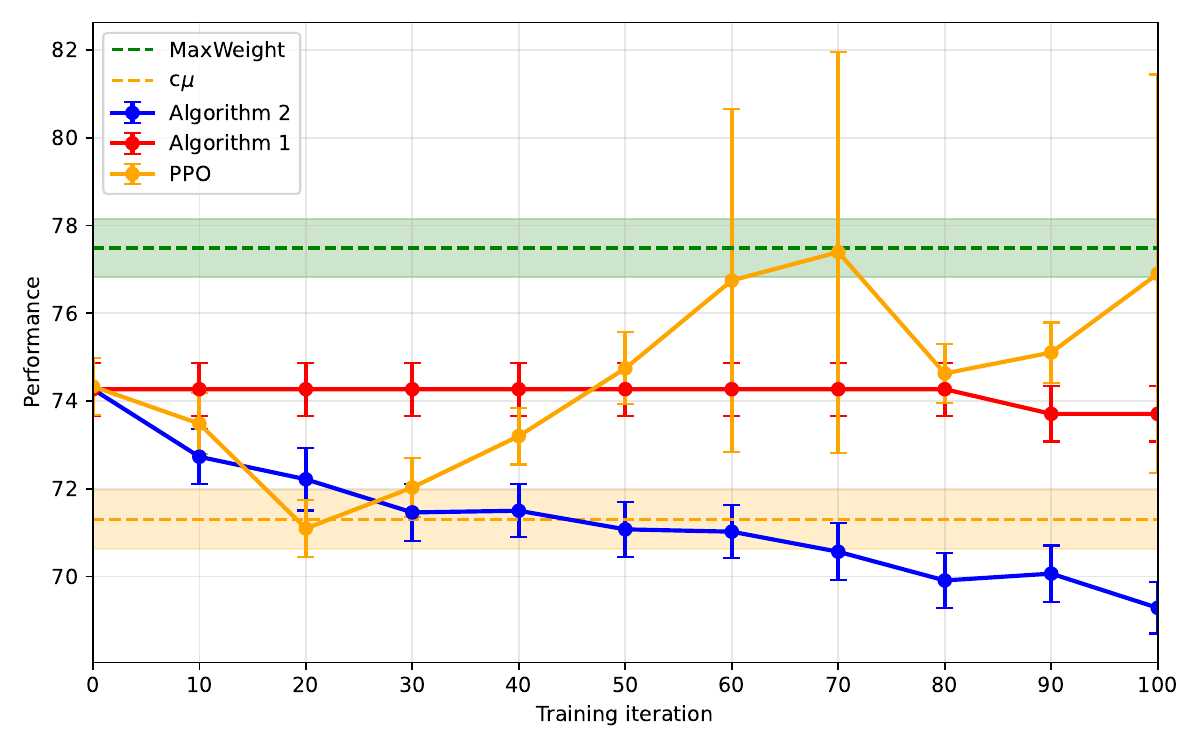}
        \caption{9-layer Exp Reentrant}
    \end{subfigure}
    \begin{subfigure}[b]{0.49\textwidth}
        \centering
        \includegraphics[width=\textwidth]{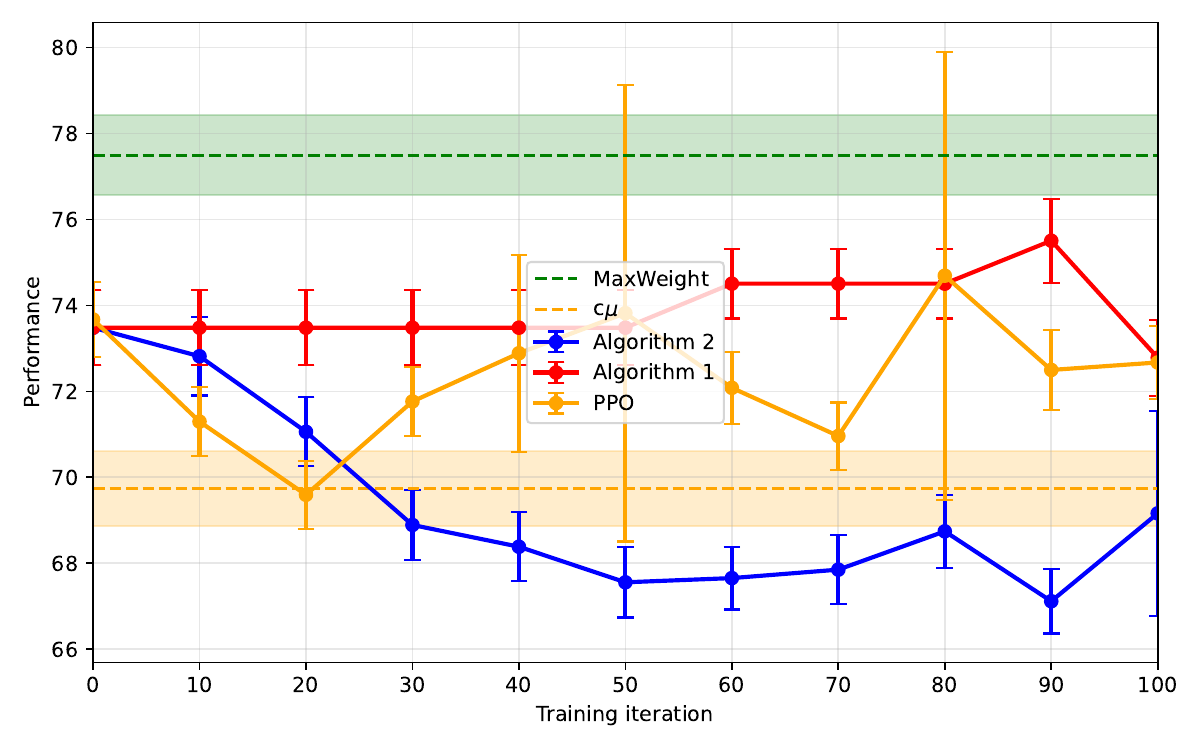}
        \caption{5-layer HyperExp Reentrant}
    \end{subfigure}
    \caption{Performances (average-costs) of Algorithm~\ref{alg:zeroth-order}, Algorithm~\ref{alg:zeroth-order-partition}, and PPO algorithm during training}
    \label{fig:final_results}
\end{figure}

For each queueing network, we evaluate the priority-based policies ($c\mu$ and/or MaxWeight), and train RL policies based on PPO, Algorithm~\ref{alg:zeroth-order}, and Algorithm~\ref{alg:zeroth-order-partition}.
For training algorithms, we evaluate the intermediate policy every $10$ training iterations.
The results are as shown in Figure~\ref{fig:final_results}.
Table~\ref{tab:performances} summarizes the performance of priority-based benchmark policies and the best RL training policies.

\begin{table}[h]
\centering
\small
\begin{tabular}{cccccc}
\toprule
Queueing Networks & $c\mu$ & MaxWeight & Algorithm~\ref{alg:zeroth-order} & Algorithm~\ref{alg:zeroth-order-partition} & PPO  \\
\midrule
Criss-cross 
& $17.51\pm0.46$
& $17.04\pm0.43$
& $15.62\pm0.38$
& $\mathbf{15.20\pm0.38}$
& $15.88\pm0.42$ \\

N-model
& $183.52\pm5.37$
& $\mathbf{44.43\pm1.80}$
& $44.62\pm1.79$
& $45.16\pm1.71$
& $44.66\pm1.75$\\

9-layer Exp Reentrant
& $71.30\pm0.67$
& $77.49\pm0.67$
& $73.70\pm0.63$
& $\mathbf{69.28\pm0.58}$ 
& $71.09\pm0.64$ \\

5-layer HyperExp Reentrant
& $69.73\pm0.87$
& $77.50\pm0.93$
& $72.78\pm0.88$
& $\mathbf{67.12\pm0.75}$ 
& $69.52\pm0.79$ \\

\bottomrule
\end{tabular}
\caption{Performances of priority-based benchmark policies and the best RL training policies.}
\label{tab:performances}
\end{table}

We observe that except for in the N-model network, Algorithm~\ref{alg:zeroth-order-partition} has the smallest average costs. For N-model, while MaxWeight has the smallest average cost,  Algorithm~\ref{alg:zeroth-order-partition} is the second best and is close to MaxWeight. 

\section{Conclusions}
This paper studied reinforcement learning for scheduling in queueing networks.
We proposed zeroth-order policy optimization algorithms that estimate only the average-cost, a scalar value, at each training iteration.
We established uniform smoothness of the back-pressure-guarded policy class under the Lipchitz and smoothness assumptions on the parameterized policies and finite-time convergence guarantees to $\epsilon$-stationary policies under exact average-cost evaluation.
The numerical experiments show that our proposed algorithms perform competitively against PPO and some traditional queueing policy baselines.

\section*{Disclosure of the use of GenAI tools}

\begin{enumerate}
    \item ChatGPT-5.5 assisted to establish the proof of Lemma~\ref{BP_quadratic_drift}, especially starting from~\eqref{ineq:cross-term-BP} in Appendix.
    \item ChatGPT-5.5 assisted in choosing the form of the exponential Lypapunov function in Lemma~\ref{geometrical_drift}.
    \item ChatGPT-5.5 Pro wrote the proof of Lemma~\ref{lemma: Uniform geometric mixing}. We revised it for readability.
    \item ChatGPT-5.5 assisted in establishing the proof of Lemma~\ref{lemma: advantage_bound}, Lemma~\ref{lemma: pre stationary distribution difference}, and Lemma~\ref{lemma: stationary distribution difference}, especially the idea of using the one-step transition bound of queueing systems.
    \item ChatGPT-5.5 suggested using Paley–Zygmund inequality in the proof of Theorem~\ref{thm:positive_prob_per_iteration}.
    \item ChatGPT-5.5 assisted to establish the proof of irreducibility in Proposition~\ref{prop:well_defined_J}.
    \item The abstract, introduction and coclusion sections were drafted by the authors and polished with GPT-Astra. GPT-Astra has been used to proofread and fix minor technical issues throughout the paper as well. 
\end{enumerate}

%% file: sections/appendix.tex
\newpage
\section*{Appendix}

\etocsettagdepth{main}{none}
\etocsettagdepth{appendix}{subsection}
\etocsettocstyle
    {\section*{Contents}}
    {}
\tableofcontents

We use the following notations throughout this appendix. 
for a policy $\pi \in \Pi_H$, let $\mP^\pi$ denote is the one-step transition matrix under policy $\pi$.
For each $t$, let $\vp^{\pi, t}_{\vq}$ be the row in matrix $(\mP^\pi)^t$ that corresponds to the transition probability of state $\vq$.
By convention, $p^{\pi, 0}_{\vq} (\vq') = 1$ if $\vq' = \vq$; $p^{\pi, 0}_{\vq} (\vq') = 0$, otherwise. 

\section{Preliminary Lemmas} \label{section: prerequisite}

To prepare the following proofs, we first provide two useful lemmas regarding the queueing network model and back-pressure scheduling. The first lemma bounds the change of queeu lengths in one step. 

\input{sections/proofs/lemma_one_step_bound}

\input{sections/proofs/lemma_BPdrift}

\section{Proof of Proposition~\ref{prop:well_defined_J}} \label{section: prop:well_defined_J: proof}

\input{sections/proofs/prop_well_defined_J_proof}

\section{Proof of Lemma~\ref{lemma: smoothness}} \label{section: lemma: smoothness: proof}

\input{sections/proofs/lemma_smoothness}

\section{Proof of Proposition~\ref{prop: well_defined_value}} \label{section: prop:well_defined_value:proof}

\input{sections/proofs/prop_well_defined_value_proof}

\section{Proof of Theorem~\ref{thm:positive_prob_per_iteration}} \label{section: thm 1 proof}

\input{sections/proofs/thm_periteration}

\section{Proof of Theorem~\ref{thm: main_thm}} \label{section: thm 2 proof}

\input{sections/proofs/thm_main}

\section{Proof of Theorem~\ref{thm: partition}} \label{section: thm 3 proof}

\input{sections/proofs/thm_partition}

%% file: sections/proofs/lemma_one_step_bound.tex
\begin{lemma} \label{lemma: one_step_bound}
   For the queueing network model considered in this paper, the following inequalities holds under any policy:
    \begin{align*}
        &\| \vq(t+1) - \vq(t) \|_1 \le N+M \le 2N,\\
        &\left| \| \vq(t+1) \|_1 - \| \vq(t) \|_1 \right| \le N.
    \end{align*}
    Moreover, for a set $\gC \subseteq \gQ$,
    and define the set of immediate next states of $\gC$ as
    \begin{align*}
        \gC^{+} \coloneqq \{ \vq' | P(\vq' | \vq, \vu) > 0,
        \vq \in \gC, \vu \in \gU \}.
    \end{align*}
    If $\gC$ is finite, then $\gC^+$ is also finite, and $\gC \subseteq \gC^+.$
\end{lemma}

\begin{proof}
    According to the queue dynamic as stated in~\eqref{eq:queue-recursion}, we have
    \begin{align} \label{eq: 1-norm one-step}
        \| \vq(t+1) - \vq(t) \|_1 = \| \va(t) + (\mR^\top - \mI) \tilde \vs(t) \|_1 \le \| \va(t) - \tilde \vs(t) \|_1 + \| \mR^\top \tilde \vs(t) \|_1.
    \end{align}
    For the term $\va(t) - \tilde \vs(t)$,
    since for each queue $i \in [N]$, $a_i(t) \in \{0, 1\}$ and $\tilde s_i(t) \in \{0, 1\}$, we have $a_i(t) - \tilde s_i(t) \in \{-1, 0, 1\}$.
    Thus, we have
    \begin{align} \label{eq: a-s one-step}
        \| \va(t) - \tilde \vs(t) \|_1 \le N.
    \end{align}
    For the term $\mR^\top \tilde \vs(t)$, we have
    \begin{align} \label{eq: Rs one-step}
        \| \mR^\top \tilde \vs(t) \|_1 = \sum_{j = 1}^N \sum_{i = 1}^N R_{ij} \tilde s_i(t) = \sum_{i = 1}^N \left( \sum_{j = 1}^N R_{ij} \right) \tilde s_i(t) \le \sum_{i = 1}^N \tilde s_i(t) \le M,
    \end{align}
    where the first inequality holds because there is at most one downstream queue per queue, i.e., for $i \in [N]$, $\sum_{j = 1}^N R_{ij} \le 1$,
    and the last inequality holds because there are $M$ servers and thus at most $M$ service completions in one time slot.
    Thus, substituting~\eqref{eq: a-s one-step} and~\eqref{eq: Rs one-step} into~\eqref{eq: 1-norm one-step} gives,
    \begin{align*}
        \| \vq(t+1) - \vq(t) \|_1 \le N+M \le 2N.
    \end{align*}

    Since $\left| \| \vq(t+1) \|_1 - \| \vq(t) \|_1 \right|$ represents the total queue length change in one time slot, and there are at most $N$ exogenous arrivals joining the queue system, and at most $M$ jobs leaving the queue system, we have,
    \begin{align*}
        \left| \| \vq(t+1) \|_1 - \| \vq(t) \|_1 \right| \le \max \{N,M\} = N.
    \end{align*}

    Since for any queue state $\vq(t) = \vq$ and feasible scheduling $\vu(t) = \vu$, there is a positive probability at least
    \begin{align} \label{eq: p_0}
        p_0 = \prod_{i=1}^N (1 - \lambda_i) \prod_{i=1}^N (1 - \mu_i) > 0
    \end{align}
    such that there is no exogenous arrival or job leaving the system in one time slot. Accordingly, we have $P(\vq|\vq,\vu) > 0$.
    Thus, given a finite set $\gC \subseteq \gQ$, $\forall \vq \in \gC$, we have $\vq \in \gC^+$, i.e., $\gC \subseteq \gC^+$.
    Moreover, since we have $\| \vq(t+1) - \vq(t) \|_1 \le 2N$, and for any $\vq \in \gQ$, the set $\{\vq': \| \vq' - \vq \|_1 \le 2N\}$ is finite, we have the set $$\gC^+ \subseteq \bigcup_{\vq \in \gC} \{\vq': \| \vq' - \vq \|_1 \le 2N\}$$ is also finite.
\end{proof}

%% file: sections/proofs/lemma_BPdrift.tex
The following lemma establishes the Lyapunov drift for a quadratic Laypunov function, where the drift depends on the traffic slackness constant $\delta_c$.

\begin{lemma} \label{BP_quadratic_drift}
    Suppose Assumption~\ref{assumption:stability-condition} holds.
    Then for any state $\vq \in \gQ$, if the policy chooses the back-pressure action $\vu_{\bp}(\vq)$, we have:
    \begin{align} \label{eq:BP-quadratic-drift}
        \E\left[
            \left.
            \|\vq(t+1)\|_2^2-\|\vq(t)\|_2^2
            \right|
            \vq(t)=\vq,\,
            \vu(t)=\vu_{\bp}(\vq)
        \right]
        \le
        -2\delta_c\|\vq\|_1+4N^2,
    \end{align}
    where the queue dynamics follow~\eqref{eq:queue-recursion}.
\end{lemma}

\begin{proof}
    Expanding
    $\|\vq(t+1)\|_2^2-\|\vq(t)\|_2^2$, we have:
    \begin{align} \label{eq:quadratic-expansion-BP}
        \|\vq(t+1)\|_2^2 - \|\vq(t)\|_2^2
        =
        &2\sum_{i=1}^N
        q_i(t) (q_i(t+1) - q_i(t)) + \sum_{i=1}^N (q_i(t+1) - q_i(t))^2 \nonumber \\
        = 
        &2\sum_{i=1}^N
        q_i(t) (q_i(t+1) - q_i(t)) + \|\vq(t+1) - \vq(t)\|_2^2.
    \end{align}
    For the second term $\|\vq(t+1) - \vq(t)\|_2^2$, which characterizes the queue increment, we use the relation between the $2$-norm and $1$-norm as follows:
    \begin{align}
        \| \vq(t+1) - \vq(t) \|_2
        \le
        \| \vq(t+1) - \vq(t) \|_1
        \le 2N, \label{square_term_1}
    \end{align}
    where the last inequality follows from Lemma~\ref{lemma: one_step_bound}. This shows the second drift is bounded. Substituting~\eqref{square_term_1} into~\eqref{eq:quadratic-expansion-BP},
    taking the conditional expectation given
    $\vq(t)=\vq$ and the fixed scheduling action $\vu$ yields:
    \begin{align} \label{eq:quadratic-drift-reduction}
        \E\left[
            \left.
            \|\vq(t+1)\|_2^2-\|\vq(t)\|_2^2
            \right|
            \vq(t)=\vq,\,
            \vu(t)=\vu
        \right]
        \leq
        2\sum_{i=1}^N
        q_i
        \E\left[
            \left.
            q_i(t+1) - q_i(t)
            \right|
            \vq(t)=\vq,\,
            \vu(t)=\vu
        \right]
        +4N^2.
    \end{align}
    We next evaluate the conditional increment $q_i(t+1) - q_i(t)$. Following the queue dynamics $q_i(t+1) = q_i(t) + a_i(t) - \tilde{s}_i(t) + \sum_{j=1}^N R_{ji} \tilde{s}_j(t)$ in~\eqref{eq:queue-recursion}, we expand the conditional expectation as follows:
    \begin{align*}
        \E\left[
            \left.
            q_i(t+1) - q_i(t)
            \right|
            \vq(t)=\vq,\,
            \vu(t)=\vu
        \right] 
        =& \E\left[
            \left.
            a_i(t)
            \right|
            \vq(t)=\vq,\,
            \vu(t)=\vu
        \right] - \E\left[
            \left.
            \tilde{s}_i(t)
            \right|
            \vq(t)=\vq,\,
            \vu(t)=\vu
        \right]\\
        &+ \sum_{j=1}^N R_{ji} \E\left[
            \left.
            \tilde{s}_j(t)
            \right|
            \vq(t)=\vq,\,
            \vu(t)=\vu
        \right].
    \end{align*}
    By definition, for any queue $j$, the actual service-completion variable $\tilde{s}_j(t)$ depends on whether queue $j$ is empty, whether the scheduling decision $u_j(t)$ will select queue $j$, and whether the potential service $s_j(t)$ will complete in one time slot when it is selected, i.e.,
    $
        \tilde{s}_j(t)
        =
        \mathbbm{1}\{q_j(t)>0\}u_j(t)s_j(t).
    $
    Since the potential service-completion variable $s_j(t)$ follows a Bernoulli distribution independent of the queue $\vq(t)$ and the scheduling decision $\vu(t)$, and its expectation is $\E[s_j(t)]=\mu_j$, we have:
    \begin{align*}
        \E\left[
            \left.
            \tilde{s}_j(t)
            \right|
            \vq(t)=\vq,\,
            \vu(t)=\vu
        \right]
        = u_j\mathbbm{1}\{q_j>0\} \E\left[ s_j(t) \right] = \mu_j u_j\mathbbm{1}\{q_j>0\}.
    \end{align*}
    Notice that for any queue $i$, the arrival process $a_i(t)$ is also independent of the queue $\vq(t)$ and the scheduling decision $\vu(t)$ with $\E[a_i(t)]=\lambda_i$. We therefore obtain $\E[a_i(t)|\vq(t)=\vq,\vu(t)=\vu] = \E[a_i(t)]=\lambda_i$. Substitute these results back into the expectation of the increment, we have:
    \begin{align} \label{eq:mean-increment-BP}
        \E\left[
            \left.
            q_i(t+1) - q_i(t)
            \right|
            \vq(t)=\vq, \vu(t)=\vu
        \right]
        =
        \lambda_i
        -
        \mu_i u_i\mathbbm{1}\{q_i>0\}
        +
        \sum_{j=1}^N
        R_{ji}\mu_j u_j
        \mathbbm{1}\{q_j>0\}.
    \end{align}
    Multiplying~\eqref{eq:mean-increment-BP} by $q_i$
    and summing over $i\in[N]$, we have:
    \begin{align} \label{eq:cross-term-BP}
        &\sum_{i=1}^N
        q_i
        \E\left[
            \left.
            q_i(t+1) - q_i(t)
            \right|
            \vq(t)=\vq, \vu(t)=\vu
        \right]\nonumber\\
        =
        &\sum_{i=1}^N\lambda_i q_i
        -
        \sum_{i=1}^N
        \mu_i u_i\mathbbm{1}\{q_i>0\} q_i
        +
        \sum_{i=1}^N q_i
        \sum_{j=1}^N
        R_{ji}\mu_j u_j
        \mathbbm{1}\{q_j>0\} \nonumber \\
        =
        &\sum_{i=1}^N\lambda_i q_i
        -
        \sum_{i=1}^N
        \mu_i 
        \left(
            q_i-\sum_{j=1}^N R_{ij}q_j
        \right) \mathbbm{1}\{q_i>0\} u_i,
    \end{align}
    where the second equality holds by exchanging the summation over $j$ and the summation over $i$. Thus, the second term on the right-hand side of~\eqref{eq:cross-term-BP} is precisely the
    back-pressure scheduling objective, which is maximized by the back-pressure action $\vu_{\bp}(\vq)$. Therefore, we have:
    \begin{align} \label{ineq:cross-term-BP}
        \sum_{i=1}^N
        q_i
        \E\left[
            \left.
            q_i(t+1) - q_i(t)
            \right|
            \vq(t)=\vq, \vu(t)=\vu_\bp (\vq)
        \right]
        = \sum_{i=1}^N\lambda_i q_i
        -
        \max_{\vu \in \gU}\sum_{i=1}^N
        w_i(\vq) u_i.
    \end{align}
    For each server $m$ and queue $i$, let $\tilde{\varrho}_i(\delta) = \tilde{\lambda}_i(\delta_c)/\mu_i$. Recall that $\tilde{\vlambda}(\delta_c)$ is the effective arrival rate for each queue when the arrival rate is inflated by $\delta_c$, i.e., $\tilde{\vlambda}(\delta_c) = (\mI - \mR^\top)^{-1}(\vlambda + \delta_c \vone)$. Notice that for each server $m$, we have $\sum_{i\in \gN_m} \tilde{\varrho}_i(\delta_c) \leq 1$.
    We now construct a random scheduling policy as follows: for each server $m$, select queue $i\in\gN_m$ with probability $\tilde \varrho_i(\delta_c)$ and leave the server idle with the remaining probability.
    Let $\rvu$ denote the scheduling decision from the random policy. Then,
    $
        \E[\ervu_i]=\tilde{\varrho}_i(\delta_c),~\forall
        i\in[N].
    $
    and the support of $\rvu$ is $\gU$. According to the definition of the back-pressure action $\vu_{\bp}(\vq)$,
    for every decision $\vu\in \gU$, we have:
    \begin{align*}
        \sum_{i=1}^N
        w_i(\vq) u_i
        \leq \max_{\vu' \in \gU} \sum_{i=1}^N
        w_i(\vq) u_i'.
    \end{align*}
    Taking expectation with respect to the random
    scheduling policy, we have:
    \begin{align} \label{eq:BP-comparison-1}
        \max_{\vu \in \gU} \sum_{i=1}^N
        w_i(\vq) u_i
        \geq& \sum_{\vu \in \gU} \prob(\rvu = \vu) \sum_{i=1}^N
        w_i(\vq) u_i 
        = \sum_{m=1}^M \sum_{i\in \gN_m} \sum_{u_i \in \{0, 1\}} \prob(\ervu_i = u_i) 
        w_i(\vq) u_i \nonumber\\
        =& \sum_{m=1}^M \sum_{i\in \gN_m} \tilde{\varrho}_i(\delta_c) w_i(\vq)
        = \sum_{m=1}^M \sum_{i\in \gN_m} 
        \tilde{\lambda}_i(\delta_c)
        \left(
            q_i-\sum_{j=1}^N R_{ij}q_j
        \right)
        \mathbbm{1}\{q_i>0\}\nonumber\\
        =& \sum_{i=1}^N
        \tilde{\lambda}_i(\delta_c)
        \left(
            q_i-\sum_{j=1}^N R_{ij}q_j
        \right)
        \mathbbm{1}\{q_i>0\},
    \end{align}
    where the second equality and the last equality use the fact that $\{\gN_m\}$ is a partition of $[N]$.
    For any queue $i$ which is empty with $q_i=0$, we have:
    $
        q_i-\sum_{j=1}^N R_{ij}q_j
        =
        -\sum_{j=1}^N R_{ij}q_j
        \leq 0.
    $
    Since
    \[
    \tilde \vlambda (\delta_c) = \left( \mI - \mR^\top \right)^{-1} (\vlambda + \delta_c \vone) = \left( \sum_{n=0}^\infty \left( \mR^\top \right)^n \right) (\vlambda + \delta_c \vone) = \vlambda + \delta_c \vone + \left( \sum_{n=1}^\infty \left( \mR^\top \right)^n \right) (\vlambda + \delta_c \vone),
    \]
    we have for any $i$, $\tilde \lambda_i (\delta_c) \ge \lambda_i + \delta_c > 0$.
    Thus, adding the terms $q_i-\sum_{j=1}^N R_{ij}q_j$ corresponding to $q_i=0$ can only decrease
    the right-hand side of~\eqref{eq:BP-comparison-1}.
    Therefore,
    \begin{align} \label{eq:BP-comparison-2}
        \max_{\vu \in \gU} \sum_{i=1}^N
        w_i(\vq) u_i
        \geq
        \sum_{i=1}^N
        \tilde{\lambda}_i(\delta_c)
        \left(
            q_i-\sum_{j=1}^N R_{ij}q_j
        \right)
        =
        \left\langle
            \tilde{\vlambda}(\delta_c),
            (\mI-\mR)\vq
        \right\rangle.
    \end{align}
    Substituting~\eqref{eq:BP-comparison-2}
    into~\eqref{ineq:cross-term-BP}, we obtain
    \begin{align}
        &\sum_{i=1}^N
        q_i
        \E\left[
            \left.
            q_i(t+1)-q_i(t)
            \right|
            \vq(t)=\vq,\,
            \vu(t)=\vu_{\bp}(\vq)
        \right]\nonumber \\
        \leq &\langle\vlambda,\vq\rangle
        -
        \langle
            \tilde{\vlambda}(\delta_c),
            (\mI-\mR)\vq
        \rangle
        \nonumber\\
        = &\langle\vlambda,\vq\rangle
        -
        \left\langle
            (\mI-\mR^\top)^{-1}(\vlambda + \delta_c \vone),
            (\mI-\mR)\vq
        \right\rangle
        \nonumber\\
        =& - \delta_c \|\vq\|_1.
        \label{eq:cross-term-negative-BP}
    \end{align}
    Here, the first equality uses the traffic equation
    $
        \tilde{\vlambda}(\delta_c)
        =
        (\mI-\mR^\top)^{-1}(\vlambda + \delta_c \vone).
    $
    Finally, substituting~\eqref{eq:cross-term-negative-BP}
    into~\eqref{eq:quadratic-drift-reduction}, we have:
    \begin{align}
        \E\left[
            \left.
            \|\vq(t+1)\|_2^2-\|\vq(t)\|_2^2
            \right|
            \vq(t)=\vq,\,
            \vu(t)=\vu_{\bp}(\vq)
        \right]
        \leq
        -2\delta_c\|\vq\|_1+4N^2,
    \end{align}
    which completes the proof.
\end{proof}

%% file: sections/proofs/prop_well_defined_J_proof.tex
    \textbf{Irreducibility:} We first show that any $\pi_{H,\vtheta}\in \Pi_H$ is weakly work-conserving, i.e., at least one nonempty queue is served when the system is nonempty.
    This is immediate when $\|\vq\|_1\le H$, because the corresponding learnable policy $\pi_\vtheta$ is weakly work-conserving by assumption. 
    When $\|\vq\|_1>H$, we consider the back-pressure scheduling. Note that the back-pressure weights are all zero if and and if all queues are empty. This holds because (1) the back-pressure weight of a  queue without a downstream queue is its queue length, so zero back-pressure weight means zero queue length for that queue; and (2) back-pressure weights are all zero mean all queue legnths are the same. 
    Hence when the system is nonempty, some back-pressure weight is strictly positive, and some nonemtpy queue will be scheduled under back-pressure.
    Next, under a weakly work-conserving policy,  if the system is nonemtpy, then at each time slot, with probability at least
    \begin{align}
        p_s
        &=
        \min_{i'}\mu_{i'}\prod_{i=1}^N(1-\lambda_i)
        >0,
        \label{eq:p-star}
    \end{align} 
    there is no external arrival and at least one job is scheduled and completes its service. 

    Note that each job is served by at most $M$ servers before leaving the system. 
    Starting from queue starting from $\vq \in \gQ$, the system becomes empty in $N\|\vq\|_1$ time slots with probability at least 
    \begin{align}
        p_s^{N\|\vq\|_1}.
        \label{eq:emptying}
    \end{align}

    Then the Markov chain defined on the state space is irreducible because any state can reach $\vq=0$. 
    Also, since each arrival rate $\lambda_i > 0$, there is a positive probability that in one step, there is one exogenous arrival to one specific queue, no arrivals to all other queues, and no service completion.
    Thus, there is a positive probability that the empty state $\vzero$ goes to any state $\vq' \in \gQ$ in a finite number of steps.

    \textbf{Aperiodicity:} applying exactly self-loop argument in~\eqref{eq: p_0} in Lemma~\ref{lemma: one_step_bound} gives that for any state $\vq \in \gQ$, there is a positive probability of the self-loop, which implies the Markov chain is aperiodic.

    \textbf{Positive recurrence:} Lemma~\ref{BP_quadratic_drift} shows that
    \begin{align*}
        \E\left[
            \left.
            \|\vq(t+1)\|_2^2-\|\vq(t)\|_2^2
            \right|
            \vq(t)=\vq,\,
            \vu(t)=\vu_{\bp}(\vq)
        \right]
        \le
        -2\delta_c\|\vq\|_1+4N^2.
    \end{align*}
    Then, for any policy $\pi_{H, \vtheta}$, when $\|\vq\|_1 > \max\{H, (4N^2+\varepsilon)/2\delta_c\}$ for some $\varepsilon > 0$, we have,
    \begin{align} \label{eq: quadratic_outfinite_bound}
        \E_{\vu(t) \sim \pi_{H,\vtheta} (\cdot|\vq(t))} \left[
            \left.
            \|\vq(t+1)\|_2^2-\|\vq(t)\|_2^2
            \right|
            \vq(t)=\vq
        \right]
        &=
        \E\left[
            \left.
            \|\vq(t+1)\|_2^2-\|\vq(t)\|_2^2
            \right|
            \vq(t)=\vq,\,
            \vu(t)=\vu_{\bp}(\vq)
        \right] \nonumber \\
        &\le
        -2\delta_c\|\vq\|_1+4N^2
        \le
        -2\delta_c\frac{4N^2+\varepsilon}{2\delta_c}+4N^2 = -\varepsilon,
    \end{align}
    where the first equality holds because when $\|\vq\|_1 > H$, the policy $\pi_{H,\vtheta}$ deploys the back-pressure action $\vu_\bp(\vq)$.
    When $\|\vq\|_1 \le \max\{H, (4N^2+\varepsilon)/2\delta_c\}$, according to the one-step transition bound in Lemma~\ref{lemma: one_step_bound}, we have,
    \begin{align} \label{eq: quadratic_finite_bound}
        &\E_{\vu(t) \sim \pi_{H,\vtheta} (\cdot|\vq(t))} \left[
            \left.
            \|\vq(t+1)\|_2^2-\|\vq(t)\|_2^2
            \right|
            \vq(t)=\vq
        \right] \nonumber \\
        \le
        &\E_{\vu(t) \sim \pi_{H,\vtheta} (\cdot|\vq(t))} \left[
            \left.
            \|\vq(t+1)\|_1^2
            \right|
            \vq(t)=\vq
        \right]
        \le
        (\|\vq\|_1+N)^2
        < \infty.
    \end{align}
    Thus, since the set $\{\vq: \|\vq\|_1 \le \max\{H, (4N^2+\varepsilon)/2\delta_c\}\}$ is finite, according to the Foster's theorem in~\cite{Foster}, the induced Markov chain is positive recurrent.

    \textbf{Finiteness of $J(\pi_{H,\vtheta})$:} For $\|\vq\|_1 \le H$, according to Lemma~\ref{lemma: one_step_bound}, we have
    \begin{align}
        \E_{\vu(t) \sim \pi_{H,\vtheta} (\cdot|\vq(t))} \left[
            \left.
            \|\vq(t+1)\|_2^2-\|\vq(t)\|_2^2
            \right|
            \vq(t)=\vq
        \right]
        &\le
        (H+N)^2
        &\le
        -2\delta_c \|\vq\|_1 + 2\delta_c H + (H+N)^2,
    \end{align}
    where the first inequality follows from the same argument used to establish ~\eqref{eq: quadratic_finite_bound}.
    Thus, combining it with Lemma \ref{BP_quadratic_drift} yields
    \begin{align}
        \E_{\vu(t) \sim \pi_{H,\vtheta} (\cdot|\vq(t))} \left[
            \left.
            \|\vq(t+1)\|_2^2-\|\vq(t)\|_2^2
            \right|
            \vq(t)=\vq
        \right]
        &\le
        -2\delta_c \|\vq\|_1 + \max\{4N^2, 2\delta_c H + (H+N)^2\}.
    \end{align}
    Taking expectation over $\vq \sim \vnu^{\pi_{H,\vtheta}}$ on both sides gives,
    \begin{align*}
        0 \le -2\delta_c J(\pi_{H,\vtheta})+\max\{4N^2, 2\delta_c H + (H+N)^2\},
    \end{align*}
    where the left-hand-side holds because the one-step transition of the stationary distribution is still the stationary distribution, and the right-hand-side holds for $c(\vq) \leq \|\vq\|_1$.
    Thus, we have $$J(\pi_{H,\vtheta}) \le \frac{\max\{4N^2, 2\delta_c H + (H+N)^2\}}{2\delta_c} < \infty,$$
    which completes the proof of $J(\pi_{H,\vtheta})$'s finiteness.

%% file: sections/proofs/lemma_smoothness.tex
Consider any two policies $\pi = \pi_{H, \vtheta}$ and $\pi' = \pi_{H, \vtheta'}$, where the dependence on $\vtheta$ and $\vtheta'$ are omitted for simplicity when there is no ambiguity.
The performance difference lemma (Lemma A.9,~\cite{murthy2024}) gives the difference between the average costs of the two policies:
\begin{align}\label{eq: performance_difference_lemma}
    J(\pi') - J(\pi) =& \sum_{\vq \in \gQ} \nu^{\pi'} (\vq) \sum_{\vu \in \gU} \left( \pi' (\vu | \vq) - \pi (\vu | \vq) \right) Q^\pi (\vq, \vu) \nonumber\\
    =& \sum_{\vq \in \gQ} \nu^{\pi'} (\vq) \sum_{\vu \in \gU} \left( \pi' (\vu | \vq) - \pi (\vu | \vq) \right) A^\pi (\vq, \vu),
\end{align}
where the second equality holds because $\sum_{\vu \in \gU} \left( \pi' (\vu | \vq) - \pi (\vu | \vq) \right) V^\pi (\vq) = 0.$ We can decompose~\eqref{eq: performance_difference_lemma} as follows:
\begin{align} \label{eq: pdl_decompose}
    J(\pi') - J(\pi) =& \sum_{\vq \in \gQ} \left( \nu^{\pi'} (\vq) - \nu^{\pi} (\vq) + \nu^{\pi} (\vq) \right) \sum_{\vu \in \gU} (\pi' (\vu | \vq) - \pi (\vu | \vq) ) A^\pi (\vq, \vu) \nonumber \\
    =&  \sum_{\vq \in \gQ} \nu^{\pi} (\vq) \sum_{\vu \in \gU} (\pi' (\vu | \vq) - \pi (\vu | \vq) ) A^\pi (\vq, \vu)  \nonumber \\
    &+  \sum_{\vq \in \gQ} \left( \nu^{\pi'} (\vq) - \nu^{\pi} (\vq) \right) \sum_{\vu \in \gU} (\pi' (\vu | \vq) - \pi (\vu | \vq) ) A^\pi (\vq, \vu),
\end{align}
From the decomposition in~\eqref{eq: pdl_decompose}, we can see the performance difference is attributed to two parts:
the first term represents the policy difference, and the second term represents the policy difference plus the stationary-distribution difference. The policy gradient theorem (Theorem 1,~\cite{Sutton1999}) gives that,
\begin{align} \label{eq: policy_gradient_theorem}
    \nabla_\vtheta J(\pi) = \sum_{\vq \in \gQ} \nu^\pi(\vq) \sum_{\vu \in \gU} \nabla_\vtheta \pi(\vu | \vq) Q^\pi (\vq, \vu) = \sum_{\vq \in \gQ} \nu^\pi(\vq) \sum_{\vu \in \gU} \nabla_\vtheta \pi(\vu | \vq) A^\pi (\vq, \vu),
\end{align}
where the second equality above holds because
$
    \sum_{\vu \in \gU} \nabla_\vtheta \pi(\vu | \vq) V^\pi (\vq) = \nabla_\vtheta \left( \sum_{\vu \in \gU} \pi(\vu | \vq) \right) V^\pi (\vq) = 0.
$ Note that the policy gradient theorem for a finite state space \cite{Sutton1999} applies here for the countable state space because  $\nabla_\vtheta \pi(\vu | \vq)=0$ when $\|\vq\|_1>H$ under a back-pressure guided policy. 
Combining~\eqref{eq: pdl_decompose} and~\eqref{eq: policy_gradient_theorem} gives,
\begin{align} \label{eq: second order expression}
    J(\pi') - J(\pi) - \nabla_\vtheta J(\pi)^\top (\vtheta' - \vtheta) =& \sum_{\vq \in \gQ} \nu^{\pi} (\vq) \sum_{\vu \in \gU} (\pi' (\vu | \vq) - \pi (\vu | \vq) - \nabla_\vtheta \pi(\vu | \vq)^\top (\vtheta' - \vtheta) ) A^\pi (\vq, \vu) \nonumber \\
    &+ \sum_{\vq \in \gQ} \left( \nu^{\pi'} (\vq) - \nu^{\pi} (\vq) \right) \sum_{\vu \in \gU} (\pi' (\vu | \vq) - \pi (\vu | \vq) ) A^\pi (\vq, \vu) \nonumber \\
    =& \underbrace{ \sum_{\vq: \|\vq\|_1 \le H} \nu^{\pi} (\vq) \sum_{\vu \in \gU} (\pi' (\vu | \vq) - \pi (\vu | \vq) - \nabla_\vtheta \pi(\vu | \vq)^\top (\vtheta' - \vtheta) ) A^\pi (\vq, \vu) }_{\mathsf{I_1}} \nonumber \\
    &+ \underbrace{ \sum_{\vq: \|\vq\|_1 \le H} \left( \nu^{\pi'} (\vq) - \nu^{\pi} (\vq) \right) \sum_{\vu \in \gU} (\pi' (\vu | \vq) - \pi (\vu | \vq) ) A^\pi (\vq, \vu) }_{\mathsf{I_2}}
\end{align}
The last equality holds because when $\|\vq\|_1 > H$, both policies $\pi$ and $\pi'$ deploy the back-pressure action $\vu_{\bp} (\vq)$, and we have for any $\vu \in \gU$, $\pi' (\vu | \vq) = \pi (\vu | \vq)$. Moreover, the probability $\pi (\vu | \vq)$ is determined by the back-pressure algorithm and does not change with $\vtheta$ for $\|\vq\|_1 > H$, so we have $\nabla_\vtheta \pi(\vu | \vq) = \vzero$.
Thus, when $\|\vq\|_1 > H$, for any $\vu \in \gU$, $\pi' (\vu | \vq) - \pi (\vu | \vq) - \nabla_\vtheta \pi(\vu | \vq)^\top (\vtheta' - \vtheta) = 0$.
    
We first deal with the term $\mathsf{I_1}$. We have:
\begin{align} \label{eq: I_1}
    \left| \mathsf{I_1} \right| &= \left| \sum_{\vq: \|\vq\|_1 \le H} \nu^{\pi} (\vq) \sum_{\vu \in \gU} \left(\pi' (\vu | \vq) - \pi (\vu | \vq) - \nabla_\vtheta \pi(\vu | \vq)^\top (\vtheta' - \vtheta) \right) A^\pi (\vq, \vu) \right| \nonumber \\
    &\le \sum_{\vq: \|\vq\|_1 \le H} \nu^{\pi} (\vq) \sum_{\vu \in \gU} \left| \pi' (\vu | \vq) - \pi (\vu | \vq) - \nabla_\vtheta \pi(\vu | \vq)^\top (\vtheta' - \vtheta) \right| \left| A^\pi (\vq, \vu) \right| \nonumber \\
    &\le \max_{\substack{\vq: \|\vq\|_1 \le H \\ \vu \in \gU}} \left| A^\pi (\vq, \vu) \right|
    \max_{\vq: \|\vq\|_1 \le H} \sum_{\vu \in \gU} \left| \pi' (\vu | \vq) - \pi (\vu | \vq) - \nabla_\vtheta \pi(\vu | \vq)^\top (\vtheta' - \vtheta) \right| \sum_{\vq: \|\vq\|_1 \le H} \nu^{\pi} (\vq) \nonumber \\
    &\le \max_{\substack{\vq: \|\vq\|_1 \le H \\ \vu \in \gU}} \left| A^\pi (\vq, \vu) \right| \frac{L_1}{2} \|\vtheta' - \vtheta\|_2^2,
\end{align} 
where the last inequality uses the fact that $\sum_{\vq: \|\vq\|_1 \le H} \nu^\pi(\vq) \le \sum_{\vq\in\gQ} \nu^\pi(\vq) = 1$, and the smoothness of the policy class assumed in Assumption~\ref{assumption:smooth-policy}. We then deal with the term $\mathsf{I_2}$. We have:
\begin{align} 
    |\mathsf{I_2}| &= \left| \sum_{\vq: \|\vq\|_1 \le H} \left( \nu^{\pi'} (\vq) - \nu^{\pi} (\vq) \right) \sum_{\vu \in \gU} (\pi' (\vu | \vq) - \pi (\vu | \vq) ) A^\pi (\vq, \vu) \right| \nonumber \\
    &\le \max_{\substack{\vq: \|\vq\|_1 \le H \\ \vu \in \gU}} \left| A^\pi (\vq, \vu) \right| 
    \max_{\vq: \|\vq\|_1 \le H}  \sum_{\vu \in \gU} \left|(\pi' (\vu | \vq) - \pi (\vu | \vq) ) \right| 
    \sum_{\vq: \|\vq\|_1 \le H} \left| \nu^{\pi'} (\vq) - \nu^{\pi} (\vq) \right| \nonumber \\
    &\le \max_{\substack{\vq: \|\vq\|_1 \le H \\ \vu \in \gU}} \left| A^\pi (\vq, \vu) \right| 
    L_0 \|\vtheta' - \vtheta\|_2 
    \|\vnu^{\pi'} - \vnu^\pi\|_1 ,
    \label{eq: I_2_pre}
\end{align}
where the third inequality follows from the Lipschitz continuity of the policy class assumed in Assumption~\ref{assumption:smooth-policy}.

Observing~\eqref{eq: I_1} and~\eqref{eq: I_2_pre}, there are two term we need to characterize:
\[
    \max_{\substack{\vq: \|\vq\|_1 \le H \\ \vu \in \gU}} \left| A^\pi (\vq, \vu) \right|
    \quad \text{and} \quad
    \|\vnu^{\pi'} - \vnu^\pi\|_1.
\]
We will first prove the geometric mixing property of Markov chains induced from policies in the back-pressure guarded policy class in Appendix~\ref{section: geometric mixing}, and bound the two term above based on this respectively in Appendix~\eqref{section: advantage bound} and~\ref{section: stationary difference}.
Finally, we will complete the proof of smoothness in Appendix~\ref{section: smoothness complete}.

\subsection{Geometric mixing} \label{section: geometric mixing}
We first establish the geometric drift condition which is the key to prove geometric mixing.

\input{sections/proofs/lemma_geodrift}

Now we provide the statement and the proof of geometric mixing.
Note that in Lemma~\ref{lemma: Uniform geometric mixing} we slightly abuse the notation $\rho$ which is used to represent the load of a server in the main part of the paper.

\input{sections/proofs/lemma_geomix}

\subsection{Advantage function bound} \label{section: advantage bound}

\input{sections/proofs/lemma_advantage}

\subsection{Stationary distribution difference bound} \label{section: stationary difference}

\input{sections/proofs/lemma_stationarydifference}

\subsection{Proof of Lemma~\ref{lemma: smoothness}} \label{section: smoothness complete}

After characterizing
\[
    \max_{\substack{\vq: \|\vq\|_1 \le H \\ \vu \in \gU}} \left| A^\pi (\vq, \vu) \right|
    \quad \text{and} \quad
    \|\vnu^{\pi'} - \vnu^\pi\|_1,
\]
we can complete the proof of smoothness.

From Lemma~\ref{lemma: stationary distribution difference}, we can characterize the stationary distribution difference as follows:
\begin{align*}
    \|\vnu^{\pi'} - \vnu^\pi\|_1 
    \leq& \frac{2 C_\mix \exp(\kappa(H+N))}{1-\rho_\mix} \max_{\vq:\|\vq\|_1 \leq H} \sum_{\vu \in \gU} \left| \pi'(\vu|\vq) - \pi(\vu|\vq) \right| \\
    \leq& \frac{2 C_\mix L_0 \exp(\kappa(H+N))}{1-\rho_\mix} \|\vtheta' - \vtheta\|_2,
\end{align*} 
where the last inequality uses the Lipschitz continuity of the policy class. Therefore, we obtain:
\begin{align}\label{eq: I_2}
    |\mathsf{I_2}| \leq \frac{2 L_0^2 C_\mix \exp(\kappa(H+N))}{1-\rho_\mix} 
    \|\vtheta' - \vtheta\|_2^2 \max_{\substack{\vq: \|\vq\|_1 \le H \\ \vu \in \gU}} \left| A^\pi (\vq, \vu) \right|.
\end{align}
Substituting~\eqref{eq: I_1} and~\eqref{eq: I_2} into~\eqref{eq: second order expression} gives,
\begin{align} \label{eq: smoothness_final}
    &\left| J(\pi') - J(\pi) - \nabla_\vtheta J(\pi)^\top (\vtheta' - \vtheta) \right| \nonumber \\
    \le& \max_{\substack{\vq: \|\vq\|_1 \le H \\ \vu \in \gU}} \left| A^\pi (\vq, \vu) \right| 
    \left( \frac{L_1}{2} + \frac{2 C_\mix \exp(\kappa(H+N))}{1-\rho_\mix} L_0^2 \right) 
    \|\vtheta' - \vtheta\|_2^2.
\end{align}
Finally, substituting the advantage function bound in Lemma~\ref{lemma: advantage_bound} into~\eqref{eq: smoothness_final} gives $L_J$ as follows:
\[
    L_J = 4 C_\mix (H+N) \exp(\kappa(H+N)) \frac{1+\rho_\mix}{\rho_\mix (1-\rho_\mix)^2} 
    \left( \frac{L_1}{2} + \frac{2 C_\mix \exp(\kappa(H+N))}{1-\rho_\mix} L_0^2 \right).
\]
Considering the dependence of constants $\kappa, C_\mix$ and $\rho_\mix$ analyzed in Lemma~\ref{geometrical_drift} and Lemma~\ref{lemma: Uniform geometric mixing},
$L_J$ depends only on $\delta_c$, other network parameters, $H$, $N$, $L_0$, and $L_1$, and not on any specific policy.

%% file: sections/proofs/lemma_geodrift.tex
Define the following constant:
\begin{align*}
    &\kappa = \min\left\{\frac{\delta_c}{2N(4N + \delta_c)}, \frac{1}{\delta_c} \right\} >0, \quad \gamma = 1-\frac{\kappa\delta_c}{8} \in (0,1), \\
    &H_{\delta_c, N} = \max \left\{ H, \frac{4N^2\sqrt{N}}{\delta_c} \right\}, \quad B = \exp \left( \kappa \left( H_{\delta_c, N} + 2N \right) \right) < \infty.
\end{align*}
Define the exponential Lyapunov function $W(\vq)$ as follows:
\begin{align} \label{eq: Lyapunov function}
    W(\vq) \coloneqq \exp \left( \kappa \| \vq \|_2 \right).
\end{align}
We show that the exponential Lyapunov function is geometrically decreasing outside a finite regime as follows.

\begin{lemma} \label{geometrical_drift}
    Suppose Assumption~\ref{assumption:stability-condition} holds.
    Then for all $\pi \in \Pi_H$, we have:
    \[
        \E_\pi\left[
            \left. W(\vq(t+1))
            \right| \vq(t)=\vq
        \right]
        \le
        \gamma W(\vq) + B \mathbbm{1} \left\{ \| \vq \|_1 \le H_{\delta_c, N}\right\},
        \qquad
        \forall \vq\in\gQ.
    \]
\end{lemma}

\begin{proof}
    Fix an arbitrary policy $\pi \in \Pi_H$, and we separate two cases: $\|\vq\|_1 \leq H_{\delta_c, N}$, where we use the learnable RL policy or the back-pressure action, and $\|\vq\|_1 > H_{\delta_c, N} \ge H$, where we use the back-pressure action. For any state $\vq$ and conditioned on the event $\vq(t)=\vq$, we consider the the increment $\| \vq(t+1) \|_2 - \| \vq(t) \|_2$ as follows:
    \begin{align} \label{eq:Z-bounded}
        \|\vq(t+1)\|_2-\|\vq(t)\|_2
        \le
        \|\vq(t+1)-\vq(t)\|_2
        \le
        \|\vq(t+1)-\vq(t)\|_1
        \le 2N,
    \end{align}
    where the first inequality is the triangle inequality, the second inequality uses the relation between the $2$-norm and the $1$-norm, and the final inequality uses Lemma~\ref{lemma: one_step_bound}. Then conditioned on $\vq(t)=\vq$, we can upper bound $W(\vq(t+1))$ where $\|\vq\|_1 \leq H_{\delta_c, N}$ as follows:
    \begin{align*}
        W(\vq(t+1)) = \exp \left( \kappa \| \vq(t+1) \|_2 \right) \leq \exp \left( \kappa \left(\| \vq(t) \|_2 + 2 N\right) \right) \leq \exp \left( \kappa \left(H_{\delta_c, N} + 2 N\right) \right) = B.
    \end{align*}
    This establishes for $\vq$ such that $\|\vq\|_1 \leq H_{\delta_c, N}$, we have:
    \begin{align}\label{eq:inside-drift}
        \E_\pi\left[
            \left. W(\vq(t+1))
            \right| \vq(t)=\vq
        \right]
        \le
        B \mathbbm{1} \{ \| \vq \|_1 \le H_{\delta_c, N}\} \leq
        \gamma W(\vq) + B \mathbbm{1} \{ \| \vq \|_1 \le H_{\delta_c, N}\}.
    \end{align}
    We then consider the case where the queue length is large, i.e., $\|\vq\|_1 > H_{\delta_c, N} \ge H$.
    By the definition of $\Pi_H$, the policy $\pi$ uses the back-pressure action $\vu_{\bp}(\vq)$ on these states. 
    Thus, we have:
    \begin{align} \label{eq:W-ratio}
        \frac{ \E_\pi \left[ \left. W(\vq(t+1)) \right| \vq(t) = \vq \right]}{W(\vq)}
        =
        \E \left[
            \left.
            \exp \left( \kappa (\| \vq(t+1) \|_2 - \| \vq(t) \|_2) \right)
            \right|
            \vq(t) = \vq,\,
            \vu(t) = \vu_{\bp}(\vq)
        \right].
    \end{align}
    The main step for proving Lemma~\ref{geometrical_drift} is  to upper bound the right-hand side of~\eqref{eq:W-ratio} by $\gamma$, which leads to deriving an upper bound for $\| \vq(t+1) \|_2 - \| \vq(t) \|_2$. We use the quadratic Lyapunov function as follows:
    \begin{align*}
        \|\vq(t+1)\|_2^2 - \|\vq(t)\|_2^2
        =
        2\|\vq(t)\|_2 \| (\vq(t+1) \|_2 - \| \vq(t) \|_2) + (\| \vq(t+1) \|_2 - \| \vq(t) \|_2)^2,
    \end{align*}
    which means the difference of $2$-norm is upper bounded by the quadratic Lyapunov function difference as follows:
    \begin{align*}
        \| \vq(t+1) \|_2 - \| \vq(t) \|_2 
        =& \frac{1}{2\|\vq(t)\|_2}\left( \|\vq(t+1)\|_2^2 - \|\vq(t)\|_2^2 - (\| \vq(t+1) \|_2 - \| \vq(t) \|_2)^2 \right)\\
        \leq & \frac{1}{2\|\vq(t)\|_2} \left( \|\vq(t+1)\|_2^2 - \|\vq(t)\|_2^2 \right).
    \end{align*}
   We then take the conditional expectation over the event $\vq(t) = \vq$ and $\vu(t) = \vu_{\bp}(\vq)$ to obtain
    \begin{align*}
        &\E\left[
            \left. \| \vq(t+1) \|_2 - \| \vq(t) \|_2
            \right|
            \vq(t) = \vq,\,
            \vu(t) = \vu_{\bp}(\vq)
        \right]\\
        \leq & \frac{1}{2\|\vq\|_2} \E\left[
            \left. \| \vq(t+1) \|_2^2 - \| \vq(t) \|_2^2
            \right|
            \vq(t) = \vq,\,
            \vu(t) = \vu_{\bp}(\vq)
        \right]\\
        \leq & \frac{1}{2\|\vq\|_2} \left( - 2 \delta_c \|\vq\|_1 + 4N^2 \right) = -\frac{ \delta_c \| \vq \|_1}{ \| \vq \|_2}
        +
        \frac{2N^2}{ \| \vq \|_2},
    \end{align*}
    where the second inequality follows from Lemma~\ref{BP_quadratic_drift}.    
    Since
    $
        {\|\vq\|_1}/{\sqrt{N}}
        \le
        \|\vq\|_2
        \le
        \|\vq\|_1,
    $
 we have
    \begin{align} \label{eq:negative-Z-drift}
        \E\left[
            \left. \| \vq(t+1) \|_2 - \| \vq(t) \|_2
            \right|
            \vq(t) = \vq,\,
            \vu(t) = \vu_{\bp}(\vq)
        \right]
        \le
        -\delta_c
        +
        \frac{2N^2\sqrt{N}}{\|\vq\|_1} \leq -\frac{\delta_c}{2},
    \end{align}
    where the last inequality holds because we are considering $\|\vq\|_1 > H_{\delta_c, N} \geq 4N^{2}\sqrt{N}/\delta_c$.
    We then convert the drift of $\|\vq\|_2$ to the exponential Lyapunov function. Recall that $\| \vq(t+1) \|_2 - \| \vq(t) \|_2\le 2N$. By Taylor's theorem for the exponential function, we have
    \begin{align} \label{eq:exp-taylor}
        \exp\left(\kappa (\| \vq(t+1) \|_2 - \| \vq(t) \|_2)\right)
        \le&
        1 + \kappa (\| \vq(t+1) \|_2 - \| \vq(t) \|_2)
        +
        \frac{\exp(2\kappa N)}{2}\kappa^2 (\| \vq(t+1) \|_2 - \| \vq(t) \|_2)^2\nonumber\\
        \leq& 1 + \kappa (\| \vq(t+1) \|_2 - \| \vq(t) \|_2)
        +
        2\kappa^2N^2\exp(2\kappa N)
    \end{align}
    Taking conditional expectation in~\eqref{eq:exp-taylor} and using
    \eqref{eq:negative-Z-drift}, we obtain
    \begin{align} \label{eq:exp-drift-pre}
        \E\left[
            \left.
            \exp\left(\kappa (\| \vq(t+1) \|_2 - \| \vq(t) \|_2)\right)
            \right|
            \vq(t) = \vq,\,
            \vu(t) = \vu_{\bp}(\vq)
        \right]
        \le
        1-\frac{\kappa\delta_c}{2}
        +
        2\kappa^2N^2\exp(2\kappa N).
    \end{align}
    If we choose $\kappa$ to be small enough such that
    \begin{align}
        \kappa = \min\left\{\frac{\delta_c}{2N(4N + \delta_c)}, \frac{1}{\delta_c} \right\}, \label{kappa}
    \end{align}
    the second term is upper bounded by $\kappa \delta_c /4$. To see this, define $\overline{\kappa} = {\delta_c}/{2N(4N + \delta_c)}$ and we have
    \begin{align*}
        2\overline{\kappa}N^2\exp(2\overline{\kappa} N) =& \frac{\delta_c N}{4N + \delta_c} \exp\left( \frac{\delta_c }{4N + \delta_c} \right) = \frac{\delta_c N}{4N + \delta_c}\exp\left( \frac{\frac{\delta_c}{4N}}{1 + \frac{\delta_c}{4N}} \right)\\
        \leq & \frac{\delta_c N}{4N + \delta_c}\left( 1 + \frac{\delta_c}{4N} \right) = \frac{\delta_c}{4},
    \end{align*}
    where the inequality uses $e^{\frac{x}{1+x}} \leq 1+x$ for $x\geq 0$.
    Since $2\kappa N^2\exp(2\kappa N)$ is monotonically increasing in $\kappa$ and the chosen $\kappa$ in \eqref{kappa} is no large than $\overline{\kappa}$, $2\kappa N^2\exp(2\kappa N) \leq {\delta_c}/{4}$.
    Then, we can further derive from  ~\eqref{eq:exp-drift-pre} that
    \begin{align}
        \E\left[
            \left.
            \exp\left(\kappa (\| \vq(t+1) \|_2 - \| \vq(t) \|_2)\right)
            \right|
            \vq(t)=\vq,\,
            \vu(t)=\vu_{\bp}(\vq)
        \right]
        \le
        1-\frac{\kappa\delta_c}{4} \le \gamma.
    \end{align}
    Combining the above inequality with~\eqref{eq:W-ratio}, we have:
    \begin{align} \label{eq:outside-drift}
        \E_\pi\left[
            \left.
            W(\vq(t+1))
            \right|
            \vq(t)=\vq
        \right]
        \le
        \gamma W(\vq),
        \qquad
        \|\vq\|_1>H_{\delta_c, N}.
    \end{align}
    Combining~\eqref{eq:outside-drift} and~\eqref{eq:inside-drift} yields the following inequality:
    \[
        \E_\pi\left[
            \left.
            W(\vq(t+1))
            \right|
            \vq(t)=\vq
        \right]
        \le
        \gamma W(\vq)
        +
        B\mathbbm{1}\{\|\vq\|_1\le H_{\delta_c, N}\},
    \]
    for every $\vq\in\gQ$ and every $\pi\in\Pi_H$, which proves the lemma.
\end{proof}

%% file: sections/proofs/lemma_geomix.tex
\begin{lemma} \label{lemma: Uniform geometric mixing}
Suppose Assumption 1 holds. Then there exist constants $C_{\rm mix}<\infty$ and $\rho_{\rm mix}\in(0,1)$ such that
\begin{align}
\left\|\vp_{\vq}^{\pi,t}-\vnu^\pi\right\|_{\rm TV}
&\le C_{\rm mix}W(\boldsymbol q)\rho_{\rm mix}^t,
\qquad \forall \pi \in \Pi_H,\ \boldsymbol q\in{\cal Q},\ t\ge0,
\label{eq:geometric-mixing}
\end{align}
$C_{\rm mix}$ and $\rho_{\rm mix}$ depend only on the common drift constants in Lemma 5, the network parameters, $H$, and $N$, and not on $\pi$.
\end{lemma}

\begin{proof}
Fix $\pi\in{\Pi}_H$.
All constants below
will be chosen independently of $\pi$. Let
${\cal C}=\{\boldsymbol q\in{\cal Q}:
\|\boldsymbol q\|_1\le H_{\delta_c,N}\}$.

Fix an initial queue-length vector $\boldsymbol q(0)\in{\cal Q}$.
We use $\mathbb P_{\boldsymbol q(0)}$ and
$\mathbb E_{\boldsymbol q(0)}$ to denote probability and expectation
conditional on this initial state, and we omit the dependency on policy $\pi$ here.
Let
${\cal F}_t=\sigma(\boldsymbol q(0),\ldots,\boldsymbol q(t))$
be the natural filtration and define
$\tau_{\cal C}=\inf\{t\ge0:\boldsymbol q(t)\in{\cal C}\}$.
Recall that an integrable process $\{M_t\}$ adapted to
$\{{\cal F}_t\}$ is a supermartingale if
\begin{align}
\mathbb E_{\boldsymbol q(0)}[M_{t+1}\mid{\cal F}_t]
&\le M_t,
\qquad t\ge0.
\label{eq:supermartingale-definition}
\end{align}
Define
$M_t=\gamma^{-(t\wedge\tau_{\cal C})}
W(\boldsymbol q(t\wedge\tau_{\cal C}))$.
On $\{\tau_{\cal C}\le t\}$, the process has already stopped, so
$M_{t+1}=M_t$. On $\{\tau_{\cal C}>t\}$, we have
$\boldsymbol q(t)\notin{\cal C}$ and hence, by
Lemma~\ref{geometrical_drift},
\begin{align}
\mathbb E_{\boldsymbol q(0)}[M_{t+1}\mid{\cal F}_t]
&=
\gamma^{-(t+1)}
\mathbb E_{\boldsymbol q(0)}
[W(\boldsymbol q(t+1))\mid{\cal F}_t]
\le
\gamma^{-t}W(\boldsymbol q(t))
=
M_t.
\label{eq:stopped-supermartingale}
\end{align}
Thus $\{M_t\}$ is a nonnegative supermartingale. In particular,
$M_0=W(\boldsymbol q(0))$, and the supermartingale property gives
$\mathbb E_{\boldsymbol q(0)}[M_t]
\le\mathbb E_{\boldsymbol q(0)}[M_0]
=W(\boldsymbol q(0))$.
Since $M_t\ge0$, we may restrict the expectation to the event
$\{\tau_{\cal C}>t\}$. On this event,
$t\wedge\tau_{\cal C}=t$, so
$M_t=\gamma^{-t}W(\boldsymbol q(t))$. Therefore,
\begin{align}
W(\boldsymbol q(0))
&=
\mathbb E_{\boldsymbol q(0)}[M_0]
\ge
\mathbb E_{\boldsymbol q(0)}[M_t]
\ge
\mathbb E_{\boldsymbol q(0)}
\left[
M_t{\bf 1}_{\{\tau_{\cal C}>t\}}
\right]
\nonumber\\
&=
\gamma^{-t}
\mathbb E_{\boldsymbol q(0)}
\left[
W(\boldsymbol q(t))
{\bf 1}_{\{\tau_{\cal C}>t\}}
\right]
\ge
\gamma^{-t}
\mathbb E_{\boldsymbol q(0)}
\left[
{\bf 1}_{\{\tau_{\cal C}>t\}}
\right]
\nonumber\\
&=
\gamma^{-t}
\mathbb P_{\boldsymbol q(0)}(\tau_{\cal C}>t),
\label{eq:geometric-return-derivation}
\end{align}
where the last inequality uses $W(\boldsymbol q(t))\ge1$.
Rearranging \eqref{eq:geometric-return-derivation} yields
\begin{align}
\mathbb P_{\boldsymbol q(0)}(\tau_{\cal C}>t)
&\le
W(\boldsymbol q(0))\gamma^t.
\label{eq:geometric-return}
\end{align}
Thus the hitting time of ${\cal C}$ has a geometric tail, with the
initial-state dependence appearing only through $W(\boldsymbol q(0))$.
For every $r\in(1,\gamma^{-1})$, using
$\mathbb E[r^\tau]
=1+(r-1)\sum_{t\ge0}r^t\mathbb P(\tau>t)$
for an integer-valued nonnegative $\tau$, we further obtain
\begin{align}
\mathbb E_{\boldsymbol q(0)}[r^{\tau_{\cal C}}]
&\le
1+\frac{r-1}{1-r\gamma}W(\boldsymbol q(0)).
\label{eq:return-exponential-moment}
\end{align}

We now prove geometric convergence directly by coupling. Consider two
independent copies $\boldsymbol X(t)$ and $\boldsymbol Y(t)$ of the
chain with kernel $P^\pi$. Define
\begin{align}
\overline W(\boldsymbol x,\boldsymbol y)
&:=
W(\boldsymbol x)+W(\boldsymbol y),
\qquad
\overline\gamma
:=
\frac{1+\gamma}{2},
\qquad
{\cal D}
:=
\left\{
(\boldsymbol x,\boldsymbol y):
\overline W(\boldsymbol x,\boldsymbol y)
\le
\frac{4B}{1-\gamma}
\right\}.
\label{eq:pair-small-set}
\end{align}
The set ${\cal D}$ is finite because
${\cal Q}=\mathbb Z_+^N$ and
$W(\boldsymbol q)\to\infty$ as $\|\boldsymbol q\|_1\to\infty$.
For the product kernel $\overline P=P^\pi \otimes P^\pi$,
Lemma~\ref{geometrical_drift} implies
\begin{align}
\sum_{\vx',\vy'} \overline P \left(\left. \vx', \vy' \right| \vx, \vy \right) \overline W (\vx',\vy')
&\le
\gamma\overline W(\boldsymbol x,\boldsymbol y)+2B
\le
\overline\gamma\,\overline W(\boldsymbol x,\boldsymbol y),
\qquad
(\boldsymbol x,\boldsymbol y)\notin{\cal D}.
\label{eq:pair-drift}
\end{align}
Let
$\tau_{\cal D}
=
\inf\{t\ge0:
(\boldsymbol X(t),\boldsymbol Y(t))\in{\cal D}\}$.
Applying exactly the supermartingale argument in
\eqref{eq:stopped-supermartingale}--\eqref{eq:return-exponential-moment}
to the product chain, for every
$s\in(1,\overline\gamma^{-1})$,
\begin{align}
\mathbb E_{\boldsymbol x,\boldsymbol y}[s^{\tau_{\cal D}}]
&\le
1+
\frac{s-1}{1-s\overline\gamma}
\overline W(\boldsymbol x,\boldsymbol y).
\label{eq:pair-return-exp}
\end{align}

Since ${\cal D}$ is finite, define
\begin{align}
K_{\cal D}
&:=
\max_{(\boldsymbol x,\boldsymbol y)\in{\cal D}}
\left(
\|\boldsymbol x\|_1
\vee
\|\boldsymbol y\|_1
\right),
\qquad
m_{\cal D}
:=
N\lceil K_{\cal D}\rceil,
\qquad
\epsilon_{\cal D}
:=
p_s^{2m_{\cal D}}
>0.
\label{eq:pair-emptying}
\end{align}
Starting from any
$(\boldsymbol x,\boldsymbol y)\in{\cal D}$,
define an \emph{emptying attempt} to be the following
$m_{\cal D}$-slot experiment. At the beginning of the attempt, the
product chain is in ${\cal D}$. Let ${\cal E}$ denote the event that,
during the next $m_{\cal D}$ slots, both copies receive no exogenous
arrivals and all remaining jobs in both copies are
served. We call the attempt \emph{successful} if ${\cal E}$
occurs and \emph{unsuccessful} otherwise.
As discussed in Appendix~\ref{section: prop:well_defined_J: proof}, the proof of Proposition~\ref{prop:well_defined_J},
in each time slot, each copy receives no exogenous arrival and serves one existing job with probability at least $p_s$.
And a successful attempt guarantees that both copies are at $\boldsymbol 0$ no later than the end of the $m_{\cal D}$ slots.
Conditional on the state at
the beginning of the attempt, the independence of the arrival and
potential-service processes gives
\begin{align}
\mathbb P_{\boldsymbol x,\boldsymbol y}({\cal E})
&\ge
\epsilon_{\cal D},
\qquad
(\boldsymbol x,\boldsymbol y)\in{\cal D}.
\label{eq:pair-emptying-probability}
\end{align}
Notice that ${\cal E}$ is only a sufficient event
for simultaneous emptying: the two copies may reach
$(\boldsymbol 0,\boldsymbol 0)$ even when ${\cal E}$ does not occur,
which can only make the coupling time smaller.

If an attempt is unsuccessful and the two copies have not already hit
$(\boldsymbol 0,\boldsymbol 0)$, the product chain need not remain in
${\cal D}$. However, by Lemma 2, during the $m_{\cal D}$ slots each
component can increase in $\ell_1$-norm by at most $2Nm_{\cal D}$.
Thus, at the end of the attempt, the pair belongs to the finite set
\begin{align}
{\cal D}^+
&:=
\left\{
(\boldsymbol x,\boldsymbol y):
\|\boldsymbol x\|_1
\vee
\|\boldsymbol y\|_1
\le
K_{\cal D}+2Nm_{\cal D}
\right\}.
\label{eq:D-plus}
\end{align}
After an unsuccessful attempt, we wait until the product chain next
returns to ${\cal D}$ and then start a new emptying attempt. By
\eqref{eq:pair-return-exp}, for
$s\in(1,\overline\gamma^{-1})$
the exponential moment of this return time is uniformly bounded by
\begin{align}
G_{\cal D}(s)
&:=
\sup_{(\boldsymbol x,\boldsymbol y)\in{\cal D}^+}
\mathbb E_{\boldsymbol x,\boldsymbol y}
[s^{\tau_{\cal D}}]
<\infty,
\qquad
\lim_{s\downarrow1}G_{\cal D}(s)=1.
\label{eq:G-D}
\end{align}
Choose $s_0\in(1,\overline\gamma^{-1})$ sufficiently close to one
so that
\begin{align}
a_{\cal D}
&:=
s_0^{m_{\cal D}}
G_{\cal D}(s_0)
(1-\epsilon_{\cal D})
<1.
\label{eq:a-D}
\end{align}
Let
\begin{align}
\tau_{(\vzero,\vzero)}
&:=
\inf\left\{
t\ge0:
(\boldsymbol X(t),\boldsymbol Y(t))
=
(\boldsymbol 0,\boldsymbol 0)
\right\}.
\label{eq:tau-00}
\end{align}
To make the repeated-attempt argument explicit, suppose first that
$(\boldsymbol X(0),\boldsymbol Y(0))\in{\cal D}$.
Since the finiteness of the exponential moment has not yet been
established, for $T<\infty$ define the truncated quantity
\begin{align}
F_{\cal D}^{(T)}
&:=
\sup_{(\boldsymbol x,\boldsymbol y)\in{\cal D}}
\mathbb E_{\boldsymbol x,\boldsymbol y}
\left[
s_0^{\tau_{(\vzero,\vzero)}\wedge T}
\right].
\label{eq:F-D-truncated}
\end{align}
During the first emptying attempt, either the pair reaches
$(\boldsymbol 0,\boldsymbol 0)$ within $m_{\cal D}$ slots, whose
contribution is at most $s_0^{m_{\cal D}}$, or the attempt is
unsuccessful and simultaneous emptying has not yet occurred.
In the latter case, the probability of failure is at most
$1-\epsilon_{\cal D}$ by \eqref{eq:pair-emptying-probability};
the pair at the end of the attempt lies in ${\cal D}^+$;
the exponential moment of the additional time required to return to
${\cal D}$ is at most $G_{\cal D}(s_0)$; and, after that return, the
strong Markov property bounds the truncated exponential moment of the
remaining time to simultaneous emptying by $F_{\cal D}^{(T)}$.
Therefore,
\begin{align}
F_{\cal D}^{(T)}
&\le
s_0^{m_{\cal D}}
+
s_0^{m_{\cal D}}
(1-\epsilon_{\cal D})
G_{\cal D}(s_0)
F_{\cal D}^{(T)}
=
s_0^{m_{\cal D}}
+
a_{\cal D}F_{\cal D}^{(T)}.
\label{eq:F-D-recursion}
\end{align}
Since $a_{\cal D}<1$, rearranging
\eqref{eq:F-D-recursion} yields, uniformly over $T$,
\begin{align}
F_{\cal D}^{(T)}
&\le
\frac{s_0^{m_{\cal D}}}{1-a_{\cal D}}.
\label{eq:F-D-truncated-bound}
\end{align}
Letting $T\to\infty$ and using monotone convergence gives
\begin{align}
\sup_{(\boldsymbol x,\boldsymbol y)\in{\cal D}}
\mathbb E_{\boldsymbol x,\boldsymbol y}
[s_0^{\tau_{(\vzero,\vzero)}}]
&\le
\frac{s_0^{m_{\cal D}}}{1-a_{\cal D}}
<\infty.
\label{eq:tau-00-from-D}
\end{align}
This recursion is equivalent to summing a geometric series over
repeated unsuccessful attempts, but it explicitly defines an attempt,
what constitutes failure, and what happens after a failure.

We next extend \eqref{eq:tau-00-from-D} to an arbitrary initial pair
$(\boldsymbol x,\boldsymbol y)$. Define
\begin{align}
K_{\cal D}^{(0)}
&:=
\frac{s_0^{m_{\cal D}}}{1-a_{\cal D}}.
\label{eq:K-D-0}
\end{align}
Then \eqref{eq:tau-00-from-D} states that
\begin{align}
\sup_{(\boldsymbol u,\boldsymbol v)\in{\cal D}}
\mathbb E_{\boldsymbol u,\boldsymbol v}
[s_0^{\tau_{(\vzero,\vzero)}}]
&\le
K_{\cal D}^{(0)}.
\label{eq:tau-00-D-K}
\end{align}
For the product chain starting from $(\boldsymbol x,\boldsymbol y)$,
first wait until it enters ${\cal D}$. Since
$(\boldsymbol 0,\boldsymbol 0)\in{\cal D}$, we have
$\tau_{\cal D}\le\tau_{(\vzero,\vzero)}$. At time $\tau_{\cal D}$, the random
state
$(\boldsymbol X(\tau_{\cal D}),
\boldsymbol Y(\tau_{\cal D}))$
belongs to ${\cal D}$. Moreover,
$\tau_{(\vzero,\vzero)}-\tau_{\cal D}$ is the additional time needed to reach
$(\boldsymbol 0,\boldsymbol 0)$ after time $\tau_{\cal D}$.
Therefore, by conditioning on ${\cal F}_{\tau_{\cal D}}$ and applying
the strong Markov property at $\tau_{\cal D}$,
\begin{align}
\mathbb E_{\boldsymbol x,\boldsymbol y}[s_0^{\tau_{(\vzero,\vzero)}}]
&=
\mathbb E_{\boldsymbol x,\boldsymbol y}
\left[
s_0^{\tau_{\cal D}}
s_0^{\tau_{(\vzero,\vzero)}-\tau_{\cal D}}
\right]
\nonumber\\
&=
\mathbb E_{\boldsymbol x,\boldsymbol y}
\left[
s_0^{\tau_{\cal D}}
\mathbb E
\left[
s_0^{\tau_{(\vzero,\vzero)}-\tau_{\cal D}}
\mid
{\cal F}_{\tau_{\cal D}}
\right]
\right]
\nonumber\\
&=
\mathbb E_{\boldsymbol x,\boldsymbol y}
\left[
s_0^{\tau_{\cal D}}
\mathbb E_{
\boldsymbol X(\tau_{\cal D}),
\boldsymbol Y(\tau_{\cal D})}
[s_0^{\tau_{0,0}}]
\right]
\nonumber\\
&\le
K_{\cal D}^{(0)}
\mathbb E_{\boldsymbol x,\boldsymbol y}
[s_0^{\tau_{\cal D}}],
\label{eq:tau-00-strong-Markov}
\end{align}
where the last inequality follows from \eqref{eq:tau-00-D-K} because
$(\boldsymbol X(\tau_{\cal D}),
\boldsymbol Y(\tau_{\cal D}))\in{\cal D}$.
Applying \eqref{eq:pair-return-exp} with $s=s_0$ to the last term in
\eqref{eq:tau-00-strong-Markov} gives
\begin{align}
\mathbb E_{\boldsymbol x,\boldsymbol y}[s_0^{\tau_{(\vzero,\vzero)}}]
&\le
K_{\cal D}^{(0)}
\left[
1+
\frac{s_0-1}{1-s_0\overline\gamma}
\overline W(\boldsymbol x,\boldsymbol y)
\right].
\label{eq:tau-00-intermediate}
\end{align}
Since $W\ge1$,
$\overline W(\boldsymbol x,\boldsymbol y)
=W(\boldsymbol x)+W(\boldsymbol y)\ge2$,
and in particular
$1\le\overline W(\boldsymbol x,\boldsymbol y)$.
Hence \eqref{eq:tau-00-intermediate} implies
\begin{align}
\mathbb E_{\boldsymbol x,\boldsymbol y}[s_0^{\tau_{(\vzero,\vzero)}}]
&\le
K_{(\vzero,\vzero)}\overline W(\boldsymbol x,\boldsymbol y)
=
K_{(\vzero,\vzero)}
\left(
W(\boldsymbol x)+W(\boldsymbol y)
\right),
\label{eq:tau-00-exp}
\end{align}
where the uniform constant $K_{(\vzero,\vzero)}$ can be chosen explicitly as
\begin{align}
K_{(\vzero,\vzero)}
&:=
K_{\cal D}^{(0)}
\left(
1+
\frac{s_0-1}{1-s_0\overline\gamma}
\right)
=
\frac{s_0^{m_{\cal D}}}{1-a_{\cal D}}
\left(
1+
\frac{s_0-1}{1-s_0\overline\gamma}
\right)
<\infty.
\label{eq:K-00}
\end{align}

Run the two copies independently until $\tau_{(\vzero,\vzero)}$. Once both are at
$\boldsymbol 0$, use the same arrivals, potential service variables,
and policy randomization for both copies thereafter. They then remain
identical. If $T_{\rm coup}$ is the resulting coupling time, then
$T_{\rm coup}\le\tau_{(\vzero,\vzero)}$. Thus, with
$\rho=s_0^{-1}\in(0,1)$, Markov's inequality and
\eqref{eq:tau-00-exp} give
\begin{align}
\mathbb P_{\boldsymbol x,\boldsymbol y}(T_{\rm coup}>t)
&\le
K_{(\vzero,\vzero)}
\left(
W(\boldsymbol x)+W(\boldsymbol y)
\right)
\rho^t.
\label{eq:coupling-tail}
\end{align}
For any ${\cal A}\subseteq{\cal Q}$, we have
\begin{align}
\left|
\sum_{\vu \in \gA} \left(P^\pi\right)^t(\boldsymbol x,\vu)
-
\sum_{\vu \in \gA} \left(P^\pi\right)^t(\boldsymbol y,\vu)
\right|
&=
\left|
\mathbb E
\left[
{\bf 1}_{\{\boldsymbol X(t)\in{\cal A}\}}
-
{\bf 1}_{\{\boldsymbol Y(t)\in{\cal A}\}}
\right]
\right|
\nonumber\\
&\le
\mathbb E
\left[
\left|
{\bf 1}_{\{\boldsymbol X(t)\in{\cal A}\}}
-
{\bf 1}_{\{\boldsymbol Y(t)\in{\cal A}\}}
\right|
\right]
\nonumber\\
&\le
\mathbb E
\left[
{\bf 1}_{\{\boldsymbol X(t)\ne\boldsymbol Y(t)\}}
\right]
=
\mathbb P(\boldsymbol X(t)\ne\boldsymbol Y(t))
\nonumber\\
&\le
\mathbb P(T_{\rm coup}>t),
\label{eq:coupling-TV-derivation}
\end{align}
where the first inequality follows from
$|\mathbb E[Z]|\le\mathbb E[|Z|]$, and the second follows because
\begin{align}
\left|
{\bf 1}_{\{\boldsymbol X(t)\in{\cal A}\}}
-
{\bf 1}_{\{\boldsymbol Y(t)\in{\cal A}\}}
\right|
&\le
{\bf 1}_{\{\boldsymbol X(t)\ne\boldsymbol Y(t)\}}.
\label{eq:indicator-coupling-bound}
\end{align}
Indeed, when $\boldsymbol X(t)=\boldsymbol Y(t)$, the two membership
indicators are identical. Taking the supremum of
\eqref{eq:coupling-TV-derivation} over all
${\cal A}\subseteq{\cal Q}$ and using
\eqref{eq:coupling-tail} gives
\begin{align}
\left\|
\vp_\vx^{\pi,t}
-
\vp_\vy^{\pi,t}
\right\|_{\rm TV}
&\le
K_{(\vzero,\vzero)}
\left(
W(\boldsymbol x)+W(\boldsymbol y)
\right)
\rho^t.
\label{eq:pairwise-geometric}
\end{align}

We now compare the distribution of the chain started from
$\boldsymbol q(0)$ directly with the stationary distribution $\vnu^\pi$.
We first focus on $\E_{\vq \in \vnu^\pi}\left[ W(\vq) \right]$.
For the geometric drift condition in Lemma~\ref{geometrical_drift}, taking expectation over $\vq \sim \vnu^\pi$ on both sides gives,
\begin{align}
    \E_{\vq\sim\vnu^\pi}\left[ \E_\pi\left[ W(\vq(t+1)) \right] \right] \le \gamma \E_{\vq\sim\vnu^\pi}\left[ W(\vq) \right] + B.
    \label{eq: expectation_geo_drift}
\end{align}
Since $\vnu^\pi$ is stationary, we have $\E_{\vq\sim\vnu^\pi}\left[ \E_\pi\left[ W(\vq(t+1)) \right] \right] = \E_{\vq\sim\vnu^\pi}\left[ W(\vq) \right]$. Thus, following from~\eqref{eq: expectation_geo_drift}, we have
\begin{align}
    \E_{\vq \in \vnu^\pi}\left[ W(\vq) \right] \le \frac{B}{1-\gamma},
    \label{eq:stationary-W-drift}
\end{align}
uniformly over $\pi\in{\Pi}_H$.
Stationarity also implies
\begin{align}
\sum_{\vy} \nu^\pi(\vy) \vp_\vy^{\pi,t} = \vnu^\pi.
\label{eq:stationary-mixture}
\end{align}
Thus,
\begin{align}
    \|\vp_\vq^{\pi,t} - \vnu^\pi\|_{\rm TV} &= \|\vp_\vq^{\pi,t} - \sum_{\vy} \nu^\pi(\vy) \vp_\vy^{\pi,t}\|_{\rm TV} \le \sum_{\vy} \nu^\pi(\vy) \| \vp_\vq^{\pi,t} - \vp_\vy^{\pi,t} \|_{\rm TV} \nonumber \\
    &\le K_{(\vzero,\vzero)}
        \left(
        W(\boldsymbol q)+ \sum_{\vy} \nu^\pi(\vy) W(\boldsymbol y)
        \right)
        \rho^t \nonumber \\
    &\le K_{(\vzero,\vzero)}
        \left(
        W(\boldsymbol q)+ \frac{B}{1-\gamma}
        \right)
        \rho^t \nonumber \\
    &\le K_{(\vzero,\vzero)}
        \left(
        1+ \frac{B}{1-\gamma}
        \right)
        W(\boldsymbol q)
        \rho^t,
        \label{eq:stationary-coupling-bound}
\end{align}
where the first equality follows from~\eqref{eq:stationary-mixture}, the first inequality follows from triangle inequality, the second inequality holds for~\eqref{eq:pairwise-geometric}, the third inequality holds for~\eqref{eq: expectation_geo_drift}, and the last inequality holds because the Lyapunouv function $W \ge 1$.

Therefore \eqref{eq:geometric-mixing} holds with
\begin{align}
C_{\rm mix}
&:=
K_{(\vzero,\vzero)}
\left(
1+\frac{B}{1-\gamma}
\right),
\qquad
\rho_{\rm mix}
:=
\rho
\in(0,1).
\label{eq:mixing-constants}
\end{align}
All constants above depend only on the common drift constants in
Lemma 5, the network parameters, $H$, and $N$, and not on $\pi$.
Therefore the geometric mixing bound is uniform over
$\pi\in{\Pi}_H$.
\end{proof}

%% file: sections/proofs/lemma_advantage.tex
\begin{lemma} \label{lemma: advantage_bound}
    Suppose Assumption 1 holds. For any policy $\pi \in \Pi_H$, we have,
    \begin{align}\label{ad-bound}
        \max_{\substack{\vq: \|\vq\|_1 \le H \\ \vu \in \gU}} \left| A^\pi (\vq, \vu) \right| 
        \le 
        2 C_{\mix} (H+N) \exp(\kappa (H+N)) \frac{1+\rho_\mix}{\rho_\mix (1 - \rho_\mix)^2}.
    \end{align}
\end{lemma}

\begin{proof}
    According to the definition of the advantage function, we have
    \begin{align} \label{advantage_function_recursion}
    A^\pi(\vq, \vu) =& Q^\pi(\vq, \vu) - V^\pi(\vq) \nonumber \\
    =& c(\vq) - J(\pi) + \sum_{\vq'} P(\vq'|\vq, \vu) V^\pi(\vq') \nonumber \\
    &- \left[ c(\vq) - J(\pi) + \sum_{\vu'} \pi(\vu'|\vq) \sum_{\vq'} P(\vq'|\vq, \vu') V^\pi(\vq') \right] \nonumber \\
    =& \sum_{\vq'} P(\vq'|\vq, \vu) V^\pi(\vq') - \sum_{\vu'} \pi(\vu'|\vq) \sum_{\vq'} P(\vq'|\vq, \vu') V^\pi(\vq').
    \end{align}
    For any state $\vq$ and action $\vu$, define $\vp_\vq^{\vu, \pi, t} = \sum_{\vq' \in \gQ} P(\vq' | \vq, \vu) \vp_{\vq'}^{\pi, t-1}$, representing the $t$-step transition probability of state $\vq$ under kernel $P^\pi$, with the first step action $\vu$.
    By convention, $p_\vq^{\vu, \pi, 0} (\vq') = 1$ if $\vq' = \vq$; $p_\vq^{\vu, \pi, 0} (\vq') = 0$, otherwise.
    Then,~\eqref{advantage_function_recursion} can be rewritten as,
    \begin{align} \label{eq: advantage_function_recursion_1}
        A^\pi(\vq, \vu) = \sum_{\vq'} \left( p_\vq^{\vu,\pi,1}(\vq') - p_\vq^{\pi,1}(\vq') \right) V^\pi(\vq').
    \end{align}
    Since according to the definition, $V^\pi(\vq') = c(\vq') - J(\pi) + \sum_{\vu'} \pi(\vu'|\vq') \sum_{\vq''} P(\vq''|\vq', \vu') V^\pi(\vq'')$,
    substituting into~\eqref{eq: advantage_function_recursion_1} gives,
    \begin{align} \label{eq: advantage_function_recursion_2}
        A^\pi(\vq, \vu) &= \sum_{\vq'} \left( p_\vq^{\vu,\pi,1}(\vq') - p_\vq^{\pi,1}(\vq') \right) \left[ c(\vq') - J(\pi) + \sum_{\vu'} \pi(\vu'|\vq') \sum_{\vq''} P(\vq''|\vq', \vu') V^\pi(\vq'') \right] \nonumber \\
        &= \sum_{\vq'} \left( p_\vq^{\vu,\pi,1}(\vq') - p_\vq^{\pi,1}(\vq') \right) c(\vq') + \sum_{\vq''} \left( p_\vq^{\vu,\pi,2}(\vq'') - p_\vq^{\pi,2}(\vq'') \right) V^\pi(\vq''),
    \end{align}
    where in the last equality, the term related to $J(\pi)$ cancels out because $\sum_{\vq'} p_\vq^{\vu,\pi,1}(\vq') = \sum_{\vq'} p_\vq^{\pi,1}(\vq') = 1$,
    and the term $\sum_{\vq''} \left( p_\vq^{\vu,\pi,2}(\vq'') - p_\vq^{\pi,2}(\vq'') \right) V^\pi(\vq'')$ comes from 
    \begin{align*}
        p_\vq^{\vu,\pi,2}(\vq'') &= \sum_{\vq'} p_\vq^{\vu,\pi,1}(\vq') \sum_{\vu'} \pi(\vu'|\vq')  P(\vq''|\vq', \vu'), \\
        p_\vq^{\pi,2}(\vq'') &= \sum_{\vq'} p_\vq^{\pi,1}(\vq') \sum_{\vu'} \pi(\vu'|\vq') P(\vq''|\vq', \vu').
    \end{align*}
    By substituting the definition of value function into~\eqref{eq: advantage_function_recursion_2} recursively, we have
    \[
    A^\pi (\vq,\vu) =  \sum_{t = 0}^\infty \sum_{\vy \in \gQ} \left( p_\vq^{\vu, \pi, t} (\vy) c(\vy) - p^{\pi, t}_{\vq} (\vy) c(\vy) \right),
    \]
    where we add the term $\vc^\top \vp_\vq^{\vu, \pi, 0}  - \vc^\top \vp^{\pi, 0}_{\vq}$ to the begining of the summation.
    This is acceptable because $\vc^\top \vp_\vq^{\vu, \pi, 0}  - \vc^\top \vp^{\pi, 0}_{\vq} = c(\vq) - c(\vq) = 0$.
    
    For each $t\ge0$, define
    \[
        \gD_{\vq,t}^{\pi, \vu}
        \coloneqq
        \left\{
            \vy\in\gQ:
            p^{\pi, t}_{\vq}(\vy)>0
            \text{ or }
            p_\vq^{\vu, \pi, t}(\vy)>0
        \right\},
    \]
    which implies that the probability that the system is in a state not in $\gD_{\vq,t}^{\pi, \vu}$ at time $t$ is zero, in other words, $p_\vq^{\vu, \pi, t}(\vy)=p^{\pi, t}_{\vq}(\vy)=0$ for $\vy\not\in\gD_{\vq,t}^{\pi, \vu}$. 
    Thus, we have 
    \begin{align} \label{finite-A}
        | A^\pi (\vq,\vu) | = \left| \sum_{t = 0}^\infty \sum_{\vy \in \gD_{\vq,t}^{\pi, \vu}} \left( p_\vq^{\vu, \pi, t} (\vy) c(\vy) - p^{\pi, t}_{\vq} (\vy) c(\vy) \right) \right|.
    \end{align}
    According to Lemma~\ref{lemma: one_step_bound}, we have, 
    \begin{align} \label{prop:Dqt}
        \max_{\vy\in\gD_{\vq,t}^{\pi, \vu}} c(\vy) \le c(\vq)+Nt,
    \end{align}
    since $c(\vq)\le\|\vq\|_1$.
    Next we add and subtract the distribution $\nu^\pi(\vy)$ in~\eqref{finite-A} to obtain
    \begin{align}
        | A^\pi (\vq,\vu) | &= 
        \left| \sum_{t = 0}^\infty \sum_{\vy \in \gD_{\vq,t}^{\pi, \vu}} \left( p_\vq^{\vu, \pi, t} (\vy) c(\vy) - \nu^\pi (\vy) c(\vy) + \nu^\pi (\vy) c(\vy) - p^{\pi, t}_{\vq} (\vy) c(\vy) \right) \right| \nonumber \\
        &\le 
        \sum_{t=0}^\infty 
        \underbrace{
        \left| \sum_{\vy \in \gD_{\vq,t}^{\pi, \vu}} \left( p_\vq^{\vu, \pi, t} (\vy) c(\vy) - \nu^\pi (\vy) c(\vy) \right) \right|
        }_{\mathsf{A}_{1,t}} 
        + 
        \sum_{t=0}^\infty 
        \underbrace{
        \left| \sum_{\vy \in \gD_{\vq,t}^{\pi, \vu}} \left( p^{\pi, t}_{\vq} (\vy) c(\vy) - \nu^\pi (\vy) c(\vy) \right) \right|
        }_{\mathsf{A}_{2,t}}, \label{ad-decompose}
    \end{align} where the last inequality follows from the triangle inequality.

    We first analyze $\mathsf{A}_{2,t}$ in~\eqref{ad-decompose}. We have
    \begin{align}
        \mathsf{A}_{2,t} 
        &\le 
        \left(\max_{\vy \in \gD_{\vq,t}^{\pi, \vu}} c(\vy)\right)  \sum_{\vy \in \gD_{\vq,t}^{\pi, \vu}} \left| p^{\pi, t}_{\vq}(\vy) - \nu^\pi(\vy) \right|
        \overset{(a)}{\le} 
        ( c(\vq) + Nt ) \left\| \vp^{\pi, t}_{\vq} - \vnu^\pi \right\|_1 
        \overset{(b)}{\le} 
        ( c(\vq) + Nt ) \cdot 2 C_\mix W (\vq) \rho_\mix^t \nonumber\\
        &\le
        2 C_\mix \left( \max_{\vq: \|\vq\|_1 \le H} c(\vq) W(\vq) \right) \rho_\mix^t + 2 C_\mix N \left( \max_{\vq: \|\vq\|_1 \le H} W(\vq) \right) t \rho_\mix^t \nonumber \\
        &\overset{(c)}{\le}
        2 C_\mix H \exp(\kappa H) \rho_\mix^t + 2 C_\mix N \exp(\kappa H) t \rho_\mix^t, \label{final-t-bound}
    \end{align}
    where Inequality $(a)$ is due to the property of $\gD_{\vq,t}^{\pi, \vu}$ stated in~\eqref{prop:Dqt}, Inequality $(b)$ follows from Lemma~\ref{lemma: Uniform geometric mixing}, 
    and Inequality $(c)$ follows from $c(\vq) = \|\vq\|_1$ and the Lyapunov function $W$ defined in Lemma~\ref{geometrical_drift}. 
    Thus, we have,
    \begin{align} \label{2-bound}
        \sum_{t=0}^\infty 
        \mathsf{A}_{2,t} 
        \le 
        &2 C_\mix H \exp(\kappa H) \sum_{t=0}^\infty \rho_\mix^t + 2 C_\mix N \exp(\kappa H) \sum_{t=0}^\infty t \rho_\mix^t \nonumber \\
        \le  
        &2 C_\mix H \exp(\kappa H) \frac{1}{1-\rho_\mix} + 2 C_\mix N \exp(\kappa H) \frac{\rho_\mix}{(1-\rho_\mix)^2}.
    \end{align}

    We now consider $\mathsf{A}_{1,t}$ in~\eqref{ad-decompose}. For $t\ge 1,$ we have,
    \begin{align}
        \mathsf{A}_{1,t} 
        &= 
        \left| \sum_{\vy \in \gD_{\vq,t}^{\pi, \vu}} \sum_{\vq'} P(\vq' | \vq, \vu) \left( p_{\vq'}^{\pi, t-1} (\vy) - \nu^\pi(\vy) \right) c(\vy) \right| \nonumber\\
        &\overset{(a)}{\le} 
        ( c(\vq) + Nt ) \sum_{\vq'} P(\vq' | \vq, \vu) \sum_{\vy \in \gD_{\vq,t}^{\pi, \vu}} \left| p_{\vq'}^{\pi, t-1} (\vy) - \nu^\pi(\vy) \right| \nonumber \\ 
        &\overset{(b)}{\le} 
        ( c(\vq) + Nt ) \max_{\vq': P(\vq' | \vq, \vu) > 0} \left\| \vp_{\vq'}^{\pi, t-1} - \vnu^\pi \right\|_1 \nonumber\\
        &\overset{(c)}{\le} 
        ( c(\vq) + Nt ) \cdot 2 C_\mix \max_{\vq': \left| \|\vq'\|_1 - \|\vq\|_1 \right| \le N} W (\vq') \rho_\mix^{t - 1} \nonumber \\
        &\overset{(d)}{\le}
        2 C_\mix H \exp (\kappa(H+N)) \rho_\mix^{t - 1} + 2 C_\mix N \exp (\kappa(H+N)) t \rho_\mix^{ t - 1 }. \label{final-2-bound}
    \end{align}
    Inequality $(a)$ is due to the property of $\gD_{\vq,t}^{\pi, \vu}$ stated in~\eqref{prop:Dqt}, Inequality $(b)$ holds because 
    \[
    \sum_{\vy \in \gD_{\vq,t}^{\pi, \vu}} \left| p_{\vq'}^{\pi, t-1} (\vy) - \nu^\pi(\vy) \right| \leq \left\| \vp_{\vq'}^{\pi, t-1} - \vnu^\pi \right\|_1
    \] 
    and $\sum_{\vq'} P(\vq' | \vq, \vu) = 1$ for any $(\vq,\vu),$ inequality $(c)$ results from Lemma~\ref{lemma: Uniform geometric mixing} and Lemma~\ref{lemma: one_step_bound}, and Inequality $(d)$ follows from $c(\vq) = \|\vq\|_1$ and the Lyapunov function $W$ defined in Lemma~\ref{geometrical_drift}. 
    Note that the bound \eqref{final-2-bound} holds for $t=0$ as well. Summarizing $\mathsf{A}_{1,t}$ over $t$ yields 
    \begin{align} \label{1-bound}
        \sum_{t=0}^\infty 
        \mathsf{A}_{1,t}
        \leq&2 C_\mix H \exp (\kappa(H+N)) \frac{1}{ \rho_\mix (1 - \rho_\mix) } + 2 C_\mix N \exp (\kappa(H+N)) \frac{1}{ (1 - \rho_\mix)^2 }.
    \end{align}
    Substituting \eqref{1-bound} and \eqref{2-bound} into \eqref{ad-decompose}, we have 
    \begin{align*}
    | A^\pi (\vq, \vu) | 
    \le&
    2 C_\mix H \exp (\kappa(H+N)) \frac{1}{ \rho_\mix (1 - \rho_\mix) } + 2 C_\mix N \exp (\kappa(H+N)) \frac{1}{ (1 - \rho_\mix)^2 } \nonumber \\
    &+ 2 C_\mix H \exp(\kappa H) \frac{1}{1-\rho_\mix} + 2 C_\mix N \exp(\kappa H) \frac{\rho_\mix}{(1-\rho_\mix)^2} \nonumber \\
    \le& 2 C_\mix H \exp (\kappa(H+N)) \frac{1-\rho_\mix^2}{ \rho_\mix (1 - \rho_\mix)^2 } + 2 C_\mix N \exp (\kappa(H+N)) \frac{\rho_\mix + \rho_\mix^2}{ \rho_\mix(1 - \rho_\mix)^2 } \nonumber \\
    \le& 2 C_\mix H \exp (\kappa(H+N)) \frac{1 + \rho_\mix}{ \rho_\mix (1 - \rho_\mix)^2 } + 2 C_\mix N \exp (\kappa(H+N)) \frac{1 + \rho_\mix}{ \rho_\mix(1 - \rho_\mix)^2 } \nonumber \\
    =&2 C_\mix (H+N) \exp(\kappa(H+N)) \frac{1+\rho_\mix}{\rho_\mix (1-\rho_\mix)^2},
    \end{align*}
    where the second inequality holds by bounding $\exp(\kappa H)$ with $\exp(\kappa(H+N))$,
    the third inequality holds because $1-\rho_\mix^2 \le 1+\rho_\mix$ and $\rho_\mix+\rho_\mix^2 \le 1+\rho_\mix$.
    This completes the proof.
\end{proof}

%% file: sections/proofs/lemma_stationarydifference.tex
Lemma~\ref{lemma: pre stationary distribution difference} shows a recursion form of the stationary distribution difference, which is the key of bounding it.

\begin{lemma} \label{lemma: pre stationary distribution difference}
    Suppose Assumption~\ref{assumption:stability-condition} holds.
    Consider any two policies $\pi, \pi' \in \Pi_H$. We have,
    \begin{align} \label{eq: sdd_geo}
        \vnu^{\pi'} - \vnu^\pi = \sum_{n=0}^\infty((\mP^\pi)^n)^\top (\mP^{\pi'} - \mP^\pi)^\top \vnu^{\pi'}.
    \end{align}
\end{lemma}

\begin{proof}
    Consider any two policies $\pi, \pi' \in \Pi_H$.
    We construct the connection between the stationary distribution difference and the transition kernel difference. 
    By the global balance equation, for any policy $\pi$, recall that $\mP^\pi$ is the one-step transition matrix under policy $\pi$, and we have $\vnu^\pi = (\mP^\pi)^\top \vnu^\pi$. The difference in the stationary distribution is therefore:
    \begin{align*}
        \vnu^{\pi'} - \vnu^\pi =& (\mP^{\pi'})^\top \vnu^{\pi'} - (\mP^\pi)^\top \vnu^\pi = (\mP^{\pi'} - \mP^\pi)^\top \vnu^{\pi'} + (\mP^\pi)^\top (\vnu^{\pi'} -\vnu^\pi).
    \end{align*}
    Then, we can continue to roll out the stationary distribution difference using the one-step rollout equality until the $N$-th time as follows:
    \begin{align*}
        \vnu^{\pi'} - \vnu^\pi = \sum_{n=0}^N((\mP^\pi)^n)^\top (\mP^{\pi'} - \mP^\pi)^\top \vnu^{\pi'} + ((\mP^\pi)^{N+1})^\top (\vnu^{\pi'} -\vnu^\pi),
    \end{align*}
    with the convention that any squared matrix raised to the power of zero is the identity matrix.
    Since the Markov chain induced by policy $\pi$ is positive recurrent, irreducible, and aperiodic, the $N$-step state distribution converges to the stationary distribution as $N$ increase to infinity. Therefore, we have:
    \begin{align*}
        \lim_{N\to \infty} ((\mP^\pi)^{N+1})^\top (\vnu^{\pi'} -\vnu^\pi) =\left(\lim_{N\to \infty} ((\mP^\pi)^{N+1})^\top \vnu^{\pi'}\right) -\vnu^\pi= 0.
    \end{align*}
    And thus the stationary distribution difference is represented as the infinite sum:
    \begin{align*}
        \vnu^{\pi'} - \vnu^\pi = \sum_{n=0}^\infty((\mP^\pi)^n)^\top (\mP^{\pi'} - \mP^\pi)^\top \vnu^{\pi'}.
    \end{align*}
\end{proof}

Now we are prepared to bound the stationary distribution difference.

\begin{lemma} \label{lemma: stationary distribution difference}
    Suppose Assumption~\ref{assumption:stability-condition} holds.
    For any two policies $\pi, \pi' \in \Pi_H$, we have
    \[
    \| \vnu^{\pi'} - \vnu^\pi \|_1 \le \frac{2 C_{\mix} \exp(\kappa (H+N))}{1 - \rho_{\mix}}  \max_{\vq} \sum_{\vu \in \gU} | \pi'(\vu | \vq)  - \pi(\vu | \vq) |.
    \]
\end{lemma}

\begin{proof}
    By Lemma~\ref{lemma: pre stationary distribution difference}, we have:
    \begin{align} \label{eq: previous lemma 1-norm}
        \left\| \vnu^{\pi'} - \vnu^\pi \right\|_1 = \left\|\sum_{n=0}^\infty((\mP^\pi)^n)^\top (\mP^{\pi'} - \mP^\pi)^\top \vnu^{\pi'}\right\|_1
        \leq \sum_{n=0}^\infty\left\|((\mP^\pi)^n)^\top (\mP^{\pi'} - \mP^\pi)^\top \vnu^{\pi'}\right\|_1.
    \end{align}
    For simplicity, let $\vmu^{\pi', \pi} = (\mP^{\pi'} - \mP^\pi)^\top \vnu^{\pi'}$ and $\mu^{\pi', \pi}(\vq)$ be its element corresponding to state $\vq$.
    We can express the vector inside the $1$-norm as follows:
    \begin{align*}
        ((\mP^\pi)^n)^\top (\mP^{\pi'} - \mP^\pi)^\top \vnu^{\pi'} 
        =& ((\mP^\pi)^n)^\top \vmu^{\pi', \pi} = \sum_{\vq \in \gQ} \mu^{\pi', \pi}(\vq) \vp^{\pi, n}_{\vq}\\
        =& \sum_{\vq \in \gQ} \mu^{\pi', \pi}(\vq) (\vp^{\pi, n}_{\vq} - \vnu^\pi) + \vnu^\pi \sum_{\vq \in \gQ} \mu^{\pi', \pi}(\vq) .
    \end{align*}
    For the second term, notice that $\vmu^{\pi', \pi} = (\mP^{\pi'} - \mP^\pi)^\top \vnu^{\pi'}$ is the difference between two next-state distributions, i.e., $(\mP^{\pi'})^\top \vnu^{\pi'}$ and $(\mP^\pi)^\top \vnu^{\pi'}$, so the sum of elements in $\vmu^{\pi', \pi}$ must equal $0$, since the sum of elements in both distributions is $1$. So we have: 
    \begin{align*}
        \left\|((\mP^\pi)^n)^\top (\mP^{\pi'} - \mP^\pi)^\top \vnu^{\pi'}\right\|_1 
        = \left\|\sum_{\vq \in \gQ} \mu^{\pi', \pi}(\vq) (\vp^{\pi, n}_{\vq} - \vnu^\pi)\right\|_1 
        \leq \sum_{\vq \in \gQ} |\mu^{\pi', \pi}(\vq)| \left\|\vp^{\pi, n}_{\vq} - \vnu^\pi\right\|_1
    \end{align*}
    We claim that under our queueing model and for any policy $\pi$ and $\pi'$ that belongs to $\Pi_H$, when a state $\vq$ satisfies $\|\vq\|_1 > H+N$, we have $\vmu^{\pi',\pi}(\vq)=0$. To see this, we first write the definition of $\vmu^{\pi',\pi}(\vq)$ as follows:
    \begin{align}\label{eq:vq-def}
        \vmu^{\pi',\pi}(\vq) = \sum_{\vq' \in \gQ} \nu^{\pi'}(\vq') \left( P^{\pi'} (\vq | \vq') - P^{\pi} (\vq | \vq') \right).
    \end{align}
    By Lemma~\ref{lemma: one_step_bound}, for any policy in one time slot, the $1$-norm of the state vector $\vq$ can only change by $N$.  
    Therefore, consider a fixed state $\vq$ such that $\|\vq\|_1 > H+N$, for any state $\vq'$ such that $\|\vq'\|_1 \leq H$, we have $P^{\pi'} (\vq | \vq') = P^{\pi'} (\vq | \vq') = 0$. For $\vq'$ such that $\|\vq'\|_1 > H$, both policies $\pi'$ and $\pi$ uses the back-pressure action $\vu_{\bp}(\vq')$, and therefore $P^{\pi'} (\vq | \vq') = P^{\pi} (\vq | \vq')$. As a consequence, for all $\vq'$, we have $P^{\pi'} (\vq | \vq') - P^{\pi} (\vq | \vq') = 0$. According to~\eqref{eq:vq-def}, this means for any state $\vq$ such that $\|\vq\|_1 > H+N$, we have $\vmu^{\pi',\pi}(\vq)=0$. Therefore, we can further bound each term in the $1$-norm of the stationary distribution difference as:
    \begin{align} \label{eq: previous lemma 1-norm per term}
        \left\|((\mP^\pi)^n)^\top (\mP^{\pi'} - \mP^\pi)^\top \vnu^{\pi'}\right\|_1 \leq& \sum_{\|\vq\|_1 \leq H+N} |\mu^{\pi', \pi}(\vq)| \left\|\vp^{\pi, n}_{\vq} - \vnu^\pi\right\|_1 \nonumber\\
        \leq &   \sum_{\|\vq\|_1 \leq H+N} |\mu^{\pi', \pi}(\vq)|  \max_{\|\vq\|_1 \leq H+N}\left\|\vp^{\pi, n}_{\vq} - \vnu^\pi\right\|_1 \nonumber\\
        \leq&  \|\vmu^{\pi', \pi}\|_1 \max_{\|\vq\|_1 \leq H+N}\left\|\vp^{\pi, n}_{\vq} - \vnu^\pi\right\|_1.
    \end{align}
    By Lemma~\ref{lemma: Uniform geometric mixing}, for any policy $\pi$, the induced Markov chain is uniformly geometric mixing, and therefore, we have:
    \begin{align} \label{eq: previous lemma 1-norm bound}
        \max_{\|\vq\|_1 \leq H+N}\left\|\vp^{\pi, n}_{\vq} - \vnu^\pi\right\|_1 \leq 2 C_{\rm mix} \rho_{\rm mix}^n \max_{\|\vq\|_1 \leq H+N} W(\vq)  \leq 2 C_{\rm mix} \rho_{\rm mix}^n \exp(\kappa (H+N)),
    \end{align}
    where $W(\vq) = \exp(\kappa \|\vq\|_2)$ is the exponential Lyapunov function, $C_{\mix}$ and $\rho_{\rm mix}$ are the geometric mixing constants defined in Lemma~\ref{lemma: Uniform geometric mixing}, and $\kappa$ is the Lyapunov function constant defined in Lemma~\ref{geometrical_drift}. The last equality
    implicitly used the fact that the $2$-norm is less than the $1$-norm and thus bounded by $H+N$. Then,
    we can upper bound the difference in stationary distribution as:
    \begin{align}
        \left\| \vnu^{\pi'} - \vnu^\pi \right\|_1 
        \leq& \sum_{n=0}^\infty \|\vmu^{\pi', \pi}\|_1 \max_{\|\vq\|_1 \leq H+N}\left\|\vp^{\pi, n}_{\vq} - \vnu^\pi\right\|_1 
        \leq 2 \|\vmu^{\pi', \pi}\|_1 C_{\rm mix} \exp(\kappa (H+N)) \sum_{n=0}^\infty \rho_{\rm mix}^n \nonumber\\
        \leq& \frac{2 C_{\rm mix} \exp(\kappa (H+N))}{1-\rho_{\mix}} \|\vmu^{\pi', \pi}\|_1,
        \label{eq: d-diff-bounded-by-nu}
    \end{align}
    where the first inequality holds by substituting~\eqref{eq: previous lemma 1-norm per term} into~\eqref{eq: previous lemma 1-norm},
    and the second inequality follows from~\eqref{eq: previous lemma 1-norm bound}.
    The rest is to show that $\|\vmu^{\pi', \pi}\|_1$ is bounded by the policy difference $\|\pi' - \pi\|_{1,\infty}$. We first use the definition of $\vmu^{\pi', \pi}$ in~\eqref{eq:vq-def} and have:
    \begin{align*}
        \|\vmu^{\pi', \pi}\|_1 =& \sum_{\vq} \left|\sum_{\vq'} \nu^{\pi'}(\vq') \left( P^{\pi'} (\vq | \vq') - P^{\pi} (\vq | \vq') \right)\right| 
        \leq \sum_{\vq'} \nu^{\pi'}(\vq')  \sum_{\vq}\left| P^{\pi'} (\vq | \vq') - P^{\pi} (\vq | \vq') \right| 
    \end{align*}
    By the definition of the transition kernel under any policy, we have:
    \begin{align*}
        \sum_{\vq}\left| P^{\pi'} (\vq | \vq') - P^{\pi} (\vq | \vq') \right| =& \sum_{\vq}\left| \sum_{\vu} (\pi'(\vu|\vq') -\pi(\vu|\vq'))  P(\vq | \vq', \vu)\right|\\
        \leq& \sum_{\vu} \left|\pi'(\vu|\vq') -\pi(\vu|\vq')\right|  \sum_{\vq}P(\vq | \vq', \vu)\\
        =& \sum_{\vu} \left|\pi'(\vu|\vq') -\pi(\vu|\vq')\right|.
    \end{align*}
    Substituting this identity back, we obtain:
    \begin{align}
        \|\vmu^{\pi', \pi}\|_1 \leq \sum_{\vq' \in \gQ} \nu^{\pi'}(\vq') \sum_{\vu} \left|\pi'(\vu|\vq') -\pi(\vu|\vq')\right| \leq \max_{\vq'} \sum_{\vu} \left|\pi'(\vu|\vq') -\pi(\vu|\vq'))\right|.
        \label{eq: nu-bound}
    \end{align}
    Substituting~\eqref{eq: nu-bound} back to~\eqref{eq: d-diff-bounded-by-nu} completes the proof.
\end{proof}

%% file: sections/proofs/prop_well_defined_value_proof.tex
\propwelldefinedvalue*

\begin{proof}
    According to the definition of the (relative) value function, we have
    \begin{align}
        \left| V^\pi(\vq) \right| &= \left| \sum_{t=0}^\infty \left(\E_\pi[c(\vq(t))|\vq(0)=\vq]-\E_\pi[c(\vq(t))|\vq(0)=0]\right) \right| \nonumber \\
        &= \left| \sum_{t=0}^\infty \sum_{\vq' \in \gQ} c(\vq') \left( p_\vq^{\pi,t}(\vq') - p_\vzero^{\pi,t}(\vq') \right) \right| \nonumber \\
        &= \left| \sum_{t=0}^\infty \sum_{\vq': \| \vq' \|_1 \le \|\vq\|_1 + Nt} c(\vq') \left( p_\vq^{\pi,t}(\vq') - p_\vzero^{\pi,t}(\vq') \right) \right|,
    \end{align}
    where the last equality holds because according to Lemma~\ref{lemma: one_step_bound}, state $\vq'$ such that $\| \vq' \|_1 > \| \vq \|_1 + Nt$ is not reachable from either from $\vq$ or from $\vzero$ within $t$ time slots, so $p_\vq^{\pi,t}(\vq') = p_\vzero^{\pi,t}(\vq') = 0$.
    Note that $c(\vq') \leq \|\vq' \|_1$, we have
    \begin{align}
        \left| V^\pi(\vq) \right| &\le \sum_{t=0}^\infty \left( \|\vq\|_1 + Nt \right) \sum_{\vq': \| \vq' \|_1 \le \|\vq\|_1 + Nt} \left| p_\vq^{\pi,t}(\vq') - p_\vzero^{\pi,t}(\vq') \right| \nonumber \\
        &= \sum_{t=0}^\infty \left( \|\vq\|_1 + Nt \right) \sum_{\vq'} \left| p_\vq^{\pi,t}(\vq') - p_\vzero^{\pi,t}(\vq') \right| \nonumber \\
        &= \sum_{t=0}^\infty \left( \|\vq\|_1 + Nt \right) \| \vp_\vq^{\pi,t} - \vp_\vzero^{\pi,t} \|_1,
    \end{align}
    where the inequality is due to the triangle inequality
    and the first equality holds still because $p_\vq^{\pi,t}(\vq') = p_\vzero^{\pi,t}(\vq') = 0$ if $\| \vq' \|_1 > \| \vq \|_1 + Nt$. 
   Adding and subtracting the stationary distribution $\vnu^\pi$ results in 
    \begin{align} \label{eq: finite-V}
        \left| V^\pi(\vq) \right| &\le \sum_{t=0}^\infty \left( \|\vq\|_1 + Nt \right) \| \vp_\vq^{\pi,t} - \vnu^\pi + \vnu^\pi - \vp_\vzero^{\pi,t} \|_1 \nonumber \\
        &\le \sum_{t=0}^\infty \left( \|\vq\|_1 + Nt \right) \left( \| \vp_\vq^{\pi,t} - \vnu^\pi \|_1 + \| \vnu^\pi - \vp_\vzero^{\pi,t} \|_1 \right) \nonumber \\
        &\le 2\sum_{t=0}^\infty \left( \|\vq\|_1 + Nt \right) C_\mix (W(\vq)+W(\vzero)) \rho_\mix^t \nonumber \\
        &= 2\|\vq\|_1 C_\mix (W(\vq)+W(\vzero)) \frac{1}{1-\rho_\mix} + 2N C_\mix (W(\vq)+W(\vzero)) \frac{\rho_\mix}{(1-\rho_\mix)^2},
    \end{align}
    where the second inequality is due to the triangle inequality,
    and the third inequality holds according to the geometric mixing property proved in Lemma~\ref{lemma: Uniform geometric mixing}. 
    
    To prove the the Bellman equation (or the Poisson equation), according to the definition of the (relative) value function, we have
    \begin{align*}
        V^\pi(\vq) &= \lim_{T\rightarrow\infty} \sum_{t=0}^T \left(\E_\pi[c(\vq(t))|\vq(0)=\vq]-\E_\pi[c(\vq(t))|\vq(0)=0]\right) \\
        &= \lim_{T\rightarrow\infty} \left( c(\vq) - \E_\pi[c(\vq(T))|\vq(0)=0] + \sum_{t=1}^T \E_\pi[c(\vq(t))|\vq(0)=\vq] - \sum_{t=0}^{T-1} \E_\pi[c(\vq(t))|\vq(0)=0] \right) \\
        &= c(\vq) - J(\pi) + \sum_{\vq'\in\gQ} P^\pi(\vq'|\vq) \lim_{T\rightarrow\infty} \sum_{t=0}^{T-1} \left(\E_\pi[c(\vq(t))|\vq(0)=\vq']-\E_\pi[c(\vq(t))|\vq(0)=0]\right) \\
        &= c(\vq) - J(\pi) + \sum_{\vq'\in\gQ} P^\pi(\vq'|\vq) V^\pi(\vq'),
    \end{align*}
    where the third equality holds because $\lim_{T\rightarrow\infty} \E_\pi[c(\vq(T))|\vq(0)=0] = J(\pi)$, and conditioning on $\vq(0)=\vq$, $\vq(1)$ is sampled from $P^\pi(\cdot|\vq)$.
    This completes the proof.
\end{proof}

%% file: sections/proofs/thm_periteration.tex
\begin{lemma}[Paley–Zygmund inequality, Lemma 19,~\cite{Paley_Zygmund_1932}] \label{lemma: Paley–Zygmund inequality}
    Let $\rz \ge 0$ be a random variable with $\E[\rz^2] < \infty$, and let $\iota \in [0,1]$, then
    \[
    \sP(\rz > \iota\E[\rz]) \ge (1 - \iota)^2 \frac{\E[\rz]^2}{\E[\rz^2]}.
    \]
\end{lemma}

\periterationthm*

\begin{proof}
Since $\vtheta_k' = \vtheta_k + \beta \vv_k$, and $J(\pi_{H,\vtheta})$ is $L_J$-smooth as shown in Lemma~\ref{lemma: smoothness}, we have
\begin{align}
   \beta \nabla_\vtheta J(\pi_{H,\vtheta_k})^\top \vv_k - \frac{L_J}{2} \beta^2 \|\vv_k\|_2^2\le J(\pi_{H,\vtheta_k'}) - J(\pi_{H,\vtheta_k})  \le \beta \nabla_\vtheta J(\pi_{H,\vtheta_k})^\top \vv_k +\frac{L_J}{2} \beta^2 \|\vv_k\|_2^2.
\end{align}
A sufficient condition that the policy performance improves by at least $\beta^2 L_J d$ after one iteration of Algorithm \ref{alg:zeroth-order}, i.e., $J(\pi_{H,\vtheta_k'}) < J(\pi_{H,\vtheta_k}) - \beta^2 L_J d$, 
is 
\begin{align} \label{eq: improve guarantee}
    \nabla_\vtheta J(\pi_{H,\vtheta_k})^\top \vv_k < -\frac{L_J \beta}{2} \|\vv_k\|_2^2 - \beta L_J d = -\frac{L_J \beta}{2} d - \beta L_J d = -\frac{3 \beta L_J d}{2},
\end{align}
where the first equality holds because the perturbation of each coordinate is sampled from a Rademacher distribution, so $v_{k,i}^2 = 1$ for all $i$ and $\|\vv_k\|_2^2 = \sum_{i = 1}^d v_{k,i}^2 = d$. Fixing $\vtheta_k$,  note that the distribution of $\nabla_\vtheta J(\pi_{H,\vtheta_k})^\top \vv_k$ is symmetric around $0$ since each coordinate $v_{k,i}$ is $1$ or $-1$ equally likely, so we have
\begin{align} \label{eq: symmetric}
    \sP \left( \nabla_\vtheta J(\pi_{H,\vtheta_k})^\top \vv_k < -\frac{3 \beta L_J d}{2} \right) &= 
    \sP \left( \nabla_\vtheta J(\pi_{H,\vtheta_k})^\top \vv_k > \frac{3 \beta L_J d}{2} \right) \nonumber \\
    &= 
    \frac{1}{2} \sP \left( \left( \nabla_\vtheta J(\pi_{H,\vtheta_k})^\top \vv_k \right)^2 > \frac{9 \beta^2 L_J^2 d^2}{4} \right).
\end{align}

The e expectation of $\left( \nabla_\vtheta J(\pi_{H,\vtheta_k})^\top \vv_k \right)^2$ satisfies
\begin{align} \label{eq: 2 moment}
    \E \left[ \left( \nabla_\vtheta J(\pi_{H,\vtheta_k})^\top \vv_k \right)^2 \right] = \E \left[ \sum_{i = 1}^d [\nabla_\vtheta J(\pi_{H,\vtheta_k})]_i^2 v_{k,i}^2 \right] = \sum_{i = 1}^d [\nabla_\vtheta J(\pi_{H,\vtheta_k})]_i^2 = \| \nabla_\vtheta J(\pi_{H,\vtheta_k}) \|_2^2,
\end{align}
where the first equality holds because for all $i\neq j$, $v_{k,i}$ and $v_{k,j}$ are independent and zero mean random variables, so $\E\left[ [\nabla_\vtheta J(\pi_{H,\vtheta_k})]_i [\nabla_\vtheta J(\pi_{H,\vtheta_k})]_j v_{k,i} v_{k,j} \right] = 0$, and the second equality holds due to $v_{k,i}$ is either $1$ or $-1$.
According to the assumption of the theorem, we have $\|\nabla_\vtheta J(\pi_{H, \vtheta_k})\|_2 \geq 2 \beta L_J d$.
Let $\eta = 3/4$, by the Paley–Zygmund inequality in Lemma~\ref{lemma: Paley–Zygmund inequality}, we have:
\begin{align}\label{eq:paley-zygmund}
    \sP \left( \left( \nabla_\vtheta J(\pi_{H,\vtheta_k})^\top \vv_k \right)^2 > \frac{9 \beta^2 L_J^2 d^2}{4} \right) 
    \geq& \sP \left( \left( \nabla_\vtheta J(\pi_{H,\vtheta_k})^\top \vv_k \right)^2 > \eta^2 \| \nabla_\vtheta J(\pi_{H,\vtheta_k}) \|_2^2 \right) \nonumber\\
    =& \sP \left( \left( \nabla_\vtheta J(\pi_{H,\vtheta_k})^\top \vv_k \right)^2 > \eta^2 \E \left[ \left( \nabla_\vtheta J(\pi_{H,\vtheta_k})^\top \vv_k \right)^2 \right] \right) \nonumber\\
    \geq & (1 - \eta^2)^2 \frac{\E \left[ \left( \nabla_\vtheta J(\pi_{H,\vtheta_k})^\top \vv_k \right)^2 \right]^2}{\E \left[ \left( \nabla_\vtheta J(\pi_{H,\vtheta_k})^\top \vv_k \right)^4 \right]}.
\end{align}
The fourth moment of $\nabla_\vtheta J(\pi_{H,\vtheta_k})^\top \vv_k$ satisfies
\begin{align*}
    \E \left[ \left( \nabla_\vtheta J(\pi_{H,\vtheta_k})^\top \vv_k \right)^4 \right]
    =&
    \E \left[ \sum_{i = 1}^d [\nabla_\vtheta J(\pi_{H,\vtheta_k})]_i^4 v_{k,i}^4 \right] + 3 \E \left[ \sum_{i \neq j} [\nabla_\vtheta J(\pi_{H,\vtheta_k})]_i^2 v_{k,i}^2 [\nabla_\vtheta J(\pi_{H,\vtheta_k})]_j^2 v_{k,j}^2  \right]\\
    =& \sum_{i = 1}^d [\nabla_\vtheta J(\pi_{H,\vtheta_k})]_i^4 + 3\sum_{i \neq j} [\nabla_\vtheta J(\pi_{H,\vtheta_k})]_i^2 [\nabla_\vtheta J(\pi_{H,\vtheta_k})]_j^2,
\end{align*}
where in the first equality, any monomial containing at least one factor \(v_{k,i}\) raised to an odd power has zero expectation. Since both terms are positive, we obtain:
\begin{align}\label{eq: 4 moment}
    \E \left[ \left( \nabla_\vtheta J(\pi_{H,\vtheta_k})^\top \vv_k \right)^4 \right] \leq & 3\left(\sum_{i = 1}^d [\nabla_\vtheta J(\pi_{H,\vtheta_k})]_i^4 + \sum_{i \neq j} \nonumber [\nabla_\vtheta J(\pi_{H,\vtheta_k})]_i^2 [\nabla_\vtheta J(\pi_{H,\vtheta_k})]_j^2\right)\\
    =& 3 \left(\sum_{i = 1}^d [\nabla_\vtheta J(\pi_{H,\vtheta_k})]_i^2\right)^2 = 3 \| \nabla_\vtheta J(\pi_{H,\vtheta_k}) \|_2^4.
\end{align}
Substituting~\eqref{eq: 2 moment} and~\eqref{eq: 4 moment} into~\eqref{eq:paley-zygmund}, we have:
\begin{align} \label{eq: square bound}
    \sP \left( \left( \nabla_\vtheta J(\pi_{H,\vtheta_k})^\top \vv_k \right)^2 > \frac{9 \beta^2 L_J^2 d^2}{4} \right)
    \ge
    (1 - \eta^2)^2 \frac{\| \nabla_\vtheta J(\pi_{H,\vtheta_k}) \|_2^4}{3 \| \nabla_\vtheta J(\pi_{H,\vtheta_k}) \|_2^4}
    = \frac{(1 - \eta^2)^2}{3},
\end{align}
Finally, using~\eqref{eq: square bound},~\eqref{eq: symmetric}, ~\eqref{eq: improve guarantee}, and substituting $\eta = 3/4$, we have:
\begin{align*}
    \prob\left( J(\pi_{H,\vtheta_k'}) < J(\pi_{H,\vtheta_k}) - \beta^2 L_J d \right) &\ge \frac{1}{2}\sP \left( \left( \nabla_\vtheta J(\pi_{H,\vtheta_k})^\top \vv_k \right)^2 > \frac{9 \beta^2 L_J^2 d^2}{4} \right) \geq \frac{(1 - \eta^2)^2}{6} \ge \frac{1}{32},
\end{align*}
which completes the proof.
\end{proof}

%% file: sections/proofs/thm_main.tex
\mainthm*

\begin{proof}
    According to Theorem~\ref{thm:positive_prob_per_iteration}, we can obtain 
    \begin{align*}
        \prob\left( \left. J(\pi_{H,\vtheta_{k+1}}) \leq J(\pi_{H,\vtheta_k}) - \beta^2 L_J d \right| \vtheta_k = \vtheta\right)
        \geq p \mathbbm{1}\{\left\| \nabla_\vtheta J(\pi_{H,\vtheta}) \right\|_2 \geq 2\beta L_J d\}.
    \end{align*}
    By the update rule of Algorithm~\ref{alg:zeroth-order}, the function value $J(\pi_{H,\vtheta_{k+1}})$ after policy iteration $k$ can only decrease compared to $J(\pi_{H,\vtheta_k})$, so we have:
    \begin{align*}
        &\E\left[ \left. J(\pi_{H,\vtheta_{k+1}}) - J(\pi_{H,\vtheta_k}) \right| \vtheta_k= \vtheta\right]\\ \leq& \E\left[ \left. \left(J(\pi_{H,\vtheta_{k+1}}) - J(\pi_{H,\vtheta_k})\right) \mathbbm{1}\left\{ J(\pi_{H,\vtheta_{k+1}}) \leq J(\pi_{H,\vtheta_k}) - \beta^2 L_J d \right\} \right| \vtheta_k = \vtheta\right]\\
        & + \E\left[ \left. \left(J(\pi_{H,\vtheta_{k+1}}) - J(\pi_{H,\vtheta_k})\right) \mathbbm{1}\left\{ J(\pi_{H,\vtheta_{k+1}}) > J(\pi_{H,\vtheta_k}) - \beta^2 L_J d \right\} \right| \vtheta_k = \vtheta\right]\\
        \leq & - \beta^2 L_J d \prob\left( \left. J(\pi_{H,\vtheta_{k+1}}) \leq J(\pi_{H,\vtheta_k}) - \beta^2 L_J d \right| \vtheta_k = \vtheta\right) + 0\\
        \leq & - p \beta^2 L_J d \mathbbm{1}\{\left\| \nabla_\vtheta J(\pi_{H,\vtheta}) \right\|_2 \geq 2\beta L_J d\}.
    \end{align*}
    Taking the expectation over random variables $\vtheta_k$, we obtain 
    \begin{align*}
        \E\left[ J(\pi_{H,\vtheta_{k+1}}) - J(\pi_{H,\vtheta_k}) \right] \leq - p \beta^2 L_J d \prob\left(\left\| \nabla_\vtheta J(\pi_{H,\vtheta_k}) \right\|_2 \geq 2\beta L_J d\right).
    \end{align*}
Taking a telescoping sum over $k=0,1,\cdots,K-1$ yields 
    \begin{align*}
        \E\left[ J(\pi_{H,\vtheta_K}) \right] - J(\pi_{H,\vtheta_0}) \leq - p \beta^2 L_J d \sum_{k=0}^{K-1} \prob\left(\left\| \nabla_\vtheta J(\pi_{H,\vtheta_k}) \right\|_2 \geq 2\beta L_J d\right),
    \end{align*}
which implies that 
    \begin{align*}
        \frac{1}{K} \sum_{k=0}^{K-1} \prob\left(\left\| \nabla_\vtheta J(\pi_{H,\vtheta_k}) \right\|_2 \geq 2\beta L_J d\right) \leq \frac{J(\pi_{H,\vtheta_0}) - \E\left[ J(\pi_{H,\vtheta_K}) \right]}{p \beta^2 L_J d K} \leq \frac{J(\pi_{H,\vtheta_0}) - J^*}{p \beta^2 L_J d K},
    \end{align*}
    where the last inequality holds because $J^*$ is a lower bound on the average cost. Since $r$ is sampled from $\{0, 1, \cdots, K-1\}$ uniformly at random, we have
    \begin{align*}
        \prob\left(\left\| \nabla_\vtheta J(\pi_{H,\vtheta_\rr}) \right\|_2 \geq 2\beta L_J d\right) =& \sum_{k=0}^{K-1} \prob\left(\vtheta_{\rr} = \vtheta_k\right)\prob\left(\left\| \nabla_\vtheta J(\pi_{H,\vtheta_k}) \right\|_2 \geq 2\beta L_J d\right)\\
        =& \frac{1}{K} \sum_{k=0}^{K-1} \prob\left(\left\| \nabla_\vtheta J(\pi_{H,\vtheta_k}) \right\|_2 \geq 2\beta L_J d\right)
        \leq  \frac{J(\pi_{H,\vtheta_0}) - J^*}{p \beta^2 L_J d K}.
    \end{align*}
    Therefore, for any $\omega \in (0,1)$, by choosing the perturbation distance and learning rate to be $\alpha = \beta = \varepsilon/ 2 L_J d$ and $K$ sufficinetly large such that 
    \begin{align*}
        K \geq  \frac{4 L_J d}{p \omega \varepsilon^2} [J(\pi_{H,\vtheta_0}) - J^*],
    \end{align*}
    we have $\prob\left(\left\| \nabla_\vtheta J(\pi_{H,\vtheta_\rr}) \right\|_2 \geq \varepsilon\right) \leq \omega.$ This proves the first part of the the theorem. Under  Algorithm~\ref{alg:zeroth-order}, the average cost $J(\pi_{H, \vtheta_k})$ is nonincreasing , so $J(\pi_{H, \vtheta_K}) \leq J(\pi_{H,\vtheta_\rr})$ almost surely.
\end{proof}

%% file: sections/proofs/thm_partition.tex
\partitionthm*

\begin{proof}
    For iteration $k$ and partition $j$, let $O_{k,j} = \mathbbm{1}\left\{ J(\pi_{H,\vtheta_{k,j}'}) < J(\pi_{H,\vtheta_k})\right\} \in \{0,1\}$ denote the performance comparison indicator between the original policy $\pi_{H,\vtheta_k}$ and the perturbed policy $\pi_{H,\vtheta_{k,j}'}$.
    According to Algorithm~\ref{alg:zeroth-order-partition}, we have
    \begin{align}
        \vtheta_{k+1} = \vtheta_k + \alpha \sum_{j=1}^D O_{k,j} \vv_k \circ \ve_{\gD_j}.
        \label{eq: partitionZO_update}
    \end{align}

    Based on the smoothness proved in Lemma~\ref{lemma: smoothness}, we have
    \begin{align}
        J(\pi_{H,\vtheta_{k+1}}) - J(\pi_{H,\vtheta_k}) &\le \left\langle \nabla_\vtheta J(\pi_{H,\vtheta_k}), \vtheta_{k+1} - \vtheta_k \right\rangle + \frac{L_J}{2} \| \vtheta_{k+1} - \vtheta_k \|_2^2 \nonumber \\
        &= \alpha \sum_{j=1}^D \left\langle \nabla_\vtheta J(\pi_{H,\vtheta_k}), O_{k,j} \vv_k \circ \ve_{\gD_j} \right\rangle + \frac{L_J}{2} \alpha^2 \| \sum_{j=1}^D O_{k,j} \vv_k \circ \ve_{\gD_j} \|_2^2 \nonumber \\
        &\le \alpha \sum_{j=1}^D \left\langle \nabla_\vtheta J(\pi_{H,\vtheta_k}), O_{k,j} \vv_k \circ \ve_{\gD_j} \right\rangle + \frac{\alpha^2 L_J d}{2},
        \label{eq: partition_per_step}
    \end{align}
    where the first inequality follows from the smoothness in Lemma~\ref{lemma: smoothness},
    the first equality is by substituting the update rule~\eqref{eq: partitionZO_update},
    and the last inequality holds because
    \[
    \| \sum_{j=1}^D O_{k,j} \vv_k \circ \ve_{\gD_j} \|_2^2 = \sum_{j=1}^D O_{k,j} \| \vv_k \circ \ve_{\gD_j} \|_2^2 \le \sum_{j=1}^D \| \vv_k \circ \ve_{\gD_j} \|_2^2 = \| \vv_k \|_2^2 = d.
    \]
    The proof of Lemma 2 in~\cite{deyi2026_fl} has shown that for each iteration $k$ and partition $j$,
    \begin{align}
        \E \left[\left. \left\langle \nabla_\vtheta J(\pi_{H,\vtheta_k}), O_{k,j} \vv_k \circ \ve_{\gD_j} \right\rangle \right| \vtheta_k \right] \le -\frac{\| \nabla_\vtheta J(\pi_{H,\vtheta_k}) \circ \ve_{\gD_j} \|_2}{2 \sqrt{3}} + \frac{\beta L_J |\gD_j|}{2}.
        \label{eq: restate_fl_lemma2}
    \end{align}
    Taking expectation conditioning on $\vtheta_k$ on both sides of~\eqref{eq: partition_per_step} and substituting~\eqref{eq: restate_fl_lemma2} gives,
    \begin{align}
        \E\left[\left. J(\pi_{H,\vtheta_{k+1}}) - J(\pi_{H,\vtheta_k})\right| \vtheta_k \right] &\le \alpha \sum_{j=1}^D \E \left[\left. \left\langle \nabla_\vtheta J(\pi_{H,\vtheta_k}), O_{k,j} \vv_k \circ \ve_{\gD_j} \right\rangle \right| \vtheta_k \right] + \frac{\alpha^2 L_J d}{2} \nonumber \\
        &\le -\frac{\alpha}{2 \sqrt{3}} \sum_{j=1}^D \| \nabla_\vtheta J(\pi_{H,\vtheta_k}) \circ \ve_{\gD_j} \|_2 + \frac{\alpha \beta L_J}{2} \sum_{j=1}^D |\gD_j| + \frac{\alpha^2 L_J d}{2} \nonumber \\
        &= -\frac{\alpha}{2 \sqrt{3}} \sum_{j=1}^D \| \nabla_\vtheta J(\pi_{H,\vtheta_k}) \circ \ve_{\gD_j} \|_2 + \alpha^2 L_J d,
        \label{eq: partition_recursion}
    \end{align}
    where the last equality holds because the definition of partition and $\sum_{j=1}^D |\gD_j| = d$, and $\alpha = \beta$.

    Taking telescoping sum of~\eqref{eq: partition_recursion} over $k=0,1,\dots,K-1$ and applying total expectation law give
    \begin{align}
        \frac{\sum_{k=0}^{K-1} \E\left[ \sum_{j=1}^D \| \nabla_\vtheta J(\pi_{H,\vtheta_k}) \circ \ve_{\gD_j} \|_2 \right]}{K} &\le \frac{2 \sqrt{3} \E\left[ J(\pi_{H,\vtheta_0}) - J(\pi_{H,\vtheta_K}) \right]}{\alpha K} + 2 \sqrt{3} \alpha L_J d \nonumber \\
        &\le \frac{2 \sqrt{3} \left( J(\pi_{H,\vtheta_0}) - J^* \right)}{\alpha K} + 2 \sqrt{3} \alpha L_J d,
    \end{align}
    where $J^*$ lower bounds the average cost.
    Since for any vector $\vv$, $\sum_{j=1}^D \| \vv \circ \ve_{\gD_j} \|_2 \ge \|\vv\|_2$, we have
    \begin{align}
        \E\left[ \| \nabla_\vtheta J(\pi_{H,\vtheta_\rr}) \|_2 \right] = \frac{\sum_{k=0}^{K-1} \E\left[ \| \nabla_\vtheta J(\pi_{H,\vtheta_k}) \|_2 \right]}{K} \le \frac{2 \sqrt{3} \left( J(\pi_{H,\vtheta_0}) - J^* \right)}{\alpha K} + 2 \sqrt{3} \alpha L_J d.
        \label{eq: partition_preconvergence}
    \end{align}
    Substituting $\alpha = \sqrt{\left( J(\pi_{H,\vtheta_0}) - J^* \right)/L_J d K}$ into~\eqref{eq: partition_preconvergence} gives
    \[
    \E\left[ \| \nabla_\vtheta J(\pi_{H,\vtheta_\rr}) \|_2 \right] \le 4 \sqrt{3} \sqrt{\frac{\left( J(\pi_{H,\vtheta_0}) - J^* \right) L_J d}{K}},
    \]
    which completes the proof.
\end{proof}